%% file: main.tex
\documentclass{article}

\usepackage{microtype}
\usepackage{graphicx}
\usepackage{subfigure}
\usepackage{booktabs} 

\usepackage{hyperref}

\usepackage[accepted]{icml2019}

\icmltitlerunning{Recurrent Kalman Networks:  Factorized Inference in High-Dimensional Deep Feature Spaces}

\usepackage{amsmath}
\usepackage{amssymb}
\usepackage{gensymb}

\usepackage{pgfplots}
\usetikzlibrary{pgfplots.groupplots}
\usetikzlibrary{backgrounds}
\pgfplotsset{compat=1.13}
\usepackage{tikz}
\usetikzlibrary{positioning, arrows, automata}
\newcommand*{\MyScale}{1}%

\newcommand*{\MyResizeBox}[2]{%
    \sbox0{#2}%
    \pgfmathsetmacro{\MyScale}{#1/\wd0}%
    #2%
}%
\newcommand{\cvec}[1]{\boldsymbol{\mathrm{#1}}}
\newcommand{\cmat}[1]{\boldsymbol{\mathrm{#1}}}

\renewcommand{\cite}[1]{\citep{#1}}

\begin{document}
\include{content/tikzGraphs}

\twocolumn[
\icmltitle{Recurrent Kalman Networks: \\ Factorized Inference in High-Dimensional Deep Feature Spaces}



\begin{icmlauthorlist}
\icmlauthor{Philipp Becker}{tud,bcai,tue}
\icmlauthor{Harit Pandya}{linc}
\icmlauthor{Gregor Gebhardt}{tud}
\icmlauthor{Cheng Zhao}{birm}
\icmlauthor{James Taylor}{lanc}
\icmlauthor{Gerhard Neumann}{linc,bcai,tue}
\end{icmlauthorlist}

\icmlaffiliation{tud}{Computational Learning for Autonomous Systems, TU Darmstadt, Darmstadt, Germany.}
\icmlaffiliation{linc}{Lincoln Center for Autonomous Systems, University of Lincoln, Lincoln, UK.}
\icmlaffiliation{birm}{Extreme Robotics Lab, University of Birmingham, Birmingham, UK.}
\icmlaffiliation{lanc}{Engineering Department, Lancaster University, Lancaster, UK.}
\icmlaffiliation{bcai}{Bosch Center for Artificial Intelligence, Renningen, Germany.}
\icmlaffiliation{tue}{University of T\"ubingen, T\"ubingen, Germany.}

\icmlcorrespondingauthor{Philipp Becker}{philippbecker93@googlemail.com}

\icmlkeywords{Machine Learning, ICML}

\vskip 0.3in
]



\printAffiliationsAndNotice{}

\begin{abstract}
In order to integrate uncertainty estimates into deep time-series modelling, Kalman Filters (KFs) \cite{kalman1960filter} have been integrated with deep learning models, however, such approaches typically rely on approximate inference techniques such as variational inference which makes learning more complex and often less scalable due to approximation errors. 
We propose a new deep approach to Kalman filtering which can be learned directly in an end-to-end manner using backpropagation without additional approximations.
Our approach uses a high-dimensional factorized latent state representation for which the Kalman updates simplify to scalar operations and thus avoids hard to backpropagate, computationally heavy and potentially unstable matrix inversions. 
Moreover, we use locally linear dynamic models to efficiently propagate the latent state to the next time step.
The resulting network architecture, which we call \emph{Recurrent Kalman Network} (RKN), can be used for any time-series data, similar to a LSTM \citep{hochreiter1997long} but uses an explicit representation of uncertainty. 
As shown by our experiments, the RKN obtains much more accurate uncertainty estimates than an LSTM or Gated Recurrent Units (GRUs) \cite{cho2014gru} while also showing a slightly improved prediction performance and outperforms various recent generative models on an image imputation task.
\end{abstract}

\input{content/1-Introduction.tex}
\input{content/2-RKN.tex}
\input{content/3-EvalAndExp.tex}

\input{content/4-Conclusion.tex}
\bibliography{main.bib}
\bibliographystyle{icml2019}

\clearpage
\section*{Appendix}
\input{content/A-Preliminaries.tex}
\input{content/B-Appendix.tex}

\end{document}

%% file: content/tikzGraphs.tex
\newcommand{\tikzMyModel}{
\begin{tikzpicture}[scale=\MyScale, thick]
		\tikzset{invisible/.style={draw=none},
	}

	\draw[densely dotted] (4, -1.7)   rectangle  (8, 1.7) node[pos=.5, align=center]{};
	\draw (9.5, -0.5) -- (9.5, 0.5) -- (10.5, 1.5) -- (10.5, -1.5) -- (9.5, -0.5) node[black, pos=.5, rotate=90, xshift=25]{Decoder};
	\draw (1.5, -1.5) -- (1.5, 1.5) -- (2.5, 0.5) -- (2.5, -0.5) -- (1.5, -1.5) node[pos=.5, rotate=90, xshift=25]{Encoder};
	\draw (5, 0.5) rectangle (7, 1.5) node[pos=.5, align=center]{Predict}; 
	\draw (5, -.5) rectangle (7, -1.5) node[pos=.5, align=center]{Update}; 

	\draw[->, >=stealth] (0, 0) -- ( 1.5, 0) node[pos=.5, yshift= 5]{$\cvec{o}_t$};
	\draw[-] (2.5, 0) -- ( 4,0) node[pos=.5, yshift=-10]{$\cvec{\sigma}_t^\mathrm{obs}$};
	\draw[-] (2.5, 0) -- ( 4, 0) node[pos=.5, yshift=5]{$\cvec{w}_t$};
	\draw[->, >=stealth] (4, 0) -- (4.5, 0) -- (4.5, -1.0) -- (5.0, -1.0);  
    \draw[->, >=stealth] (7, -1) -- (7.5, -1) -- (7.5, 0) -- (8, 0) -- (7.5, 0) -- (7.5,1 ) -- (7,1);
    \draw[->, >=stealth] (6, 0.5) -- (6, -0.5) node[pos=.5, xshift=-7]{$\cvec{z}_t^-$};
    \draw[->, >=stealth] (6, 0.5) -- (6, -0.5) node[pos=.5, xshift=12]{$\cmat{\Sigma}_t^-$};
	\draw[->, >=stealth] (8, 0) -- ( 9.5, 0) node[pos=.5, yshift=7]{$\cvec{z}_{t}^+$};
	\draw[->, >=stealth] (8, 0) -- ( 9.5, 0) node[pos=.5, yshift=-8] {$\cmat{\Sigma}_{t}^+$};

	\draw[->, >=stealth] (10.5, -1) -- (12, -1) node[black, pos=.5, yshift=-10]{$\cvec{s}_{t}^+, \cvec{\sigma}_t^+ $};
	\draw[->, >=stealth] (10.5, 1) -- (12, 1) node[black, pos=.5, yshift=-10]{$\cvec{o}^+_{t}$};

	
\end{tikzpicture}
}

\newcommand{\tikzMyModelSave}{
\begin{tikzpicture}[scale=\MyScale, thick]
		\tikzset{invisible/.style={draw=none},
	}

	\draw (4, 1)   rectangle  (7, 2) node[pos=.5, align=center]{Transition Layer \\ ($\cmat{A}_t$ learned)};
	\draw (8.5, 1) -- (8.5, 2) -- (9.5, 3) -- (9.5, 0) -- (8.5, 1) node[black, pos=.5, rotate=90, xshift=27.5]{Decoder};
	\draw (1.5, 0) -- (1.5, 3) -- (2.5, 2) -- (2.5, 1) -- (1.5, 0) node[pos=.5, rotate=90, xshift=27.5]{Encoder};
	
	\draw[<-, >=stealth] (5, 2) arc(45:-225:-0.75) node[pos=.5, yshift=9] {$(\cvec{z}_{t-1}^+, \cvec{\sigma}_{t-1}^{\textrm{upper},+}, \cvec{\sigma}_{t-1}^{\textrm{lower},+}, \cvec{\sigma}_{t-1}^{\textrm{side},+})$};
	
	\draw[->, >=stealth] (0, 1.5) -- ( 1.5, 1.5) node[pos=.5, yshift= 5]{$\cvec{o}_t$};
	\draw[->, >=stealth] (2.5, 1.5) -- ( 4, 1.5) node[pos=.5, yshift=-10]{$\cvec{\sigma}_t^\mathrm{obs}$};
	\draw[->, >=stealth] (3, 1.5) -- ( 4, 1.5) node[pos=.5, yshift=5, xshift=-10]{$\cvec{w}_t$};
	\draw[->, >=stealth] (7, 1.5) -- ( 8.5, 1.5) node[pos=.5, yshift=7, xshift=-10]{$\cvec{z}_{t}^+$};
	\draw[->, >=stealth] (7, 1.5) -- ( 8.5, 1.5) node[pos=.5, yshift=-8] {$\cvec{\sigma}_{t}^{\textrm{upper},+}$};
	\draw[->, >=stealth] (7, 1.5) -- ( 8.5, 1.5) node[pos=.5, yshift=-23] {$\cvec{\sigma}_{t}^{\textrm{lower},+}$};
	\draw[->, >=stealth] (7, 1.5) -- ( 8.5, 1.5) node[pos=.5, yshift=-38] {$\cvec{\sigma}_{t}^{\textrm{side},+}$};
	
	\draw[->, >=stealth] (9.5, 0.5) -- (11, 0.5) node[black, pos=.5, yshift=-10]{$\left(\cvec{s}_{t}^+, \cvec{\sigma}_t^+ \right)$};
	\draw[->, >=stealth] (9.5, 2.5) -- (11, 2.5) node[black, pos=.5, yshift=-10]{$\left(\cvec{o}^+_{t}, \sigma_t^+ \right)$};

	
\end{tikzpicture}
}
\newcommand{\tikzBackpropKF}{
\begin{tikzpicture}[scale=\MyScale, thick]
		\tikzset{invisible/.style={draw=none},
	}

	\draw (4, 1)   rectangle  (7, 2) node[pos=.5, align=center]{(Extended) \\ Kalman Filter};
	\draw (8, 1)   rectangle  (9, 2) node[pos=.5]{$\cmat{H}_s$};
	\draw (2, 0) -- (2, 3) -- (3, 2) -- (3, 1) -- (2, 0) node[pos=.5, rotate=90, xshift=25]{CNN};
	
	\draw[<-, >=stealth] (5, 1) arc(-45:225:-0.75) node[pos=.5, yshift=-9] {$(\cvec{x}_{t-1}^+, \cmat{\Sigma}_{t-1}^+)$};
	
	\draw[->, >=stealth] (1, 1.5) -- ( 2, 1.5) node[pos=.5, yshift=  5]{$\cvec{o}_t$};
	\draw[->, >=stealth] (3, 1.5) -- ( 4, 1.5) node[pos=.5, yshift=-10]{$\cmat{\Sigma}_t^\mathrm{obs}$};
	\draw[->, >=stealth] (3, 1.5) -- ( 4, 1.5) node[pos=.5, yshift=  5]{$\cvec{w}_t$};
	\draw[->, >=stealth] (7, 1.5) -- ( 8, 1.5) node[pos=.5, yshift=-10]{$\cvec{x}_{t}^+$};
	\draw[->, >=stealth] (7, 1.5) -- ( 8, 1.5);
	\draw[->, >=stealth] (9, 1.5) -- (10, 1.5) node[pos=.5, yshift=-10]{$\cvec{s}_{t}^+$};

	
\end{tikzpicture}
}

\newcommand{\tikzBackpropKFCell}{
\begin{tikzpicture}[scale=\MyScale, thick]
	\draw (0, 0.5) rectangle (1, 1.25) node[pos=0.5]{$\cmat{A}\cvec{x}^+_{t-1}$};
	\draw (3, 0.5) rectangle (6, 1.25) node[pos=0.5]{$\cmat{A}\cmat{\Sigma}_{t-1}^+\cmat{A}^T + \cmat{\Sigma}^{trans}$};
	
	\draw (2, 1.75) rectangle (6, 2.5) node[pos=0.5]{$\cmat{\Sigma}_t^- \cmat{H}^T \left(\cmat{H} \cmat{\Sigma}^-_t \cmat{H}^T + \cmat{\Sigma}_t^{obs} \right)^{-1}$};
	
	\draw (0  , 3) rectangle (3, 3.75) node[pos=0.5]{$\cvec{x}_t^- + \cmat{Q}_t \left(\cvec{w}_t - \cmat{H}\cvec{x}_t^-  \right)$};
	\draw (3.5, 3) rectangle (6, 3.75) node[pos=0.5]{$\left( \cmat{I} - \cmat{Q}_t\cmat{H} \right)^{-1} \cmat{\Sigma}_t^-$};
	
    \draw [->, >=stealth] (0.5, 0   ) -- (0.5  , 0.5 ) node[pos=.5, xshift= 12] {$\cvec{x}_{t-1}^+$};
    \draw [->, >=stealth] (4.5, 0   ) -- (4.5, 0.5 ) node[pos=.5, xshift=-12] {$\cmat{\Sigma}_{t-1}^+$};
    \draw [->, >=stealth] (4.5, 1.25) -- (4.5, 1.75) node[pos=.5, xshift=-10] {$\cmat{\Sigma}_{t}^-$};
    \draw [->, >=stealth] (5 ,  2.5 ) -- (5,   3   ) node[pos=.5, xshift=-10] {$\cmat{Q}_{t}$};
    \draw [->, >=stealth] (2.5, 2.5 ) -- (2.5, 3   ) node[pos=.5, xshift=-10] {$\cmat{Q}_{t}$};
    \draw [->, >=stealth] (0.5,   1.25) -- (0.5  , 3   ) node[pos=.5, xshift=-10] {$\cvec{x}_{t}^-$};
    \draw [->, >=stealth] (0.5,   3.75) -- (0.5,   4.25) node[pos=.5, xshift= 10] {$\cvec{x}_{t}^+$};
    \draw [->, >=stealth] (4.5, 3.75) -- (4.5, 4.25) node[pos=.5, xshift=-10] {$\cmat{\Sigma}_{t}^+$};
    \draw [->, >=stealth] (1.5, 0) -- (1.5, 3) node[pos=.5, xshift=-10] {$\cmat{w}_{t}$};
    \draw [->, >=stealth] (2.5, 0) -- (2.5, 1.75) node[pos=.5, xshift=-10] {$\cmat{\Sigma}_{t}^{obs}$};
    
    \draw [dashed, ->, >=stealth] (0.5,   4) -- (-0.5, 4) -- (-0.5, 0.25) -- (  0.5, 0.25);
    \draw [dashed, ->, >=stealth] (4.5, 4) -- ( 6.5, 4) -- ( 6.5, 0.25) -- (4.5, 0.25);
\end{tikzpicture}
}

\newcommand{\tikzBiHMM}{
\begin{tikzpicture}[scale=\MyScale, thick, auto, node distance=1.5cm, ->,>=stealth,
 state/.style={circle, draw, minimum size=1cm}]
\tikzset{inv state/.style={draw=none},}

\node[state]                	(x_t0)	 	{$\cvec{x}_t$};
\node[state, right of=x_t0]		(x_t1)  	{$\cvec{x}_{t+1}$};
\node[state, right of=x_t1]		(x_t2)  	{$\cvec{x}_{t+2}$};
\node[state, below of=x_t0]	    (o_t0)		{$\cvec{o}_t$};
\node[state, below of=x_t1]	    (o_t1)		{$\cvec{o}_{t+1}$};
\node[state, below of=x_t2]	    (o_t2)		{$\cvec{o}_{t+2}$};

\node[inv state, left of=x_t0] (x_s)		{};
\node[inv state, right of=x_t2] (x_d)		{};

\path	(x_s) 	edge node{} (x_t0)	
		(x_t0)	edge node{}	(x_t1)
		(x_t1)	edge node{}	(x_t2)
		(x_t2)	edge node{}	(x_d)
		(x_t0)	edge node{}	(o_t0)
		(x_t1)	edge node{}	(o_t1)
		(x_t2)	edge node{}	(o_t2);
\end{tikzpicture}
}

%% file: content/1-Introduction.tex
\section{Introduction}
State-estimation in unstructured environments is a very challenging task as observations or measurements of the environment are often high-dimensional and only provide partial information about the state. 
Images are a good example: Even for low resolution, the number of pixels can quickly exceed tens or hundreds of thousands and it is impossible to obtain any information about the dynamics, such as velocities, from a single image.
Additionally, the observations may be noisy or may not contain useful information for the task at hand. Such noise can, for example, be introduced by poor illumination or motion blur and occlusions can prevent us from observing some or all relevant aspects of the scene.
In addition to state estimation, it is also often desirable to predict future states or observations, for example, in order to assess the consequences of future actions. 
To this end, an initial estimate of the current state is necessary which again has to be inferred from observations.
In such environments, we typically also have to deal with high uncertainties in the state estimates. 
Being able to model this uncertainty is crucial in many decision making scenarios, e.g., if we need to decide to perform an action now or wait until more information about the scene is available.

Deep learning models have been very successful for time-series modelling in unstructured environments. 
Classical models such as LSTMs \citep{hochreiter1997long} or GRUs \cite{cho2014gru} perform well but fail to capture the uncertainty of the state estimate.
Recent probabilistic deep learning approaches have used the Kalman Filter (KF) as a tool to integrate uncertainty estimates into deep time-series modelling \cite{haarnoja2016backprop,watter2015embed,archer2015blackBoxVI,fraccaro2017kvae,krishnan2017sin}. 
These approaches use the KF to perform inference in a low-dimensional (latent) state space that is typically defined by a deep encoder.
However, using KF in such a state space comes with two main limitations.
In order to be usable for non-linear dynamics, we have to introduce approximations such as the extended KF \cite{haarnoja2016backprop} and variational inference methods \cite{krishnan2017sin,fraccaro2017kvae}.
Moreover, the KF equations require computationally expensive matrix inversions that are hard to scale to high dimensional latent spaces for more complex systems and computationally demanding to fully backpropagate in an end-to-end manner.
Most of these methods are implemented as (variational) auto-encoders and are therefore also limited to predicting future observations or imputing missing observations and can not be directly be applied to state estimation. 

We introduce the \emph{Recurrent Kalman Network}, an end-to-end learning approach for Kalman filtering and prediction. 
While Kalman filtering in the original state space requires approximations due to the non-linear models, the RKN uses a learned high-dimensional latent state representation that allows for efficient inference using locally linear transition models and a factorized belief state representation.
Exploiting this representation allows us to avoid the expensive and numerically problematic matrix inversions involved in the KF equations.

Conceptually, this idea is related to kernel methods which use high-dimensional feature spaces to approximate nonlinear functions with linear models \cite{gebhardt2017kkr}.
However, in difference to kernel feature spaces, our feature space is given by a deep encoder and learned in an end-to-end manner. 

The RKN can be used for any time-series data set for which LSTMs and GRUs are currently the state of the art.
In contrast to those, the RKN uses an explicit representation of uncertainty which governs the gating between keeping the current information in memory or updating it with the current observation.
While the RKN shows a slightly improved performance in terms of state estimation errors, both LSTMs and GRUs struggle with estimating the uncertainty of the prediction while the RKN can provide accurate uncertainty estimates.
In relation to existing KF-based approaches, our approach can be used for state estimation as well as for generative tasks such as image imputation.
We also show that we outperform state of the art methods on a complex image imputation task.

\subsection{Related Work}
Using encoders for time-series modelling of high-dimensional data such as images is a common approach. 
Such encoders can also be easily integrated with well known deep time-series models such as LSTMs \cite{hochreiter1997long} or GRUs \cite{cho2014gru}. 
These models are very effective but do not provide good uncertainty estimates as shown in our experiments.

Pixel to Torques (P2T) \cite{wahlstrom2015pixelsToTorques} employs an autoencoder to obtain low dimensional latent representations from images together with a transition model.
They subsequently use the models to perform control in the latent space.
Embed to Control (E2C) \cite{watter2015embed} can be seen as an extension of the previous approach with the difference that a variational autoencoder \cite{kingma2013auto} is used.
However, both of these approaches are not recurrent and rely on observations which allow inferring the whole state from a single observation.
They can therefore not deal with noisy or missing data.

The BackpropKF \cite{haarnoja2016backprop} applies a CNN to estimate the observable parts of the true state given the observation. 
Similar to our approach, this CNN additionally outputs a covariance matrix indicating the model's certainty about the estimate and allows the subsequent use of an (extended) Kalman filter with known transition model.
In contrast, we let our model chose the feature space that is used for the inference such that locally linear models can be learned and the KF computations can be simplified due to our factorization assumptions. 

Another family of approaches interprets encoder-decoder models as latent variable models that can be optimized efficiently by variational inference.
They derive a corresponding lower bound and optimize it using the stochastic gradient variational Bayes approach \citep{kingma2013auto}.
Black Box Variational Inference (BB-VI) \cite{archer2015blackBoxVI} proposes a structured Gaussian variational approximation of the posterior, which simplifies the inference step at the cost of maintaining a tri-diagonal covariance matrix of the full state.
To circumvent this issue, Structured Inference Networks (SIN) \cite{krishnan2017sin} employ a flexible recurrent neural network to approximate the dynamic state update.
Deep Variational Bayes Filters (DVBF) \cite{karl2016dvbf} integrate general Bayes filters into deep feature spaces while the Kalman Variational Autoencoder (KVAE) \cite{fraccaro2017kvae} employs the classical Kalman Filter and allows not only filtering but also smoothing. 
Variational Sequential Monte Carlo (VSMC) \cite{naesseth2017vsmc} uses particle filters instead, however, they are only learning the proposal function and are not working in a learned latent space.
Disentangled Sequential Autoencoder (DSA) \cite{yingzhen2018disentangled} explicitly partitions latent variables for separating static and dynamic content from a sequence of observations however, they do not model state space noise.
Yet, these models cannot be directly used for state estimation as they are formulated as generative models of the observations without the notion of the real state of the system. 
Moreover, the use of variational inference introduces an additional approximation that might affect the performance of the algorithms.

\begin{table}
\begin{center}
\caption{Qualitative comparison of our approach to recent related work.}
\label{tab:qual_comp}
\vskip 0.15in
\begin{tabular}{lccccc}
\toprule
 & scale-  & state   & uncer- & noise  & direct \\
& able & est. & tainty &   & opt.\\
\midrule
LSTM  & $\checkmark$ & $\checkmark$ & $\times$/$\checkmark$ & $\checkmark$ & $\checkmark$ \\ 
GRU & $\checkmark$ & $\checkmark$ & $\times$/$\checkmark$ & $\checkmark$ & $\checkmark$ \\ 
\midrule
P2T & $\checkmark$ & $\checkmark$ & $\times$/$\checkmark$ & $\times$ & $\checkmark$ \\
E2C & $\checkmark$ & $\times$  & $\checkmark$ & $\times$ & $\times$  \\
\midrule
BB-VI & $\times$ & $\times$ & $\checkmark$ & $\checkmark$ & $\times$  \\
SIN & $\checkmark$ & $\times$ & $\checkmark$ & $\checkmark$  & $\times$ \\
KVAE & $\times$ & $\times$ & $\checkmark$ & $\checkmark$ & $\times$ \\
DVBF & $\checkmark$ & $\times$ & $\checkmark$ & $\checkmark$ & $\times$ \\
VSMC & $\checkmark$ & $\times$ & $\checkmark$ & $\checkmark$  & $\times$  \\
DSA & $\checkmark$ & $\times$ & $\checkmark$ & $\times$  & $\times$  \\
\midrule
DSSM & $\times$ & $\checkmark$ (1D) & $\checkmark$ & $\times$ & $\checkmark$ \\
PRSSM & $\times$ & $\checkmark$  & $\checkmark$  & $\checkmark$ & $\times$\\
\midrule
\textbf{RKN} & $\checkmark$ & $\checkmark$ & $\checkmark$ & $\checkmark$  & $\checkmark$ \\
\bottomrule
\end{tabular}
\end{center}
\vskip -0.1in
\end{table}

Recently, \citet{rangapuram2018dssm} introduced Deep State Space Models for Time Series Forecasting (DSSM). 
They employ a recurrent network to learn the parameters of a time varying linear state space model. The emissions of that state space model are the model's final output, i.e. there is no decoder.
While this makes the likelihood of the targets given the predicted state space model parameters tractable it limits them to simple latent spaces.
Additionally, their approach is only derived for $1$ dimensional output data. 
Probabilistic Recurrent State-Space Models (PRSSM) \cite{doerr2018prssm} use Gaussian processes to capture state uncertainties, however, their approach is not demonstrated for high dimensional observations such as images. 

A summary of all the approaches and their basic properties can be seen in Table~\ref{tab:qual_comp}.
We compare whether the approaches are scalable to high dimensional latent states, whether they can be used for state estimation, whether they can provide uncertainty estimates, whether they can handle noise or missing data, and whether the objective can be optimized directly or via a lower bound.
All probabilistic generative models rely on variational inference which optimizes a lower bound, which potentially affects the performance of the algorithms.
P2T and E2C rely on the Markov assumption and therefore can not deal with noise or need very large window sizes.
The approaches introduced in   \cite{archer2015blackBoxVI,fraccaro2017kvae,haarnoja2016backprop} directly use the Kalman update equations in the latent state, which limits these approaches to rather low dimensional latent states due to the expensive matrix inversions.

Traditional recurrent models such as  LSTMs or GRUs can be trained directly by backpropagation through time and therefore typically yield very good performance but are lacking uncertainty estimates.
However, as in our experiments, they can be artificially added.
Our RKN approach combines the advantages of all methods above as it can be learned by direct optimization without the use of a lower bound and it provides a principled way of representing uncertainty inside the neural network. 

%% file: content/2-RKN.tex
\section{Factorized Inference in Deep Feature Spaces}

Lifting the original input space to a high-dimensional feature space where linear operations are feasible is a common approach in machine learning, e.g., in kernel regression and SVMs. The \emph{Recurrent Kalman Network} (RKN) transfers this idea to state estimation and filtering, i.e., we learn a high dimensional deep feature space that allows for efficient inference using the Kalman update equations even for complex systems with high dimensional observations. To achieve this, we assume that the belief state representation can be factorized into independent Gaussian distributions as described in the following sections. 

\subsection{Latent Observation and State Spaces}
\begin{figure*}[t]
	\begin{center}
		 \MyResizeBox{0.65\textwidth}{\tikzMyModel}
	\end{center}
\caption{The \emph{Recurrent Kalman Network}. 
An encoder network extracts latent features $\cvec{w}_t$ from the current observation $\cvec{o}_t$. 
Additionally, it emits an estimate of the uncertainty in the features via the variance $\cvec{\sigma}_t^\mathrm{obs}$.
The transition model $\cmat{A}_t$ is used to predict the current latent prior $\left(\cvec{z}_t^-, \cmat{\Sigma}_t^- \right)$  using the last posterior $\left(\cvec{z}_{t-1}^+, \cmat{\Sigma}_{t-1}^+ \right)$ and subsequently update the prior using the latent observation $(\cvec{w}_t, \cvec{\sigma}_t^\mathrm{obs} )$.
As we use a factorized representation of $\cmat{\Sigma}_t$, the Kalman update simplifies to scalar operations. 
The current latent state $\cvec{z}_t$ consists of the observable units $\cvec{p}_t$ as well as the corresponding memory units $\cvec{m}_t$.
Finally, a decoder produces either $\left(\cvec{s}^+_t, \cvec{\sigma}_t^+ \right)$, a low dimensional observation and an element-wise uncertainty estimate, or $\cvec{o}^+_t$,  a noise free image.}
\label{fig:our}
\end{figure*}

The RKN encoder learns a mapping to a high-dimensional latent observation space $\mathcal{W} =\mathbb{R}^m$. 
The encoder also outputs a vector of uncertainty estimates $\cvec{\sigma}_t^\textrm{obs}$, one for each entry of the latent observation $\cmat{w}_t$.
The latent state space $\mathcal{Z} = \mathbb{R}^n$ of the RKN is related to the observation space $\mathcal{W}$ by the linear latent observation model $\cmat{H} = \left[\begin{array}{cc} \cmat{I}_m & \cmat{0}_{m \times (n-m)} \end{array}\right]$, i.e., $ \cvec{w} = \cmat{H} \cvec{z}$ with  $\cvec{w} \in \mathcal{W}$ and $\cvec{z} \in \mathcal{Z}$. $\cmat{I}_m$ denotes the $m \times m$ identity matrix and $\cmat{0}_{m \times (n-m)}$ denotes a $m \times (n-m)$ matrix filled with zeros. 

The idea behind this choice is to split the latent state vector $\cvec{z}_t$ into two parts, a vector $\cvec{p}_t$ for holding information that can directly be extracted from the observations and a vector $\cvec{m}_t$ to store information inferred over time, e.g., velocities. 
We refer to the former as the observation or upper part and the latter as the memory or lower part of the latent state.
For an ordinary dynamical system and images as observations the former may correspond to positions while the latter corresponds to velocities. Clearly, this choice only makes sense for $m \leq n$ and in this work we assume $n = 2m$, i.e., for each observation unit $p_i$, we also represent a memory unit $m_i$ that stores its velocity information.

We initialize $\cvec{z}_0^+$ with an all zeros vector and $\cmat{\Sigma}_0^+$ with $10 \cdot \cmat{I}$. In practice, it is beneficial to normalize $\cvec{w}_t$ since the statistics of noisy and noise free images differ considerably.

\subsection{The Transition Model}
To obtain a locally linear transition model we learn $K$ constant transition matrices $\cmat{A}^{(k)}$ and combine them using state dependent coefficients $\alpha^{(k)}(\cvec{z_t})$, i.e.,
$$ \cmat{A}_t = \sum_{k=1}^K \alpha^{(k)}(\cvec{z_t}) \cmat{A}^{(k)}. $$
A small neural network with softmax output is used to learn $\alpha^{(k)}$. Similar approaches are used in \citep{fraccaro2017kvae, karl2016dvbf}. 

Using a dense transition matrix in high-dimensional latent spaces
is not feasible as it contains too many parameters and causes numerical instabilities and overfitting, as preliminary experiments showed. 
Therefore, we design each $\cmat{A}^{(k)}$ to consist of four band matrices
$$ \cmat{A}^{(k)} = \left[\begin{array}{cc}\cmat{B}^{(k)}_{11} & \cmat{B}^{(k)}_{12} \\
\cmat{B}^{(k)}_{21} & \cmat{B}^{(k)}_{22} \end{array}\right] $$
with bandwidth $b$. 
Here, $\cmat{B}_{11}^{(k)}$, $\cmat{B}_{12}^{(k)}$, $\cmat{B}_{21}^{(k)}$, $\cmat{B}_{22}^{(k)}$ $\in \mathbb{R}^{m \times m}$ since we assume $n=2m$.
This choice reduces the number of parameters while not affecting performance since the network is free to choose the state representation.

We assume the covariance of the transition noise to be diagonal and denote the vector containing the diagonal values by $\cvec{\sigma}^{\textrm{trans}}$. 
The noise is learned and independent of the state. 
Moreover, it is crucial to correctly initialize the transition matrix. Initially, the transition model should focus on copying the encoder output so that the encoder can learn how to extract good features if observations are available and useful. 
Additionally, it is crucial that $\cmat{A}$ does not yield an instable system. We choose $\cmat{B}^{(k)}_{11} = \cmat{B}^{(k)}_{22} = \cmat{I}$ and $\cmat{B}^{(k)}_{12} =  -\cmat{B}^{(k)}_{21} = 0.2 \cdot \cmat{I}$.

\subsection{Factorized Covariance Representation}
Since the RKN learns high-dimensional representations, we can not work with the full state covariance matrices $\cmat{\Sigma}_t^+$ and $\cmat{\Sigma}_t^-$. 
We can also not fully factorize the state covariances by diagonal matrices as this neglects the correlation between the memory and the observation parts.
As the memory part is excluded from the observation model $\cmat H$, the Kalman update step would not change the memory units nor their uncertainty if we would only use a diagonal covariance matrix $\cmat{\Sigma}_t^+$.
Hence, for each observation unit $p_i$, we compute the covariance with its corresponding memory unit $m_i$. All the other covariances are neglected.
This might be a crude approximation for many systems, however, as our network is free to choose its own state representation it can find a representation where such a factorization works well in practice.
Thus, we use matrices of the form
$$ \cmat{\Sigma}_t = \left[\begin{array}{cc}\cmat{\Sigma}_t^\mathrm{u} & \cmat{\Sigma}_t^\mathrm{s} \\
\cmat{\Sigma}_t^\mathrm{s} &  \cmat{\Sigma}_t^\mathrm{l} \end{array}\right] $$
where each of $\cmat{\Sigma}_t^\mathrm{u}, \cmat{\Sigma}_t^\mathrm{s}, \cmat{\Sigma}_t^\mathrm{l} \in \mathbb{R}^{m \times m}$ is a diagonal matrix. Again, we denote the vectors containing the diagonal values by $\cvec{\sigma}_t^\mathrm{u}, \cmat{\sigma}_t^\mathrm{l}$ and $\cmat{\sigma}_t^\mathrm{s}$. 

\subsection{Factorized Inference in the Latent Space}
Inference in the latent state space can now be implemented, similar to a standard KF, using a prediction and an observation update. 
\paragraph{Prediction Update.}
Equivalently to the classical Kalman Filter, the next prior $\left(\cvec{z}_{t+1}^-,  \cmat{\Sigma}^-_{t+1} \right)$ is obtained from the current posterior $\left(\cvec{z}_{t}^+, \cmat{\Sigma}_{t}^+ \right)$ by
\begin{equation}\cvec{z}^-_{t+1} = \cmat{A}_t \cvec{z}^+_{t} \quad \text{and} \quad \cmat{\Sigma}^-_{t+1} = \cmat{A}_t \cmat{\Sigma}_{t}^+ \cmat{A}_t^T + \cmat{I} \cdot \cvec{\sigma}^\mathrm{trans}.
\end{equation}
However, the special structure of the covariances enables us to significantly simplify the equation for the covariance.
While being straight forward, the full derivations are rather lengthy, thus, we refer to the supplementary material where the equations are given in Eqs. \ref{eq:var1_pred},\ref{eq:var2_pred} and \ref{eq:var3_pred}.

\paragraph{Observation Update.}
Next, the prior is updated using the latent observation $(\cvec{w}_t, \cmat{\sigma}_t^\textrm{obs})$.
Similar to the state, we split the Kalman gain matrix $\cmat{Q}_t$ into an upper $\cmat{Q}_t^\textrm{u}$ and a lower part $\cmat{Q}_t^\textrm{l}$. 
Both  $\cmat{Q}_t^\textrm{u}$ and $\cmat{Q}_t^\textrm{l}$ are squared matrices.
Due to the simple latent observation model $\cmat{H} = \left[\begin{array}{cc} \cmat{I}_m & \cmat{0}_{m \times (n-m)} \end{array}\right]$ and the factorized covariances all off-diagonal entries of $\cmat{Q}_t^\textrm{u}$ and $\cmat{Q}_t^\textrm{l}$ are zero and we can work with vectors representing the diagonals, i.e., $\cvec{q}_t^\textrm{u}$ and $\cvec{q}_t^\textrm{l}$. 
Those are obtained by
\begin{align}
\cvec{q}_t^\mathrm{u} &= \cvec{\sigma}_t^{\mathrm{u},-} \oslash \left( \cvec{\sigma}_t^{\mathrm{u},-} + \cvec{\sigma}_t^\mathrm{obs}  \right) \\ \text{and} \quad \cvec{q}_t^\mathrm{l} &= \cvec{\sigma}_t^{\mathrm{s},-} \oslash \left( \cmat{\sigma}_t^{\mathrm{u},-} + \cmat{\sigma}_t^\mathrm{obs} \right),
\end{align}
where $\oslash$ denotes an elementwise vector division. 
With this the update equation for the mean simplifies to 
\begin{align}
\cvec{z}_t^+ = \cvec{z}_t^- +
\left[\begin{array}{c} \cmat{q}^\mathrm{u}_t \\
\cmat{q}^\mathrm{l}_t \end{array}\right]
\odot
\left[\begin{array}{c}\cvec{w}_t - \cvec{z}^{\mathrm{u},-}_t \\
\cvec{w}_t - \cvec{z}^{\mathrm{u},-}_t  \end{array}\right]
\end{align}
where $\odot$ denotes the elementwise vector product. 
The update equations for the individual covariance parts are given by
\begin{align}
\cvec{\sigma}^{\mathrm{u},+}_t &= \left( \cvec{1}_m - \cvec{q}^\mathrm{u}_t \right) \odot \cvec{\sigma}^{\mathrm{u},-}_t \text{,}\\
\cvec{\sigma}^{\mathrm{s},+}_t &= \left( \cvec{1}_m - \cvec{q}^\mathrm{u}_t \right) \odot \cvec{\sigma}^{\mathrm{s},-}_t \\ \text{and} \quad
\cvec{\sigma}^{\mathrm{l},+}_t &= \cvec{\sigma}^{\mathrm{l}, -}_t - \cvec{q}^\mathrm{l}_t \odot \cvec{\sigma}^{\mathrm{s},-}_t,
\end{align}
where $\cvec{1}_m$ denotes the $m$ dimensional vector consisting of ones.
Again, we refer to the supplementary material for a more detailed derivations.

Besides avoiding the matrix inversion in the Kalman gain computation, the factorization of the covariance matrices reduces the total amount of numbers to store per matrix from $n^2$ to $3 m$.
Additionally, working with $\cvec{\sigma}^\mathrm{s}$ makes it trivial to ensure that the symmetry and positive definiteness of the state covariance are not affected by numerical issues.

\subsection{Loss and Training}

We consider two different potential output distributions, Gaussian distributions for estimating low dimensional observations and Bernoulli distributions for predicting high dimensional observations such as images. 
Both distributions are computed by decoders that use the current latent state estimate $\cvec z^+_t$ as well its uncertainty estimates $\cvec{\sigma}^{\mathrm{u},+}_t,
\cvec{\sigma}^{\mathrm{s},+}_t,
\cvec{\sigma}^{\mathrm{l},+}_t$.

\paragraph{Inferring states.}
Let $\cvec{s}_{1:T}$ be the ground truth sequence with dimension $D_s$, the Gaussian log-likelihood for a single sequence is then computed as  
\begin{align}
&\mathcal{L}\left(\cvec{s}_{(1:T)}\right) = \\&
\dfrac{1}{T} \sum_{t=1}^T  \log \mathcal{N}\left(\cvec{s}_t \bigg| \textrm{dec}_{\mu}(\cvec{z}_t^+), \textrm{dec}_{\Sigma} (\cvec{\sigma}^{\mathrm{u},+}_t,
\cvec{\sigma}^{\mathrm{s},+}_t,
\cvec{\sigma}^{\mathrm{l},+}_t )\right), \nonumber
\end{align}
where $\textrm{dec}_{\mu}(\cdot)$ and $\textrm{dec}_{\Sigma}(\cdot)$ denote the parts of the decoder that are responsible for decoding the latent mean and latent variance respectively.

\paragraph{Inferring images.}
Let $\cvec{o}_{1:T}$ be images with $D_\textrm{o}$ pixels. The Bernoulli log-likelihood for a single sequence is then given by
\begin{align}
\mathcal{L}\left(\cvec{o}_{(1:T)} \right) =& \dfrac{1}{T} \sum_{t=1}^T \sum_{i=0}^{D_\textrm{o}}
o_t^{(i)} \log \left( \textrm{dec}_{o,i} \left(\cvec{z}_t^+ \right) \right)  ... \nonumber \\ ... +&\left(1 - o_t^{(i)}\right) \log \left( 1 - \textrm{dec}_{o,i} \left(\cvec{z}_t^+ \right) \right),
\end{align}
where $o_t^{(i)}$ is the $i$th pixel of the $t$th image. The pixels are in this case represented by grayscale values in the range of $[0;1]$. The $i$th dimension of the decoder is denoted by $\textrm{dec}_{o,i} \left(\cvec{z}_t^+ \right)$, where we use a sigmoid transfer function as output units for the decoder. 

Gradients are computed using (truncated) backpropagation through time (BPTT) \citep{werbos1990bptt} and clipped.   We optimize the objective using the Adam \citep{kingma2014adam} stochastic gradient descent optimizer with default parameters. 

\subsection{The Recurrent Kalman Network}
The prediction and observation updates result in a new type of recurrent neural network, that we call \emph{Recurrent Kalman Network}, which allows working in high dimensional state spaces while keeping numerical stability, computational efficiency and (relatively) low memory consumption. 

Similar to the input gate in LSTMs \citep{hochreiter1997long} and GRUs \citep{cho2014gru} the Kalman gain can be seen as a gate controlling how much the current observation influences the state.
However, this gating explicitly depends on the uncertainty estimates of the latent state and observation and is computed in a principled manner. 
While the sparse transition models and factorization assumptions may seem restrictive, they allow stable and efficient computations in high dimensional spaces.
Since the high dimensional representation is learned jointly with the dynamics, this can yield very powerful models, as shown in our experiments.

In comparison to LSTMs and GRUs the number of parameters is considerably smaller. 
For a fixed bandwidth $b$ and number of basis matrices $k$ it scales linearly in the state size for the RKN while it scales quadratically for LSTMs and GRUs. 

Moreover, the RKN provides a principled method to deal with absent inputs by just omitting the update step and setting the posterior to the prior.

%% file: content/3-EvalAndExp.tex
\section{Evaluation and Experiments}

A full listing of hyperparameters and data set specifications can be found in the supplementary material. 
Code is available online\footnote{https://github.com/LCAS/RKN}.
We compare to LSTM and GRU baselines for which we replaced the RKN transition layer with generic LSTM and GRU layers.
Those were given the encoder output as inputs and have an internal state size of $2 n$. 
The internal state was split into two equally large parts, the first part was used to compute the mean and the second to compute the variance.
We additionally executed most of the following experiments using the root mean squared error to illustrate that our approach is also competitive in prediction accuracy.
The RMSE results can be found in the appendix.

\subsection{Pendulum}
\label{sec:EvalPend}
We evaluated the RKN on a simple simulated pendulum with images of size $24 \times 24$ pixels as observations.
Gaussian transition noise with standard deviation of $\sigma=0.1$ was added to the angular velocity after each step.
In the first experiment, we evaluated filtering in the presence of high observation noise.
We compare against LSTMs and GRUs as these are the only methods that can also perform state estimation. The amount of noise varies between no noise at all and the whole image consisting of pure noise.
Furthermore, the noise is correlated over time, i.e., the model may observe pure noise for several consecutive time steps. 
Details about the noise sampling process can be found in the appendix.
We represent the joint angle $\theta_t$ as a two dimensional vector $\cvec{s}_t = \cvec{\theta}_t = \left(\sin(\theta_t), \cos(\theta_t)\right)^T$ to avoid discontinuities.
We compared different instances of our model to evaluate the effects of assuming sparse transition models and factorized state covariances.
The results are given in \autoref{tab:pend_res}.
The results show that our assumptions do not affect the performance, while we need to learn fewer parameters and can use much more efficient computations using the factorized representation.
Note that for the more complex experiments with a higher dimensional latent state space, we were not able to experiment with full covariance matrices due to lack of memory and massive computation times. 
Moreover, the RKN outperforms LSTM and GRU in all settings.

\begin{table}[]
\caption{Our approach outperforms the generic LSTM and GRU baselines. The GRU with $m=8$ and the LSTM with $m=6$ were designed to have roughly the same amount of parameters as the RKN with $b=3$. In the case where $m=b$ the RKN uses a full transition matrix. {\em fc} stands for full covariance matrix, i.e., we do not use factorization of the belief state.}
\label{tab:pend_res}
\begin{center}
\begin{tabular}{lc}
\toprule
Model& Log Likelihood  \\
\midrule
RKN $(m = 15, b = 3, K = 15)$ & $6.182 \pm 0.155$  \\
RKN $m = b = 15, K = 3)$ & $6.248 \pm 0.1715$\\
RKN $(m = 15, b = 3, K = 15,$ fc $)$ & $6.161 \pm 0.23$ \\
RKN $(m = b = 15, K = 15,$ fc $)$ & $6.197 \pm 0.249$ \\
\midrule
LSTM $m=50$ & $5.773 \pm 0.231$ \\
LSTM $m=6$ & $6.019 \pm 0.122$ \\
\midrule
GRU $m=50$ & $5.649 \pm 0.197$\\
GRU $m=8$ & $6.051 \pm 0.145$ \\
\bottomrule
\end{tabular}
\end{center}
\end{table}

In a second experiment, we evaluated the image prediction performance and compare against existing variational inference approaches.
We randomly removed half of the images from the sequences and tasked the models with imputing those missing frames, i.e., we train the models to predict images instead of the position. We compared our approach to the Kalman Variational Autoencoder (KVAE) \citep{fraccaro2017kvae}, Embed to Control (E2C) \cite{watter2015embed} and Structured Inference Networks (SIN) \cite{krishnan2017sin}. The results can be found in \autoref{tab:pend_imp_res}. Again, our RKN outperforms all other models. This is surprising as the variational inference models use much more complex inference methods and in some cases even more information such as in the KVAE smoothing case.
Sample sequences can be found in the supplementary material. All hyperparameters are the same as for the previous experiment. 

\begin{table}[]
\caption{Comparison on the image imputation task. The informed models were given a mask of booleans indicating which images are valid and which not. The uninformed models were given a black image whenever the image was not valid. E2C and SIN only work in the informed case. Since the KVAE is capable of smoothing in addition to filtering, we evaluated both. Our approach outperforms all models. Only the informed KVAE yields comparable, but still slightly worse results while E2C and SIN fail to capture the dynamics. The uninformed KVAE fails at identifying the invalid images.}
\label{tab:pend_imp_res}
\begin{center}
\begin{tabular}{lc}
\toprule
Model& Log Likelihood \\
\midrule
RKN (informed) &  $- 12.782 \pm 0.0160$ \\
RKN (uninformed) & $- 12.788 \pm 0.0142$ \\
\midrule
KVAE (informed, smooth) & $-13.337 \pm 0.236$\\
KVAE (informed, filter) & $- 14.383 \pm 0.229$ \\  
KVAE (uninformed, filter) & $-46.320 \pm 6.488$\\
KVAE (uninformed, smooth) &  $-38.170 \pm 5.399$\\
\midrule
E2C (informed) & $-95.539 \pm 1.754$ \\
SIN (informed) & $-101.268 \pm 0.567$ \\ 
\bottomrule
\end{tabular}
\end{center}
\end{table}

\subsection{Multiple Pendulums}
Dealing with uncertainty in a principled manner becomes even more important if the observation noise affects different parts of the observation in different ways. In such a scenario the model has to infer which parts are currently observable and which parts are not. 
To obtain a simple experiment with that property, we repeated the pendulum experiments with three colored pendulums in one image. The image was split into four equally sized square parts and the noise was generated individually for each part such that some pendulums may be occluded while others are visible. Exemplary images are shown in the supplementary material. A comparison to LSTM and GRU baselines can be found in \autoref{fig:mult_pendRes} and \autoref{tab:mult_pend_res} as well as an exemplary trajectory in \autoref{fig:mult_pendTraj}. The RKN again clearly outperforms the competing methods. We also computed the quality of the uncertainty prediction by showing the histograms of the normalized prediction errors. While this histogram has clearly a Gaussian shape for the RKN, it looks like a less regular distribution for the LSTMs. 
\begin{table}[t]
\caption{Results of the multiple pendulum experiments.}
\label{tab:mult_pend_res}
\vskip 0.15in
\begin{center}
\begin{tabular}{lc}
\toprule
Model& Log Likelihood  \\
\midrule
RKN $m=45$  & $11.51 \pm 1.703$\\
$b=3, k=15$ &  \\
\midrule  
LSTM $m = 50$  & $7.5224\pm 1.564$  \\
LSTM $m = 12$ & $7.429 \pm 1.307$ \\
\midrule
GRU $m = 50$& $7.541 \pm 1.547$\\
GRU $m = 12$ & $5.602 \pm 1.468$ \\
\bottomrule
\end{tabular}
\end{center}
\end{table}

\begin{figure}[]
\begin{minipage}[t]{0.24\textwidth}
\centering
\MyResizeBox{0.75\textwidth}{\input{img/llrkn_hist.tex}}
\end{minipage}%
\begin{minipage}[t]{0.24\textwidth}
\centering
\MyResizeBox{0.75\textwidth}{\input{img/lstm_hist.tex}}
\end{minipage}%
\caption{To evaluate the quality of our uncertainty prediction we compute the normalized error $\frac{s^{(j)} - s^{(j), +}}{\sigma^{(j),+}}$ for each entry $j$ of $\cvec{s}$ for all time steps in all test sequences. This normalized error should follow a Gaussian distribution with unit variance if the prediction is correct. We compare the resulting error histograms with a unit variance Gaussian. The left histogram shows the RKN, the right one the LSTM. The RKN perfectly fits the normal distribution while the LSTM's normalized error distribution has several modes. Again we designed the smaller LSTM and GRU to have roughly the same amount of parameters as the RKN.}  
\label{fig:mult_pendRes}
\end{figure}
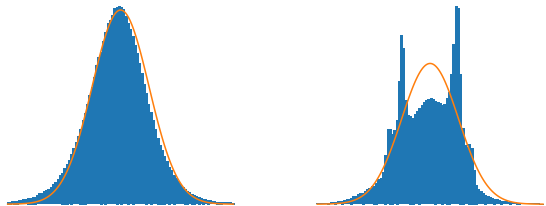
\begin{figure*}[t]
    \centering
    \MyResizeBox{0.85\textwidth}{\input{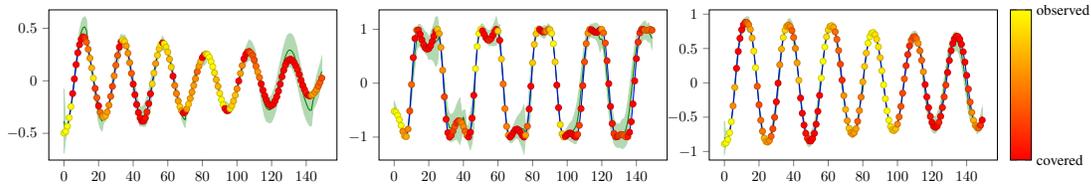}}
    \caption{Predicted sine value of the tree links with 2 times standard deviation (green). Ground truth displayed in blue. The crosses visualize the current visibility of the link with yellow corresponding to fully visible and red to fully occluded. If there is no observation for a considerable time the predictions become less precise due to transition noise, however, the variance also increases.}   \label{fig:mult_pendTraj}
\end{figure*}

\subsection{Quad Link}
We repeated the filtering experiment on a system with much more complicated dynamics, a quad link pendulum on images of size $48 \times 48$ pixels. Since the individual links of the quad link may occlude each other different amounts of noise are induced for each link. Two versions of this experiment were evaluated. One without additional noise and one were we added noise generated by the same process used in the pendulum experiments. You can find the results in \autoref{tab:quad_res}.

Furthermore, we repeated the imputation experiment with the quad link. We compared only to the informed KVAE, since it was the only model producing competitive results for the pendulum. Our approach achieved $-44.470 \pm 0.208$ (informed) and $-44.584 \pm 0.236$ (uninformed). The KVAE achieved $-52.608 \pm 0.602$ for smoothing and $-59.0218 \pm 0.580$ for filtering . Sample images can be found in the appendix.

\begin{table}[t]
\caption{Comparison of our approach with the LSTM and GRU Baselines on the Quad Link Pendulum. Again the RKN performs significantly better than LSTM and GRU who fail to perform well.}
\label{tab:quad_res}
\begin{center}
\begin{tabular}{lcc}
\toprule
Model & without noise & with noise \\
      & Log Likelihood   & Log Likelihood \\
\midrule
RKN $(m =100,$ & $14.534 \pm 0.176$ & $6.259 \pm 0.412$ \\
$b = 3, K = 15)$ & & \\ 
\midrule
LSTM $(m = 50)$ & $ 11.960 \pm  1.24$ &  $ 5.21 \pm 0.305$ \\
LSTM $(m = 100)$  & $7.858 \pm 4.680$    & $3.87  \pm 0.938$ \\
\midrule
GRU $(m=50)$ & $10.346 \pm 2.70$ &  $4.696 \pm 0.699$ \\
GRU $(m=100)$ & $ 5.82 \pm 2.80 $ & $1.2  \pm 1.105$\\
\bottomrule
\end{tabular}
\end{center}
\end{table}

\subsection{KITTI Dataset for Visual Odometry}
\begin{table*}[t]
    \caption{Comparison on the KITTI dataset using the common evaluation scheme. The RKN performs better than the non-recurrent sfmlearner-s baseline as well as the recurrent LSTM and GRU baselines. It performs comparably to DeepVO \cite{wang2017deepvo}, an approach tailored to the task of visual odometry.}
    \vskip 0.15in
    \centering
    \begin{tabular}{lcccccccccc}
    \toprule
 & \multicolumn{2}{c}{Deep VO} & \multicolumn{2}{c}{sfmlearner-s
} & \multicolumn{2}{c}{LSTM} & \multicolumn{2}{c}{GRU} & \multicolumn{2}{c}{RKN} \\
Sequence & $t_\textrm{rel} (\%)$ &  $r_\textrm{rel} (\degree)$ & $t_\textrm{rel} (\%)$ &  $r_\textrm{rel} (\degree)  $ & $t_\textrm{rel} (\%)$ &  $r_\textrm{rel} (\degree)  $ & $t_\textrm{rel} (\%)$ &  $r_\textrm{rel} (\degree)  $ & $t_\textrm{rel} (\%)$ &  $r_\textrm{rel} (\degree)  $ \\ \midrule
 $3$ & $8.49$ & $6.89$ & $13.21$ & $6.70$ & $ 8.99$ & $4.55$ &  $9.34$ & $3.81$ & $ 7.83$ & $3.57$ \\
 $4$ & $7.19$ & $6.97$ & $13.56$ & $4.56$ & $11.88$ & $3.44$ & $12.36$ & $2.89$ & $11.61$ & $2.61$ \\
 $5$ & $2.62$ &	$3.61$ & $13.06$ & $5.70$ &	$ 8.96$ & $3.43$ & $10.02$ & $3.43$ & $ 7.29$ &	$2.77$ \\
 $6$ & $5.42$ &	$5.82$ & $10.87$ & $4.47$ &	$ 9.66$	& $2.80$ & $10.99$ & $3.22$	& $ 8.08$ &	$2.32$ \\
 $7$ & $3.91$ &	$4.60$ & $13.47$ & $8.41$ &	$ 9.83$	& $5.48$ & $13.70$ & $6.52$	& $ 9.38$ &	$4.83$ \\
$10$ & $8.11$ & $8.83$ & $16.69$ & $6.26$ &	$13.85$	& $3.49$ & $13.37$ & $3.25$	& $12.71$ &	$3.09$ \\
\midrule
mean & $5.96$ & $6.12$ & $13.48$ & $6.02$ &	$10.53$ & $3.87$ & $11.63$ & $3.85$ &  $9.48$ & $3.20$ \\
\bottomrule
\end{tabular}
\label{tab:kitti_res}
\end{table*}
Next, we evaluate the RKN on the task of learning visual odometry from monocular images on the KITTI visual odometry dataset \cite{geiger2012kitti}. It consists of 11 labeled image sequences collected by driving a vehicle in an urban environment. 

We use the unsupervised approach proposed by \cite{zhou2017sfmlearner} to extract features from the images. Using these features, we first evaluate a simple forward network, which we refer to as sfmlearner-s. We then augment this architecture with LSTMs, GRUs, and our proposed RKN. Additionally, we compare to DeepVO \cite{wang2017deepvo}, an approach tailored to the problem of visual odometry. In \autoref{tab:kitti_res} we show results using the common KITTI evaluation scheme. The results show that the RKN shows better performance in comparison to LSTMs and GRUs, and even performs comparably to the tailored DeepVO approach.  

\subsection{Pneumatic Brook Robot Arm}
We consider another real world task, learning the dynamics of a pneumatically actuated joint of a Hydro-Lek HLK-7W robot arm. The data consists of measured joint positions and the input current to the controller of the joint sampled at 100Hz. Pneumatic robots are very hard to model as there is a large amount of hysteresis, requiring recurrent prediction models.  

During training, we work with sequences of length $500$. For the first $300$ time steps those sequences consist of the full observation, i.e. joint position and current. The remaining $200$ time steps only the current is given. The models have to impute the missing joint positions in an uninformed fashion, i.e., the absence of a position is only indicated by unrealistically high values. 

In order to evaluate the learned dynamics, we used them to predict future joint positions. After an initial burn-in phase, the model was tasked with predicting the joint position 2 seconds into the future, given only the future control inputs. Afterwards, the next observation was given to the model and the prediction process repeated.
 
\begin{figure}
    \centering
    \includegraphics[width=0.4\textwidth]{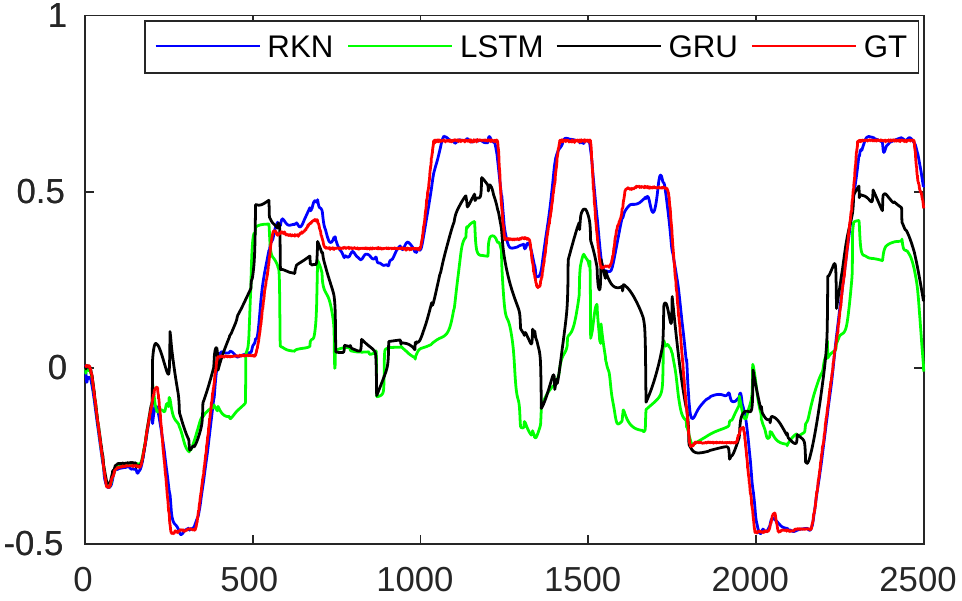}
    \caption{Predicted joint values for the pneumatic joint prediction task. The initial burn-in phase is 200 time steps long, afterwards only the predictions are displayed. GT denotes the ground truth. As can be seen, only our approach manages to give feasible predictions.}
    \label{fig:pneu_joint}
\end{figure}

 Using a $60$ dimensional latent space, $32$ basis matrices and a bandwidth of $3$, the RKN achieved a log likelihood of $4.930$. This is considerably better than the LSTM and GRU baselines, which achieved $0.952$ and $1.186$ respectively. \autoref{fig:pneu_joint} shows exemplary predictions of all approaches.

%% file: img/llrkn_hist.tex
\begin{tikzpicture}[scale=\MyScale]

\definecolor{color0}{rgb}{0.12156862745098,0.466666666666667,0.705882352941177}
\definecolor{color1}{rgb}{1,0.498039215686275,0.0549019607843137}

\begin{axis}[
hide x axis,
hide y axis,
tick align=outside,
tick pos=left,
x grid style={lightgray!92.02614379084967!black},
xmin=-4.4, xmax=4.4,
y grid style={lightgray!92.02614379084967!black},
ymin=0, ymax=0.426677967354563
]
\draw[fill=color0,draw opacity=0] (axis cs:-4,0) rectangle (axis cs:-3.92,0.00447995555884085);
\draw[fill=color0,draw opacity=0] (axis cs:-3.92,0) rectangle (axis cs:-3.84,0.00475995278126841);
\draw[fill=color0,draw opacity=0] (axis cs:-3.84,0) rectangle (axis cs:-3.76,0.00562794417079382);
\draw[fill=color0,draw opacity=0] (axis cs:-3.76,0) rectangle (axis cs:-3.68,0.00618793861564896);
\draw[fill=color0,draw opacity=0] (axis cs:-3.68,0) rectangle (axis cs:-3.6,0.00646793583807648);
\draw[fill=color0,draw opacity=0] (axis cs:-3.6,0) rectangle (axis cs:-3.52,0.0070839297274171);
\draw[fill=color0,draw opacity=0] (axis cs:-3.52,0) rectangle (axis cs:-3.44,0.00761592445002945);
\draw[fill=color0,draw opacity=0] (axis cs:-3.44,0) rectangle (axis cs:-3.36,0.00907191000665273);
\draw[fill=color0,draw opacity=0] (axis cs:-3.36,0) rectangle (axis cs:-3.28,0.00974390334047891);
\draw[fill=color0,draw opacity=0] (axis cs:-3.28,0) rectangle (axis cs:-3.2,0.0127678733426964);
\draw[fill=color0,draw opacity=0] (axis cs:-3.2,0) rectangle (axis cs:-3.12,0.0124038769535406);
\draw[fill=color0,draw opacity=0] (axis cs:-3.12,0) rectangle (axis cs:-3.04,0.01453185584399);
\draw[fill=color0,draw opacity=0] (axis cs:-3.04,0) rectangle (axis cs:-2.96,0.0148958522331458);
\draw[fill=color0,draw opacity=0] (axis cs:-2.96,0) rectangle (axis cs:-2.88,0.0178358230686351);
\draw[fill=color0,draw opacity=0] (axis cs:-2.88,0) rectangle (axis cs:-2.8,0.0214477872379506);
\draw[fill=color0,draw opacity=0] (axis cs:-2.8,0) rectangle (axis cs:-2.72,0.0222317794607477);
\draw[fill=color0,draw opacity=0] (axis cs:-2.72,0) rectangle (axis cs:-2.64,0.0254517475186646);
\draw[fill=color0,draw opacity=0] (axis cs:-2.64,0) rectangle (axis cs:-2.56,0.0285037172431251);
\draw[fill=color0,draw opacity=0] (axis cs:-2.56,0) rectangle (axis cs:-2.48,0.0301277011332047);
\draw[fill=color0,draw opacity=0] (axis cs:-2.48,0) rectangle (axis cs:-2.4,0.0344116586363463);
\draw[fill=color0,draw opacity=0] (axis cs:-2.4,0) rectangle (axis cs:-2.32,0.0403475997518106);
\draw[fill=color0,draw opacity=0] (axis cs:-2.32,0) rectangle (axis cs:-2.24,0.0447715558661658);
\draw[fill=color0,draw opacity=0] (axis cs:-2.24,0) rectangle (axis cs:-2.16,0.050119502814532);
\draw[fill=color0,draw opacity=0] (axis cs:-2.16,0) rectangle (axis cs:-2.08,0.0588274164320289);
\draw[fill=color0,draw opacity=0] (axis cs:-2.08,0) rectangle (axis cs:-2,0.0637553675467539);
\draw[fill=color0,draw opacity=0] (axis cs:-2,0) rectangle (axis cs:-1.92,0.0741712642210589);
\draw[fill=color0,draw opacity=0] (axis cs:-1.92,0) rectangle (axis cs:-1.84,0.0822911836714579);
\draw[fill=color0,draw opacity=0] (axis cs:-1.84,0) rectangle (axis cs:-1.76,0.0898231089547591);
\draw[fill=color0,draw opacity=0] (axis cs:-1.76,0) rectangle (axis cs:-1.68,0.103122977020068);
\draw[fill=color0,draw opacity=0] (axis cs:-1.68,0) rectangle (axis cs:-1.6,0.114378865361656);
\draw[fill=color0,draw opacity=0] (axis cs:-1.6,0) rectangle (axis cs:-1.52,0.126838741759682);
\draw[fill=color0,draw opacity=0] (axis cs:-1.52,0) rectangle (axis cs:-1.44,0.138794623157338);
\draw[fill=color0,draw opacity=0] (axis cs:-1.44,0) rectangle (axis cs:-1.36,0.154194470390854);
\draw[fill=color0,draw opacity=0] (axis cs:-1.36,0) rectangle (axis cs:-1.28,0.172898284849014);
\draw[fill=color0,draw opacity=0] (axis cs:-1.28,0) rectangle (axis cs:-1.2,0.187542139581975);
\draw[fill=color0,draw opacity=0] (axis cs:-1.2,0) rectangle (axis cs:-1.12,0.205153964872669);
\draw[fill=color0,draw opacity=0] (axis cs:-1.12,0) rectangle (axis cs:-1.04,0.2239697782198);
\draw[fill=color0,draw opacity=0] (axis cs:-1.04,0) rectangle (axis cs:-0.96,0.240265616565083);
\draw[fill=color0,draw opacity=0] (axis cs:-0.96,0) rectangle (axis cs:-0.88,0.26095741130248);
\draw[fill=color0,draw opacity=0] (axis cs:-0.88,0) rectangle (axis cs:-0.8,0.285513167709376);
\draw[fill=color0,draw opacity=0] (axis cs:-0.8,0) rectangle (axis cs:-0.72,0.301277011332047);
\draw[fill=color0,draw opacity=0] (axis cs:-0.72,0) rectangle (axis cs:-0.64,0.31267289828485);
\draw[fill=color0,draw opacity=0] (axis cs:-0.64,0) rectangle (axis cs:-0.56,0.334008686633828);
\draw[fill=color0,draw opacity=0] (axis cs:-0.56,0) rectangle (axis cs:-0.48,0.354084487481884);
\draw[fill=color0,draw opacity=0] (axis cs:-0.48,0) rectangle (axis cs:-0.4,0.371360316105664);
\draw[fill=color0,draw opacity=0] (axis cs:-0.4,0) rectangle (axis cs:-0.32,0.377128258887671);
\draw[fill=color0,draw opacity=0] (axis cs:-0.32,0) rectangle (axis cs:-0.24,0.387488156117491);
\draw[fill=color0,draw opacity=0] (axis cs:-0.24,0) rectangle (axis cs:-0.16,0.402216010017183);
\draw[fill=color0,draw opacity=0] (axis cs:-0.16,0) rectangle (axis cs:-0.0800000000000001,0.405295979463883);
\draw[fill=color0,draw opacity=0] (axis cs:-0.0800000000000001,0) rectangle (axis cs:0,0.406359968909108);
\draw[fill=color0,draw opacity=0] (axis cs:0,0) rectangle (axis cs:0.0800000000000001,0.405351978908369);
\draw[fill=color0,draw opacity=0] (axis cs:0.0800000000000001,0) rectangle (axis cs:0.16,0.4005920261271);
\draw[fill=color0,draw opacity=0] (axis cs:0.16,0) rectangle (axis cs:0.24,0.386144169449839);
\draw[fill=color0,draw opacity=0] (axis cs:0.24,0) rectangle (axis cs:0.32,0.372704302773316);
\draw[fill=color0,draw opacity=0] (axis cs:0.32,0) rectangle (axis cs:0.4,0.359460434152493);
\draw[fill=color0,draw opacity=0] (axis cs:0.4,0) rectangle (axis cs:0.48,0.345376573864387);
\draw[fill=color0,draw opacity=0] (axis cs:0.48,0) rectangle (axis cs:0.56,0.325608769961002);
\draw[fill=color0,draw opacity=0] (axis cs:0.56,0) rectangle (axis cs:0.64,0.309032934393294);
\draw[fill=color0,draw opacity=0] (axis cs:0.64,0) rectangle (axis cs:0.72,0.288845134656264);
\draw[fill=color0,draw opacity=0] (axis cs:0.72,0) rectangle (axis cs:0.8,0.26806934075214);
\draw[fill=color0,draw opacity=0] (axis cs:0.8,0) rectangle (axis cs:0.88,0.246089558791577);
\draw[fill=color0,draw opacity=0] (axis cs:0.88,0) rectangle (axis cs:0.96,0.228281735445184);
\draw[fill=color0,draw opacity=0] (axis cs:0.96,0) rectangle (axis cs:1.04,0.205181964594911);
\draw[fill=color0,draw opacity=0] (axis cs:1.04,0) rectangle (axis cs:1.12,0.189110124027569);
\draw[fill=color0,draw opacity=0] (axis cs:1.12,0) rectangle (axis cs:1.2,0.171974294015003);
\draw[fill=color0,draw opacity=0] (axis cs:1.2,0) rectangle (axis cs:1.28,0.153970472612912);
\draw[fill=color0,draw opacity=0] (axis cs:1.28,0) rectangle (axis cs:1.36,0.136190648988762);
\draw[fill=color0,draw opacity=0] (axis cs:1.36,0) rectangle (axis cs:1.44,0.121854791200471);
\draw[fill=color0,draw opacity=0] (axis cs:1.44,0) rectangle (axis cs:1.52,0.111634892581865);
\draw[fill=color0,draw opacity=0] (axis cs:1.52,0) rectangle (axis cs:1.6,0.0988950189614118);
\draw[fill=color0,draw opacity=0] (axis cs:1.6,0) rectangle (axis cs:1.68,0.0876391306198251);
\draw[fill=color0,draw opacity=0] (axis cs:1.68,0) rectangle (axis cs:1.76,0.0765232408894503);
\draw[fill=color0,draw opacity=0] (axis cs:1.76,0) rectangle (axis cs:1.84,0.069467310884276);
\draw[fill=color0,draw opacity=0] (axis cs:1.84,0) rectangle (axis cs:1.92,0.0601434033774384);
\draw[fill=color0,draw opacity=0] (axis cs:1.92,0) rectangle (axis cs:2,0.0523314808717097);
\draw[fill=color0,draw opacity=0] (axis cs:2,0) rectangle (axis cs:2.08,0.0470395333678289);
\draw[fill=color0,draw opacity=0] (axis cs:2.08,0) rectangle (axis cs:2.16,0.040963593641151);
\draw[fill=color0,draw opacity=0] (axis cs:2.16,0) rectangle (axis cs:2.24,0.0343556591918608);
\draw[fill=color0,draw opacity=0] (axis cs:2.24,0) rectangle (axis cs:2.32,0.0311356911339439);
\draw[fill=color0,draw opacity=0] (axis cs:2.32,0) rectangle (axis cs:2.4,0.0268237339085596);
\draw[fill=color0,draw opacity=0] (axis cs:2.4,0) rectangle (axis cs:2.48,0.023799763906342);
\draw[fill=color0,draw opacity=0] (axis cs:2.48,0) rectangle (axis cs:2.56,0.021279788904494);
\draw[fill=color0,draw opacity=0] (axis cs:2.56,0) rectangle (axis cs:2.64,0.0186758147359178);
\draw[fill=color0,draw opacity=0] (axis cs:2.64,0) rectangle (axis cs:2.72,0.0160998402895845);
\draw[fill=color0,draw opacity=0] (axis cs:2.72,0) rectangle (axis cs:2.8,0.0150358508443596);
\draw[fill=color0,draw opacity=0] (axis cs:2.8,0) rectangle (axis cs:2.88,0.0122638783423268);
\draw[fill=color0,draw opacity=0] (axis cs:2.88,0) rectangle (axis cs:2.96,0.0108638922301891);
\draw[fill=color0,draw opacity=0] (axis cs:2.96,0) rectangle (axis cs:3.04,0.0099679011184209);
\draw[fill=color0,draw opacity=0] (axis cs:3.04,0) rectangle (axis cs:3.12,0.00837191695058384);
\draw[fill=color0,draw opacity=0] (axis cs:3.12,0) rectangle (axis cs:3.2,0.00744792611657292);
\draw[fill=color0,draw opacity=0] (axis cs:3.2,0) rectangle (axis cs:3.28,0.00568394361527933);
\draw[fill=color0,draw opacity=0] (axis cs:3.28,0) rectangle (axis cs:3.36,0.00587994167097862);
\draw[fill=color0,draw opacity=0] (axis cs:3.36,0) rectangle (axis cs:3.44,0.0047039533367829);
\draw[fill=color0,draw opacity=0] (axis cs:3.44,0) rectangle (axis cs:3.52,0.00459195444781187);
\draw[fill=color0,draw opacity=0] (axis cs:3.52,0) rectangle (axis cs:3.6,0.0041999583364133);
\draw[fill=color0,draw opacity=0] (axis cs:3.6,0) rectangle (axis cs:3.68,0.00363996389155823);
\draw[fill=color0,draw opacity=0] (axis cs:3.68,0) rectangle (axis cs:3.76,0.00296797055773206);
\draw[fill=color0,draw opacity=0] (axis cs:3.76,0) rectangle (axis cs:3.84,0.00293997083548931);
\draw[fill=color0,draw opacity=0] (axis cs:3.84,0) rectangle (axis cs:3.92,0.00291197111324655);
\draw[fill=color0,draw opacity=0] (axis cs:3.92,0) rectangle (axis cs:4,0.00260397416857625);
\addplot [very thick, color1, forget plot]
table [row sep=\\]{%
-4	0.000133830225764885 \\
-3.99199199199199	0.000138182046410498 \\
-3.98398398398398	0.000142666228042307 \\
-3.97597597597598	0.000147286481503254 \\
-3.96796796796797	0.000152046611333821 \\
-3.95995995995996	0.000156950517814782 \\
-3.95195195195195	0.00016200219904485 \\
-3.94394394394394	0.00016720575305353 \\
-3.93593593593594	0.000172565379949496 \\
-3.92792792792793	0.00017808538410479 \\
-3.91991991991992	0.000183770176375138 \\
-3.91191191191191	0.000189624276356684 \\
-3.9039039039039	0.000195652314679408 \\
-3.8958958958959	0.000201859035337517 \\
-3.88788788788789	0.000208249298057069 \\
-3.87987987987988	0.000214828080701088 \\
-3.87187187187187	0.000221600481712426 \\
-3.86386386386386	0.0002285717225946 \\
-3.85585585585586	0.000235747150430845 \\
-3.84784784784785	0.000243132240441603 \\
-3.83983983983984	0.000250732598580649 \\
-3.83183183183183	0.000258553964170059 \\
-3.82382382382382	0.000266602212574213 \\
-3.81581581581582	0.000274883357912987 \\
-3.80780780780781	0.000283403555814325 \\
-3.7997997997998	0.000292169106206322 \\
-3.79179179179179	0.000301186456148957 \\
-3.78378378378378	0.0003104622027056 \\
-3.77577577577578	0.000320003095854415 \\
-3.76776776776777	0.000329816041439723 \\
-3.75975975975976	0.000339908104163431 \\
-3.75175175175175	0.00035028651061658 \\
-3.74374374374374	0.000360958652351047 \\
-3.73573573573574	0.000371932088991451 \\
-3.72772772772773	0.000383214551387261 \\
-3.71971971971972	0.000394813944805093 \\
-3.71171171171171	0.000406738352161197 \\
-3.7037037037037	0.000418996037294058 \\
-3.6956956956957	0.00043159544827708 \\
-3.68768768768769	0.00044454522077123 \\
-3.67967967967968	0.000457854181417586 \\
-3.67167167167167	0.000471531351269624 \\
-3.66366366366366	0.000485585949265103 \\
-3.65565565565566	0.000500027395737409 \\
-3.64764764764765	0.000514865315966112 \\
-3.63963963963964	0.000530109543766577 \\
-3.63163163163163	0.000545770125118336 \\
-3.62362362362362	0.000561857321831991 \\
-3.61561561561562	0.00057838161525435 \\
-3.60760760760761	0.000595353710011453 \\
-3.5995995995996	0.000612784537789192 \\
-3.59159159159159	0.000630685261151105 \\
-3.58358358358358	0.000649067277392973 \\
-3.57557557557558	0.000667942222433796 \\
-3.56756756756757	0.000687321974742665 \\
-3.55955955955956	0.000707218659301081 \\
-3.55155155155155	0.000727644651600166 \\
-3.54354354354354	0.000748612581672245 \\
-3.53553553553554	0.000770135338156225 \\
-3.52752752752753	0.000792226072396121 \\
-3.51951951951952	0.00081489820257214 \\
-3.51151151151151	0.000838165417863603 \\
-3.5035035035035	0.000862041682643007 \\
-3.4954954954955	0.000886541240700502 \\
-3.48748748748749	0.00091167861949796 \\
-3.47947947947948	0.000937468634451866 \\
-3.47147147147147	0.000963926393244149 \\
-3.46346346346346	0.000991067300160059 \\
-3.45545545545546	0.0010189070604522 \\
-3.44744744744745	0.00104746168472971 \\
-3.43943943943944	0.0010767474933716 \\
-3.43143143143143	0.00110678112096323 \\
-3.42342342342342	0.00113757952075483 \\
-3.41541541541542	0.00116915996914087 \\
-3.40740740740741	0.00120154007015926 \\
-3.3993993993994	0.00123473776000907 \\
-3.39139139139139	0.00126877131158549 \\
-3.38338338338338	0.00130365933903088 \\
-3.37537537537538	0.00133942080230046 \\
-3.36736736736737	0.00137607501174126 \\
-3.35935935935936	0.001413641632683 \\
-3.35135135135135	0.00145214069003938 \\
-3.34334334334334	0.0014915925729182 \\
-3.33533533533534	0.00153201803923895 \\
-3.32732732732733	0.00157343822035606 \\
-3.31931931931932	0.00161587462568624 \\
-3.31131131131131	0.00165934914733831 \\
-3.3033033033033	0.00170388406474362 \\
-3.2952952952953	0.00174950204928538 \\
-3.28728728728729	0.00179622616892505 \\
-3.27927927927928	0.00184407989282385 \\
-3.27127127127127	0.00189308709595757 \\
-3.26326326326326	0.00194327206372255 \\
-3.25525525525526	0.00199465949653091 \\
-3.24724724724725	0.00204727451439301 \\
-3.23923923923924	0.00210114266148476 \\
-3.23123123123123	0.00215628991069791 \\
-3.22322322322322	0.00221274266817093 \\
-3.21521521521522	0.00227052777779815 \\
-3.20720720720721	0.00232967252571501 \\
-3.1991991991992	0.00239020464475689 \\
-3.19119119119119	0.00245215231888918 \\
-3.18318318318318	0.00251554418760605 \\
-3.17517517517518	0.00258040935029545 \\
-3.16716716716717	0.00264677737056774 \\
-3.15915915915916	0.00271467828054534 \\
-3.15115115115115	0.00278414258511061 \\
-3.14314314314314	0.00285520126610952 \\
-3.13513513513514	0.00292788578650791 \\
-3.12712712712713	0.00300222809449793 \\
-3.11911911911912	0.00307826062755151 \\
-3.11111111111111	0.00315601631641804 \\
-3.1031031031031	0.00323552858906328 \\
-3.0950950950951	0.00331683137454643 \\
-3.08708708708709	0.00339995910683238 \\
-3.07907907907908	0.00348494672853595 \\
-3.07107107107107	0.00357182969459489 \\
-3.06306306306306	0.00366064397586864 \\
-3.05505505505506	0.00375142606265932 \\
-3.04704704704705	0.00384421296815189 \\
-3.03903903903904	0.00393904223176991 \\
-3.03103103103103	0.00403595192244368 \\
-3.02302302302302	0.0041349806417872 \\
-3.01501501501502	0.00423616752718052 \\
-3.00700700700701	0.00433955225475399 \\
-2.998998998999	0.00444517504227068 \\
-2.99099099099099	0.00455307665190355 \\
-2.98298298298298	0.00466329839290368 \\
-2.97497497497497	0.00477588212415564 \\
-2.96696696696697	0.00489087025661667 \\
-2.95895895895896	0.00500830575563558 \\
-2.95095095095095	0.00512823214314763 \\
-2.94294294294294	0.00525069349974172 \\
-2.93493493493493	0.00537573446659573 \\
-2.92692692692693	0.00550340024727644 \\
-2.91891891891892	0.00563373660939975 \\
-2.91091091091091	0.0057667898861475 \\
-2.9029029029029	0.00590260697763674 \\
-2.89489489489489	0.00604123535213746 \\
-2.88688688688689	0.00618272304713484 \\
-2.87887887887888	0.00632711867023164 \\
-2.87087087087087	0.00647447139988703 \\
-2.86286286286286	0.00662483098598741 \\
-2.85485485485485	0.0067782477502453 \\
-2.84684684684685	0.0069347725864219 \\
-2.83883883883884	0.00709445696036946 \\
-2.83083083083083	0.00725735290988893 \\
-2.82282282282282	0.00742351304439893 \\
-2.81481481481481	0.00759299054441176 \\
-2.80680680680681	0.0077658391608121 \\
-2.7987987987988	0.00794211321393437 \\
-2.79079079079079	0.00812186759243432 \\
-2.78278278278278	0.00830515775195071 \\
-2.77477477477477	0.00849203971355293 \\
-2.76676676676677	0.00868257006197001 \\
-2.75875875875876	0.00887680594359712 \\
-2.75075075075075	0.00907480506427511 \\
-2.74274274274274	0.00927662568683898 \\
-2.73473473473473	0.00948232662843099 \\
-2.72672672672673	0.00969196725757422 \\
-2.71871871871872	0.00990560749100246 \\
-2.71071071071071	0.0101233077902421 \\
-2.7027027027027	0.0103451291579421 \\
-2.69469469469469	0.0105711331339476 \\
-2.68668668668669	0.0108013817911136 \\
-2.67867867867868	0.011035937730854 \\
-2.67067067067067	0.0112748640784219 \\
-2.66266266266266	0.0115182244779185 \\
-2.65465465465465	0.0117660830870247 \\
-2.64664664664665	0.0120185045714526 \\
-2.63863863863864	0.0122755540991133 \\
-2.63063063063063	0.0125372973339956 \\
-2.62262262262262	0.0128038004297541 \\
-2.61461461461461	0.0130751300230007 \\
-2.60660660660661	0.0133513532262973 \\
-2.5985985985986	0.0136325376208454 \\
-2.59059059059059	0.0139187512488692 \\
-2.58258258258258	0.014210062605689 \\
-2.57457457457457	0.0145065406314803 \\
-2.56656656656657	0.0148082547027168 \\
-2.55855855855856	0.0151152746232934 \\
-2.55055055055055	0.0154276706153247 \\
-2.54254254254254	0.0157455133096184 \\
-2.53453453453453	0.0160688737358178 \\
-2.52652652652653	0.016397823312213 \\
-2.51851851851852	0.0167324338352151 \\
-2.51051051051051	0.0170727774684938 \\
-2.5025025025025	0.0174189267317727 \\
-2.49449449449449	0.0177709544892815 \\
-2.48648648648649	0.0181289339378627 \\
-2.47847847847848	0.0184929385947284 \\
-2.47047047047047	0.0188630422848682 \\
-2.46246246246246	0.0192393191281025 \\
-2.45445445445445	0.0196218435257817 \\
-2.44644644644645	0.0200106901471285 \\
-2.43843843843844	0.020405933915221 \\
-2.43043043043043	0.0208076499926159 \\
-2.42242242242242	0.0212159137666093 \\
-2.41441441441441	0.0216308008341344 \\
-2.40640640640641	0.0220523869862944 \\
-2.3983983983984	0.0224807481925294 \\
-2.39039039039039	0.022915960584417 \\
-2.38238238238238	0.0233581004391047 \\
-2.37437437437437	0.023807244162374 \\
-2.36636636636637	0.0242634682713357 \\
-2.35835835835836	0.0247268493767558 \\
-2.35035035035035	0.0251974641650118 \\
-2.34234234234234	0.0256753893796796 \\
-2.33433433433433	0.0261607018027501 \\
-2.32632632632633	0.0266534782354775 \\
-2.31831831831832	0.0271537954788581 \\
-2.31031031031031	0.0276617303137407 \\
-2.3023023023023	0.0281773594805703 \\
-2.29429429429429	0.0287007596587648 \\
-2.28628628628629	0.0292320074457259 \\
-2.27827827827828	0.029771179335487 \\
-2.27027027027027	0.0303183516969979 \\
-2.26226226226226	0.0308736007520491 \\
-2.25425425425425	0.0314370025528369 \\
-2.24624624624625	0.0320086329591723 \\
-2.23823823823824	0.0325885676153351 \\
-2.23023023023023	0.0331768819265761 \\
-2.22222222222222	0.0337736510352706 \\
-2.21421421421421	0.0343789497967251 \\
-2.20620620620621	0.0349928527546413 \\
-2.1981981981982	0.0356154341162401 \\
-2.19019019019019	0.0362467677270497 \\
-2.18218218218218	0.0368869270453615 \\
-2.17417417417417	0.0375359851163567 \\
-2.16616616616617	0.03819401454591 \\
-2.15815815815816	0.0388610874740724 \\
-2.15015015015015	0.03953727554824 \\
-2.14214214214214	0.0402226498960119 \\
-2.13413413413413	0.0409172810977437 \\
-2.12612612612613	0.0416212391588003 \\
-2.11811811811812	0.0423345934815158 \\
-2.11011011011011	0.0430574128368641 \\
-2.1021021021021	0.043789765335848 \\
-2.09409409409409	0.0445317184006117 \\
-2.08608608608609	0.0452833387352837 \\
-2.07807807807808	0.0460446922965577 \\
-2.07007007007007	0.0468158442640166 \\
-2.06206206206206	0.0475968590102091 \\
-2.05405405405405	0.0483878000704834 \\
-2.04604604604605	0.0491887301125894 \\
-2.03803803803804	0.0499997109060533 \\
-2.03003003003003	0.0508208032913361 \\
-2.02202202202202	0.0516520671487823 \\
-2.01401401401401	0.0524935613673682 \\
-2.00600600600601	0.053345343813258 \\
-1.997997997998	0.0542074712981785 \\
-1.98998998998999	0.0550799995476186 \\
-1.98198198198198	0.0559629831688668 \\
-1.97397397397397	0.0568564756188931 \\
-1.96596596596597	0.0577605291720876 \\
-1.95795795795796	0.0586751948878651 \\
-1.94994994994995	0.0596005225781457 \\
-1.94194194194194	0.0605365607747232 \\
-1.93393393393393	0.0614833566965309 \\
-1.92592592592593	0.062440956216817 \\
-1.91791791791792	0.06340940383024 \\
-1.90990990990991	0.0643887426198963 \\
-1.9019019019019	0.0653790142242912 \\
-1.89389389389389	0.0663802588042656 \\
-1.88588588588589	0.0673925150098903 \\
-1.87787787787788	0.0684158199473405 \\
-1.86986986986987	0.0694502091457625 \\
-1.86186186186186	0.0704957165241465 \\
-1.85385385385385	0.0715523743582165 \\
-1.84584584584585	0.0726202132473521 \\
-1.83783783783784	0.0736992620815557 \\
-1.82982982982983	0.0747895480084762 \\
-1.82182182182182	0.0758910964005057 \\
-1.81381381381381	0.0770039308219614 \\
-1.80580580580581	0.0781280729963669 \\
-1.7977977977978	0.0792635427738473 \\
-1.78978978978979	0.0804103580986525 \\
-1.78178178178178	0.0815685349768227 \\
-1.77377377377377	0.0827380874440109 \\
-1.76576576576577	0.0839190275334777 \\
-1.75775775775776	0.0851113652442723 \\
-1.74974974974975	0.086315108509615 \\
-1.74174174174174	0.0875302631654975 \\
-1.73373373373373	0.0887568329195139 \\
-1.72572572572573	0.0899948193199405 \\
-1.71771771771772	0.0912442217250772 \\
-1.70970970970971	0.0925050372728685 \\
-1.7017017017017	0.0937772608508177 \\
-1.69369369369369	0.0950608850662112 \\
-1.68568568568569	0.0963559002166691 \\
-1.67767767767768	0.0976622942610365 \\
-1.66966966966967	0.0989800527906321 \\
-1.66166166166166	0.100309159000872 \\
-1.65365365365365	0.10164959366328 \\
-1.64564564564565	0.103001335097907 \\
-1.63763763763764	0.10436435914617 \\
-1.62962962962963	0.105738639144131 \\
-1.62162162162162	0.107124145896225 \\
-1.61361361361361	0.108520847649465 \\
-1.60560560560561	0.10992871006813 \\
-1.5975975975976	0.111347696208955 \\
-1.58958958958959	0.112777766496842 \\
-1.58158158158158	0.114218878701105 \\
-1.57357357357357	0.115670987912262 \\
-1.56556556556557	0.117134046519405 \\
-1.55755755755756	0.11860800418814 \\
-1.54954954954955	0.120092807839133 \\
-1.54154154154154	0.121588401627275 \\
-1.53353353353353	0.123094726921466 \\
-1.52552552552553	0.124611722285057 \\
-1.51751751751752	0.12613932345695 \\
-1.50950950950951	0.127677463333369 \\
-1.5015015015015	0.129226071950339 \\
-1.49349349349349	0.130785076466857 \\
-1.48548548548549	0.132354401148798 \\
-1.47747747747748	0.133933967353555 \\
-1.46946946946947	0.13552369351543 \\
-1.46146146146146	0.137123495131801 \\
-1.45345345345345	0.13873328475006 \\
-1.44544544544545	0.140352971955359 \\
-1.43743743743744	0.141982463359159 \\
-1.42942942942943	0.143621662588612 \\
-1.42142142142142	0.14527047027677 \\
-1.41341341341341	0.146928784053658 \\
-1.40540540540541	0.148596498538208 \\
-1.3973973973974	0.150273505331067 \\
-1.38938938938939	0.151959693008306 \\
-1.38138138138138	0.153654947116025 \\
-1.37337337337337	0.155359150165875 \\
-1.36536536536537	0.157072181631514 \\
-1.35735735735736	0.158793917945997 \\
-1.34934934934935	0.160524232500121 \\
-1.34134134134134	0.16226299564173 \\
-1.33333333333333	0.164010074675994 \\
-1.32532532532533	0.165765333866673 \\
-1.31731731731732	0.167528634438376 \\
-1.30930930930931	0.169299834579821 \\
-1.3013013013013	0.17107878944811 \\
-1.29329329329329	0.172865351174024 \\
-1.28528528528529	0.174659368868355 \\
-1.27727727727728	0.176460688629273 \\
-1.26926926926927	0.178269153550739 \\
-1.26126126126126	0.180084603731981 \\
-1.25325325325325	0.181906876288021 \\
-1.24524524524525	0.183735805361283 \\
-1.23723723723724	0.185571222134271 \\
-1.22922922922923	0.187412954843324 \\
-1.22122122122122	0.18926082879347 \\
-1.21321321321321	0.191114666374361 \\
-1.20520520520521	0.192974287077311 \\
-1.1971971971972	0.194839507513435 \\
-1.18918918918919	0.19671014143289 \\
-1.18118118118118	0.198585999745227 \\
-1.17317317317317	0.200466890540856 \\
-1.16516516516517	0.202352619113618 \\
-1.15715715715716	0.204242987984477 \\
-1.14914914914915	0.206137796926331 \\
-1.14114114114114	0.208036842989938 \\
-1.13313313313313	0.209939920530956 \\
-1.12512512512513	0.21184682123811 \\
-1.11711711711712	0.21375733416247 \\
-1.10910910910911	0.215671245747847 \\
-1.1011011011011	0.217588339862305 \\
-1.09309309309309	0.219508397830784 \\
-1.08508508508509	0.221431198468836 \\
-1.07707707707708	0.223356518117463 \\
-1.06906906906907	0.225284130679063 \\
-1.06106106106106	0.22721380765447 \\
-1.05305305305305	0.229145318181096 \\
-1.04504504504505	0.231078429072154 \\
-1.03703703703704	0.233012904856966 \\
-1.02902902902903	0.234948507822352 \\
-1.02102102102102	0.236884998055083 \\
-1.01301301301301	0.238822133485401 \\
-1.00500500500501	0.24075966993159 \\
-0.996996996996997	0.242697361145593 \\
-0.988988988988989	0.244634958859672 \\
-0.980980980980981	0.246572212834085 \\
-0.972972972972973	0.248508870905785 \\
-0.964964964964965	0.250444679038131 \\
-0.956956956956957	0.252379381371581 \\
-0.948948948948949	0.254312720275385 \\
-0.940940940940941	0.256244436400228 \\
-0.932932932932933	0.258174268731856 \\
-0.924924924924925	0.260101954645624 \\
-0.916916916916917	0.262027229961987 \\
-0.908908908908909	0.263949829002911 \\
-0.900900900900901	0.265869484649172 \\
-0.892892892892893	0.267785928398552 \\
-0.884884884884885	0.269698890424904 \\
-0.876876876876877	0.271608099638066 \\
-0.868868868868869	0.273513283744617 \\
-0.860860860860861	0.275414169309444 \\
-0.852852852852853	0.277310481818125 \\
-0.844844844844845	0.279201945740075 \\
-0.836836836836837	0.281088284592474 \\
-0.828828828828829	0.28296922100493 \\
-0.820820820820821	0.284844476784867 \\
-0.812812812812813	0.286713772983616 \\
-0.804804804804805	0.288576829963198 \\
-0.796796796796797	0.290433367463758 \\
-0.788788788788789	0.292283104671645 \\
-0.780780780780781	0.294125760288111 \\
-0.772772772772773	0.295961052598604 \\
-0.764764764764765	0.29778869954263 \\
-0.756756756756757	0.299608418784177 \\
-0.748748748748749	0.301419927782648 \\
-0.740740740740741	0.303222943864308 \\
-0.732732732732733	0.305017184294205 \\
-0.724724724724725	0.306802366348536 \\
-0.716716716716717	0.308578207387456 \\
-0.708708708708709	0.310344424928268 \\
-0.700700700700701	0.312100736719007 \\
-0.692692692692693	0.313846860812361 \\
-0.684684684684685	0.31558251563992 \\
-0.676676676676677	0.317307420086723 \\
-0.668668668668669	0.319021293566066 \\
-0.660660660660661	0.320723856094559 \\
-0.652652652652653	0.322414828367396 \\
-0.644644644644645	0.324093931833804 \\
-0.636636636636636	0.325760888772662 \\
-0.628628628628629	0.327415422368237 \\
-0.620620620620621	0.329057256786028 \\
-0.612612612612613	0.330686117248681 \\
-0.604604604604605	0.33230173011195 \\
-0.596596596596596	0.333903822940662 \\
-0.588588588588589	0.335492124584683 \\
-0.580580580580581	0.337066365254824 \\
-0.572572572572573	0.338626276598684 \\
-0.564564564564565	0.340171591776383 \\
-0.556556556556556	0.341702045536168 \\
-0.548548548548549	0.343217374289848 \\
-0.54054054054054	0.344717316188045 \\
-0.532532532532533	0.346201611195215 \\
-0.524524524524525	0.347670001164422 \\
-0.516516516516516	0.349122229911825 \\
-0.508508508508509	0.350558043290856 \\
-0.5005005005005	0.351977189266054 \\
-0.492492492492492	0.353379417986526 \\
-0.484484484484485	0.354764481859011 \\
-0.476476476476476	0.356132135620502 \\
-0.468468468468469	0.357482136410418 \\
-0.46046046046046	0.358814243842273 \\
-0.452452452452452	0.360128220074839 \\
-0.444444444444445	0.361423829882744 \\
-0.436436436436436	0.362700840726495 \\
-0.428428428428429	0.363959022821904 \\
-0.42042042042042	0.365198149208855 \\
-0.412412412412412	0.366417995819425 \\
-0.404404404404405	0.3676183415453 \\
-0.396396396396396	0.368798968304465 \\
-0.388388388388389	0.369959661107154 \\
-0.38038038038038	0.37110020812101 \\
-0.372372372372372	0.372220400735444 \\
-0.364364364364365	0.373320033625156 \\
-0.356356356356356	0.374398904812801 \\
-0.348348348348348	0.375456815730761 \\
-0.34034034034034	0.376493571282005 \\
-0.332332332332332	0.377508979900006 \\
-0.324324324324325	0.378502853607702 \\
-0.316316316316316	0.379475008075448 \\
-0.308308308308308	0.380425262677969 \\
-0.3003003003003	0.38135344055026 \\
-0.292292292292292	0.382259368642425 \\
-0.284284284284285	0.38314287777343 \\
-0.276276276276276	0.384003802683735 \\
-0.268268268268268	0.3848419820868 \\
-0.26026026026026	0.385657258719425 \\
-0.252252252252252	0.386449479390916 \\
-0.244244244244244	0.387218495031049 \\
-0.236236236236236	0.387964160736805 \\
-0.228228228228228	0.38868633581787 \\
-0.22022022022022	0.389384883840873 \\
-0.212212212212212	0.390059672672329 \\
-0.204204204204204	0.390710574520297 \\
-0.196196196196196	0.391337465974705 \\
-0.188188188188188	0.39194022804635 \\
-0.18018018018018	0.392518746204527 \\
-0.172172172172172	0.393072910413305 \\
-0.164164164164164	0.393602615166401 \\
-0.156156156156156	0.394107759520658 \\
-0.148148148148148	0.394588247128102 \\
-0.14014014014014	0.395043986266573 \\
-0.132132132132132	0.3954748898689 \\
-0.124124124124124	0.395880875550626 \\
-0.116116116116116	0.396261865636259 \\
-0.108108108108108	0.396617787184037 \\
-0.1001001001001	0.396948572009207 \\
-0.0920920920920922	0.397254156705792 \\
-0.084084084084084	0.397534482666846 \\
-0.0760760760760761	0.39778949610319 \\
-0.0680680680680683	0.39801914806061 \\
-0.06006006006006	0.398223394435522 \\
-0.0520520520520522	0.398402195989086 \\
-0.0440440440440439	0.398555518359772 \\
-0.0360360360360361	0.398683332074364 \\
-0.0280280280280278	0.398785612557403 \\
-0.02002002002002	0.398862340139061 \\
-0.0120120120120122	0.398913500061449 \\
-0.00400400400400391	0.398939082483343 \\
0.00400400400400436	0.398939082483343 \\
0.0120120120120122	0.398913500061449 \\
0.02002002002002	0.398862340139061 \\
0.0280280280280278	0.398785612557403 \\
0.0360360360360357	0.398683332074364 \\
0.0440440440440444	0.398555518359772 \\
0.0520520520520522	0.398402195989086 \\
0.06006006006006	0.398223394435522 \\
0.0680680680680679	0.39801914806061 \\
0.0760760760760757	0.39778949610319 \\
0.0840840840840844	0.397534482666846 \\
0.0920920920920922	0.397254156705792 \\
0.1001001001001	0.396948572009207 \\
0.108108108108108	0.396617787184037 \\
0.116116116116116	0.396261865636259 \\
0.124124124124124	0.395880875550626 \\
0.132132132132132	0.3954748898689 \\
0.14014014014014	0.395043986266573 \\
0.148148148148148	0.394588247128102 \\
0.156156156156156	0.394107759520658 \\
0.164164164164164	0.393602615166401 \\
0.172172172172172	0.393072910413305 \\
0.18018018018018	0.392518746204527 \\
0.188188188188188	0.39194022804635 \\
0.196196196196196	0.391337465974705 \\
0.204204204204204	0.390710574520297 \\
0.212212212212212	0.390059672672329 \\
0.22022022022022	0.389384883840873 \\
0.228228228228228	0.388686335817871 \\
0.236236236236236	0.387964160736805 \\
0.244244244244245	0.387218495031049 \\
0.252252252252252	0.386449479390916 \\
0.26026026026026	0.385657258719425 \\
0.268268268268268	0.3848419820868 \\
0.276276276276276	0.384003802683735 \\
0.284284284284285	0.38314287777343 \\
0.292292292292292	0.382259368642425 \\
0.3003003003003	0.38135344055026 \\
0.308308308308308	0.380425262677969 \\
0.316316316316316	0.379475008075448 \\
0.324324324324325	0.378502853607702 \\
0.332332332332332	0.377508979900006 \\
0.34034034034034	0.376493571282005 \\
0.348348348348348	0.375456815730762 \\
0.356356356356356	0.374398904812801 \\
0.364364364364365	0.373320033625156 \\
0.372372372372372	0.372220400735444 \\
0.38038038038038	0.37110020812101 \\
0.388388388388388	0.369959661107154 \\
0.396396396396397	0.368798968304465 \\
0.404404404404405	0.3676183415453 \\
0.412412412412412	0.366417995819425 \\
0.42042042042042	0.365198149208855 \\
0.428428428428428	0.363959022821904 \\
0.436436436436437	0.362700840726495 \\
0.444444444444445	0.361423829882744 \\
0.452452452452452	0.360128220074839 \\
0.46046046046046	0.358814243842273 \\
0.468468468468468	0.357482136410418 \\
0.476476476476477	0.356132135620502 \\
0.484484484484485	0.354764481859011 \\
0.492492492492492	0.353379417986526 \\
0.5005005005005	0.351977189266054 \\
0.508508508508508	0.350558043290856 \\
0.516516516516517	0.349122229911825 \\
0.524524524524525	0.347670001164422 \\
0.532532532532533	0.346201611195215 \\
0.54054054054054	0.344717316188045 \\
0.548548548548548	0.343217374289848 \\
0.556556556556557	0.341702045536168 \\
0.564564564564565	0.340171591776383 \\
0.572572572572573	0.338626276598684 \\
0.58058058058058	0.337066365254824 \\
0.588588588588588	0.335492124584683 \\
0.596596596596597	0.333903822940662 \\
0.604604604604605	0.33230173011195 \\
0.612612612612613	0.330686117248681 \\
0.62062062062062	0.329057256786028 \\
0.628628628628628	0.327415422368237 \\
0.636636636636637	0.325760888772662 \\
0.644644644644645	0.324093931833804 \\
0.652652652652653	0.322414828367396 \\
0.66066066066066	0.32072385609456 \\
0.668668668668668	0.319021293566066 \\
0.676676676676677	0.317307420086723 \\
0.684684684684685	0.31558251563992 \\
0.692692692692693	0.313846860812361 \\
0.7007007007007	0.312100736719007 \\
0.708708708708708	0.310344424928268 \\
0.716716716716717	0.308578207387456 \\
0.724724724724725	0.306802366348536 \\
0.732732732732733	0.305017184294205 \\
0.74074074074074	0.303222943864308 \\
0.748748748748748	0.301419927782648 \\
0.756756756756757	0.299608418784177 \\
0.764764764764765	0.29778869954263 \\
0.772772772772773	0.295961052598604 \\
0.780780780780781	0.294125760288111 \\
0.788788788788788	0.292283104671645 \\
0.796796796796797	0.290433367463758 \\
0.804804804804805	0.288576829963198 \\
0.812812812812813	0.286713772983616 \\
0.820820820820821	0.284844476784867 \\
0.828828828828828	0.282969221004931 \\
0.836836836836837	0.281088284592474 \\
0.844844844844845	0.279201945740075 \\
0.852852852852853	0.277310481818125 \\
0.860860860860861	0.275414169309444 \\
0.868868868868869	0.273513283744617 \\
0.876876876876877	0.271608099638066 \\
0.884884884884885	0.269698890424904 \\
0.892892892892893	0.267785928398552 \\
0.900900900900901	0.265869484649172 \\
0.908908908908909	0.263949829002911 \\
0.916916916916917	0.262027229961987 \\
0.924924924924925	0.260101954645624 \\
0.932932932932933	0.258174268731856 \\
0.940940940940941	0.256244436400228 \\
0.948948948948949	0.254312720275384 \\
0.956956956956957	0.252379381371581 \\
0.964964964964965	0.250444679038131 \\
0.972972972972973	0.248508870905785 \\
0.980980980980981	0.246572212834085 \\
0.988988988988989	0.244634958859672 \\
0.996996996996997	0.242697361145593 \\
1.00500500500501	0.24075966993159 \\
1.01301301301301	0.238822133485401 \\
1.02102102102102	0.236884998055083 \\
1.02902902902903	0.234948507822352 \\
1.03703703703704	0.233012904856966 \\
1.04504504504505	0.231078429072154 \\
1.05305305305305	0.229145318181096 \\
1.06106106106106	0.22721380765447 \\
1.06906906906907	0.225284130679062 \\
1.07707707707708	0.223356518117463 \\
1.08508508508509	0.221431198468836 \\
1.09309309309309	0.219508397830784 \\
1.1011011011011	0.217588339862305 \\
1.10910910910911	0.215671245747847 \\
1.11711711711712	0.21375733416247 \\
1.12512512512513	0.21184682123811 \\
1.13313313313313	0.209939920530956 \\
1.14114114114114	0.208036842989938 \\
1.14914914914915	0.206137796926331 \\
1.15715715715716	0.204242987984477 \\
1.16516516516517	0.202352619113618 \\
1.17317317317317	0.200466890540856 \\
1.18118118118118	0.198585999745228 \\
1.18918918918919	0.19671014143289 \\
1.1971971971972	0.194839507513435 \\
1.20520520520521	0.192974287077311 \\
1.21321321321321	0.191114666374361 \\
1.22122122122122	0.18926082879347 \\
1.22922922922923	0.187412954843324 \\
1.23723723723724	0.185571222134271 \\
1.24524524524525	0.183735805361283 \\
1.25325325325325	0.181906876288021 \\
1.26126126126126	0.180084603731981 \\
1.26926926926927	0.178269153550739 \\
1.27727727727728	0.176460688629273 \\
1.28528528528529	0.174659368868355 \\
1.29329329329329	0.172865351174024 \\
1.3013013013013	0.17107878944811 \\
1.30930930930931	0.169299834579821 \\
1.31731731731732	0.167528634438376 \\
1.32532532532533	0.165765333866673 \\
1.33333333333333	0.164010074675994 \\
1.34134134134134	0.16226299564173 \\
1.34934934934935	0.160524232500121 \\
1.35735735735736	0.158793917945997 \\
1.36536536536537	0.157072181631514 \\
1.37337337337337	0.155359150165875 \\
1.38138138138138	0.153654947116025 \\
1.38938938938939	0.151959693008306 \\
1.3973973973974	0.150273505331067 \\
1.40540540540541	0.148596498538208 \\
1.41341341341341	0.146928784053658 \\
1.42142142142142	0.14527047027677 \\
1.42942942942943	0.143621662588612 \\
1.43743743743744	0.141982463359159 \\
1.44544544544545	0.140352971955359 \\
1.45345345345345	0.13873328475006 \\
1.46146146146146	0.137123495131801 \\
1.46946946946947	0.13552369351543 \\
1.47747747747748	0.133933967353555 \\
1.48548548548549	0.132354401148798 \\
1.49349349349349	0.130785076466857 \\
1.5015015015015	0.129226071950339 \\
1.50950950950951	0.127677463333369 \\
1.51751751751752	0.12613932345695 \\
1.52552552552553	0.124611722285057 \\
1.53353353353353	0.123094726921466 \\
1.54154154154154	0.121588401627275 \\
1.54954954954955	0.120092807839133 \\
1.55755755755756	0.11860800418814 \\
1.56556556556557	0.117134046519405 \\
1.57357357357357	0.115670987912262 \\
1.58158158158158	0.114218878701105 \\
1.58958958958959	0.112777766496842 \\
1.5975975975976	0.111347696208955 \\
1.60560560560561	0.10992871006813 \\
1.61361361361361	0.108520847649465 \\
1.62162162162162	0.107124145896225 \\
1.62962962962963	0.105738639144131 \\
1.63763763763764	0.10436435914617 \\
1.64564564564565	0.103001335097907 \\
1.65365365365365	0.10164959366328 \\
1.66166166166166	0.100309159000872 \\
1.66966966966967	0.0989800527906321 \\
1.67767767767768	0.0976622942610365 \\
1.68568568568569	0.0963559002166692 \\
1.69369369369369	0.0950608850662112 \\
1.7017017017017	0.0937772608508176 \\
1.70970970970971	0.0925050372728685 \\
1.71771771771772	0.0912442217250772 \\
1.72572572572573	0.0899948193199405 \\
1.73373373373373	0.088756832919514 \\
1.74174174174174	0.0875302631654974 \\
1.74974974974975	0.086315108509615 \\
1.75775775775776	0.0851113652442723 \\
1.76576576576577	0.0839190275334778 \\
1.77377377377377	0.082738087444011 \\
1.78178178178178	0.0815685349768226 \\
1.78978978978979	0.0804103580986525 \\
1.7977977977978	0.0792635427738473 \\
1.80580580580581	0.0781280729963669 \\
1.81381381381381	0.0770039308219615 \\
1.82182182182182	0.0758910964005057 \\
1.82982982982983	0.0747895480084762 \\
1.83783783783784	0.0736992620815557 \\
1.84584584584585	0.0726202132473522 \\
1.85385385385385	0.0715523743582165 \\
1.86186186186186	0.0704957165241465 \\
1.86986986986987	0.0694502091457625 \\
1.87787787787788	0.0684158199473405 \\
1.88588588588589	0.0673925150098903 \\
1.89389389389389	0.0663802588042655 \\
1.9019019019019	0.0653790142242912 \\
1.90990990990991	0.0643887426198963 \\
1.91791791791792	0.06340940383024 \\
1.92592592592593	0.0624409562168171 \\
1.93393393393393	0.0614833566965309 \\
1.94194194194194	0.0605365607747232 \\
1.94994994994995	0.0596005225781457 \\
1.95795795795796	0.0586751948878651 \\
1.96596596596597	0.0577605291720876 \\
1.97397397397397	0.056856475618893 \\
1.98198198198198	0.0559629831688668 \\
1.98998998998999	0.0550799995476186 \\
1.997997997998	0.0542074712981785 \\
2.00600600600601	0.0533453438132581 \\
2.01401401401401	0.0524935613673681 \\
2.02202202202202	0.0516520671487823 \\
2.03003003003003	0.0508208032913361 \\
2.03803803803804	0.0499997109060533 \\
2.04604604604605	0.0491887301125894 \\
2.05405405405405	0.0483878000704834 \\
2.06206206206206	0.0475968590102091 \\
2.07007007007007	0.0468158442640166 \\
2.07807807807808	0.0460446922965577 \\
2.08608608608609	0.0452833387352837 \\
2.09409409409409	0.0445317184006117 \\
2.1021021021021	0.043789765335848 \\
2.11011011011011	0.0430574128368641 \\
2.11811811811812	0.0423345934815158 \\
2.12612612612613	0.0416212391588004 \\
2.13413413413413	0.0409172810977437 \\
2.14214214214214	0.0402226498960119 \\
2.15015015015015	0.03953727554824 \\
2.15815815815816	0.0388610874740724 \\
2.16616616616617	0.03819401454591 \\
2.17417417417417	0.0375359851163567 \\
2.18218218218218	0.0368869270453615 \\
2.19019019019019	0.0362467677270497 \\
2.1981981981982	0.0356154341162401 \\
2.20620620620621	0.0349928527546413 \\
2.21421421421421	0.0343789497967251 \\
2.22222222222222	0.0337736510352706 \\
2.23023023023023	0.0331768819265761 \\
2.23823823823824	0.0325885676153351 \\
2.24624624624625	0.0320086329591724 \\
2.25425425425425	0.0314370025528369 \\
2.26226226226226	0.0308736007520491 \\
2.27027027027027	0.0303183516969979 \\
2.27827827827828	0.029771179335487 \\
2.28628628628629	0.0292320074457259 \\
2.29429429429429	0.0287007596587648 \\
2.3023023023023	0.0281773594805703 \\
2.31031031031031	0.0276617303137407 \\
2.31831831831832	0.0271537954788581 \\
2.32632632632633	0.0266534782354776 \\
2.33433433433433	0.0261607018027501 \\
2.34234234234234	0.0256753893796796 \\
2.35035035035035	0.0251974641650118 \\
2.35835835835836	0.0247268493767558 \\
2.36636636636637	0.0242634682713357 \\
2.37437437437437	0.023807244162374 \\
2.38238238238238	0.0233581004391047 \\
2.39039039039039	0.022915960584417 \\
2.3983983983984	0.0224807481925294 \\
2.40640640640641	0.0220523869862944 \\
2.41441441441441	0.0216308008341344 \\
2.42242242242242	0.0212159137666093 \\
2.43043043043043	0.0208076499926159 \\
2.43843843843844	0.020405933915221 \\
2.44644644644645	0.0200106901471285 \\
2.45445445445445	0.0196218435257817 \\
2.46246246246246	0.0192393191281025 \\
2.47047047047047	0.0188630422848682 \\
2.47847847847848	0.0184929385947284 \\
2.48648648648649	0.0181289339378626 \\
2.49449449449449	0.0177709544892815 \\
2.5025025025025	0.0174189267317727 \\
2.51051051051051	0.0170727774684938 \\
2.51851851851852	0.0167324338352152 \\
2.52652652652653	0.016397823312213 \\
2.53453453453453	0.0160688737358178 \\
2.54254254254254	0.0157455133096184 \\
2.55055055055055	0.0154276706153247 \\
2.55855855855856	0.0151152746232934 \\
2.56656656656657	0.0148082547027168 \\
2.57457457457457	0.0145065406314803 \\
2.58258258258258	0.014210062605689 \\
2.59059059059059	0.0139187512488692 \\
2.5985985985986	0.0136325376208454 \\
2.60660660660661	0.0133513532262973 \\
2.61461461461461	0.0130751300230007 \\
2.62262262262262	0.0128038004297541 \\
2.63063063063063	0.0125372973339956 \\
2.63863863863864	0.0122755540991133 \\
2.64664664664665	0.0120185045714526 \\
2.65465465465465	0.0117660830870247 \\
2.66266266266266	0.0115182244779185 \\
2.67067067067067	0.0112748640784219 \\
2.67867867867868	0.011035937730854 \\
2.68668668668669	0.0108013817911136 \\
2.69469469469469	0.0105711331339476 \\
2.7027027027027	0.0103451291579421 \\
2.71071071071071	0.0101233077902421 \\
2.71871871871872	0.00990560749100247 \\
2.72672672672673	0.00969196725757422 \\
2.73473473473473	0.00948232662843099 \\
2.74274274274274	0.00927662568683898 \\
2.75075075075075	0.00907480506427511 \\
2.75875875875876	0.00887680594359712 \\
2.76676676676677	0.00868257006197001 \\
2.77477477477477	0.00849203971355293 \\
2.78278278278278	0.00830515775195071 \\
2.79079079079079	0.00812186759243432 \\
2.7987987987988	0.00794211321393438 \\
2.80680680680681	0.0077658391608121 \\
2.81481481481481	0.00759299054441176 \\
2.82282282282282	0.00742351304439893 \\
2.83083083083083	0.00725735290988893 \\
2.83883883883884	0.00709445696036947 \\
2.84684684684685	0.0069347725864219 \\
2.85485485485485	0.0067782477502453 \\
2.86286286286286	0.00662483098598741 \\
2.87087087087087	0.00647447139988703 \\
2.87887887887888	0.00632711867023165 \\
2.88688688688689	0.00618272304713484 \\
2.89489489489489	0.00604123535213746 \\
2.9029029029029	0.00590260697763674 \\
2.91091091091091	0.0057667898861475 \\
2.91891891891892	0.00563373660939975 \\
2.92692692692693	0.00550340024727644 \\
2.93493493493493	0.00537573446659573 \\
2.94294294294294	0.00525069349974172 \\
2.95095095095095	0.00512823214314764 \\
2.95895895895896	0.00500830575563558 \\
2.96696696696697	0.00489087025661667 \\
2.97497497497497	0.00477588212415564 \\
2.98298298298298	0.00466329839290368 \\
2.99099099099099	0.00455307665190356 \\
2.998998998999	0.00444517504227067 \\
3.00700700700701	0.00433955225475399 \\
3.01501501501502	0.00423616752718052 \\
3.02302302302302	0.0041349806417872 \\
3.03103103103103	0.00403595192244368 \\
3.03903903903904	0.00393904223176991 \\
3.04704704704705	0.00384421296815189 \\
3.05505505505506	0.00375142606265932 \\
3.06306306306306	0.00366064397586864 \\
3.07107107107107	0.00357182969459489 \\
3.07907907907908	0.00348494672853594 \\
3.08708708708709	0.00339995910683238 \\
3.0950950950951	0.00331683137454643 \\
3.1031031031031	0.00323552858906328 \\
3.11111111111111	0.00315601631641805 \\
3.11911911911912	0.00307826062755151 \\
3.12712712712713	0.00300222809449793 \\
3.13513513513514	0.00292788578650791 \\
3.14314314314314	0.00285520126610952 \\
3.15115115115115	0.00278414258511062 \\
3.15915915915916	0.00271467828054533 \\
3.16716716716717	0.00264677737056774 \\
3.17517517517518	0.00258040935029545 \\
3.18318318318318	0.00251554418760605 \\
3.19119119119119	0.00245215231888919 \\
3.1991991991992	0.00239020464475689 \\
3.20720720720721	0.00232967252571501 \\
3.21521521521522	0.00227052777779815 \\
3.22322322322322	0.00221274266817093 \\
3.23123123123123	0.00215628991069792 \\
3.23923923923924	0.00210114266148476 \\
3.24724724724725	0.00204727451439301 \\
3.25525525525526	0.00199465949653091 \\
3.26326326326326	0.00194327206372255 \\
3.27127127127127	0.00189308709595757 \\
3.27927927927928	0.00184407989282385 \\
3.28728728728729	0.00179622616892505 \\
3.2952952952953	0.00174950204928538 \\
3.3033033033033	0.00170388406474362 \\
3.31131131131131	0.00165934914733832 \\
3.31931931931932	0.00161587462568624 \\
3.32732732732733	0.00157343822035606 \\
3.33533533533534	0.00153201803923895 \\
3.34334334334334	0.0014915925729182 \\
3.35135135135135	0.00145214069003938 \\
3.35935935935936	0.001413641632683 \\
3.36736736736737	0.00137607501174126 \\
3.37537537537538	0.00133942080230046 \\
3.38338338338338	0.00130365933903089 \\
3.39139139139139	0.00126877131158549 \\
3.3993993993994	0.00123473776000907 \\
3.40740740740741	0.00120154007015926 \\
3.41541541541542	0.00116915996914087 \\
3.42342342342342	0.00113757952075483 \\
3.43143143143143	0.00110678112096324 \\
3.43943943943944	0.0010767474933716 \\
3.44744744744745	0.00104746168472971 \\
3.45545545545546	0.0010189070604522 \\
3.46346346346346	0.000991067300160061 \\
3.47147147147147	0.000963926393244148 \\
3.47947947947948	0.000937468634451866 \\
3.48748748748749	0.00091167861949796 \\
3.4954954954955	0.000886541240700502 \\
3.5035035035035	0.000862041682643008 \\
3.51151151151151	0.000838165417863601 \\
3.51951951951952	0.00081489820257214 \\
3.52752752752753	0.000792226072396121 \\
3.53553553553554	0.000770135338156225 \\
3.54354354354354	0.000748612581672247 \\
3.55155155155155	0.000727644651600165 \\
3.55955955955956	0.000707218659301081 \\
3.56756756756757	0.000687321974742665 \\
3.57557557557558	0.000667942222433796 \\
3.58358358358358	0.000649067277392974 \\
3.59159159159159	0.000630685261151104 \\
3.5995995995996	0.000612784537789192 \\
3.60760760760761	0.000595353710011453 \\
3.61561561561562	0.00057838161525435 \\
3.62362362362362	0.000561857321831992 \\
3.63163163163163	0.000545770125118335 \\
3.63963963963964	0.000530109543766577 \\
3.64764764764765	0.000514865315966112 \\
3.65565565565566	0.000500027395737409 \\
3.66366366366366	0.000485585949265104 \\
3.67167167167167	0.000471531351269623 \\
3.67967967967968	0.000457854181417586 \\
3.68768768768769	0.00044454522077123 \\
3.6956956956957	0.00043159544827708 \\
3.7037037037037	0.000418996037294059 \\
3.71171171171171	0.000406738352161196 \\
3.71971971971972	0.000394813944805093 \\
3.72772772772773	0.000383214551387261 \\
3.73573573573574	0.000371932088991452 \\
3.74374374374374	0.000360958652351047 \\
3.75175175175175	0.000350286510616579 \\
3.75975975975976	0.000339908104163431 \\
3.76776776776777	0.000329816041439723 \\
3.77577577577578	0.000320003095854415 \\
3.78378378378378	0.0003104622027056 \\
3.79179179179179	0.000301186456148956 \\
3.7997997997998	0.000292169106206322 \\
3.80780780780781	0.000283403555814325 \\
3.81581581581582	0.000274883357912987 \\
3.82382382382382	0.000266602212574214 \\
3.83183183183183	0.000258553964170059 \\
3.83983983983984	0.000250732598580649 \\
3.84784784784785	0.000243132240441603 \\
3.85585585585586	0.000235747150430845 \\
3.86386386386386	0.000228571722594601 \\
3.87187187187187	0.000221600481712426 \\
3.87987987987988	0.000214828080701088 \\
3.88788788788789	0.000208249298057069 \\
3.8958958958959	0.000201859035337517 \\
3.9039039039039	0.000195652314679408 \\
3.91191191191191	0.000189624276356684 \\
3.91991991991992	0.000183770176375138 \\
3.92792792792793	0.00017808538410479 \\
3.93593593593594	0.000172565379949497 \\
3.94394394394394	0.00016720575305353 \\
3.95195195195195	0.00016200219904485 \\
3.95995995995996	0.000156950517814782 \\
3.96796796796797	0.000152046611333821 \\
3.97597597597598	0.000147286481503255 \\
3.98398398398398	0.000142666228042307 \\
3.99199199199199	0.000138182046410498 \\
4	0.000133830225764885 \\
};
\end{axis}

\end{tikzpicture}

%% file: img/lstm_hist.tex
\begin{tikzpicture}[scale=\MyScale]

\definecolor{color0}{rgb}{0.12156862745098,0.466666666666667,0.705882352941177}
\definecolor{color1}{rgb}{1,0.498039215686275,0.0549019607843137}

\begin{axis}[
hide x axis,
hide y axis,
tick align=outside,
tick pos=left,
x grid style={white!69.01960784313725!black},
xmin=-4.4, xmax=4.4,
y grid style={white!69.01960784313725!black},
ymin=0, ymax=0.588677561262602
]
\draw[fill=color0,draw opacity=0] (axis cs:-4,0) rectangle (axis cs:-3.92,0.00203354400477799);
\draw[fill=color0,draw opacity=0] (axis cs:-3.92,0) rectangle (axis cs:-3.84,0.00194997370321177);
\draw[fill=color0,draw opacity=0] (axis cs:-3.84,0) rectangle (axis cs:-3.76,0.00242353874542035);
\draw[fill=color0,draw opacity=0] (axis cs:-3.76,0) rectangle (axis cs:-3.68,0.0023121116766654);
\draw[fill=color0,draw opacity=0] (axis cs:-3.68,0) rectangle (axis cs:-3.6,0.00256282258136404);
\draw[fill=color0,draw opacity=0] (axis cs:-3.6,0) rectangle (axis cs:-3.52,0.00303638762357262);
\draw[fill=color0,draw opacity=0] (axis cs:-3.52,0) rectangle (axis cs:-3.44,0.0045685098189533);
\draw[fill=color0,draw opacity=0] (axis cs:-3.44,0) rectangle (axis cs:-3.36,0.0045685098189533);
\draw[fill=color0,draw opacity=0] (axis cs:-3.36,0) rectangle (axis cs:-3.28,0.00640705645341014);
\draw[fill=color0,draw opacity=0] (axis cs:-3.28,0) rectangle (axis cs:-3.2,0.00729847300344978);
\draw[fill=color0,draw opacity=0] (axis cs:-3.2,0) rectangle (axis cs:-3.12,0.00841274369099936);
\draw[fill=color0,draw opacity=0] (axis cs:-3.12,0) rectangle (axis cs:-3.04,0.00944344407698272);
\draw[fill=color0,draw opacity=0] (axis cs:-3.04,0) rectangle (axis cs:-2.96,0.0105020012301548);
\draw[fill=color0,draw opacity=0] (axis cs:-2.96,0) rectangle (axis cs:-2.88,0.0132041076474626);
\draw[fill=color0,draw opacity=0] (axis cs:-2.88,0) rectangle (axis cs:-2.8,0.0135383888537274);
\draw[fill=color0,draw opacity=0] (axis cs:-2.8,0) rectangle (axis cs:-2.72,0.0179954716039258);
\draw[fill=color0,draw opacity=0] (axis cs:-2.72,0) rectangle (axis cs:-2.64,0.0216168513384619);
\draw[fill=color0,draw opacity=0] (axis cs:-2.64,0) rectangle (axis cs:-2.56,0.0220347028462931);
\draw[fill=color0,draw opacity=0] (axis cs:-2.56,0) rectangle (axis cs:-2.48,0.0267146397340013);
\draw[fill=color0,draw opacity=0] (axis cs:-2.48,0) rectangle (axis cs:-2.4,0.0303917330029149);
\draw[fill=color0,draw opacity=0] (axis cs:-2.4,0) rectangle (axis cs:-2.32,0.0320909958014282);
\draw[fill=color0,draw opacity=0] (axis cs:-2.32,0) rectangle (axis cs:-2.24,0.035628805234398);
\draw[fill=color0,draw opacity=0] (axis cs:-2.24,0) rectangle (axis cs:-2.16,0.0409494477674472);
\draw[fill=color0,draw opacity=0] (axis cs:-2.16,0) rectangle (axis cs:-2.08,0.0446822545707383);
\draw[fill=color0,draw opacity=0] (axis cs:-2.08,0) rectangle (axis cs:-2,0.0500028971037876);
\draw[fill=color0,draw opacity=0] (axis cs:-2,0) rectangle (axis cs:-1.92,0.0520085843413768);
\draw[fill=color0,draw opacity=0] (axis cs:-1.92,0) rectangle (axis cs:-1.84,0.06033775773081);
\draw[fill=color0,draw opacity=0] (axis cs:-1.84,0) rectangle (axis cs:-1.76,0.0695862044374715);
\draw[fill=color0,draw opacity=0] (axis cs:-1.76,0) rectangle (axis cs:-1.68,0.0727340191297995);
\draw[fill=color0,draw opacity=0] (axis cs:-1.68,0) rectangle (axis cs:-1.6,0.0905344933634037);
\draw[fill=color0,draw opacity=0] (axis cs:-1.6,0) rectangle (axis cs:-1.52,0.139506690081208);
\draw[fill=color0,draw opacity=0] (axis cs:-1.52,0) rectangle (axis cs:-1.44,0.211154295290646);
\draw[fill=color0,draw opacity=0] (axis cs:-1.44,0) rectangle (axis cs:-1.36,0.212797844554782);
\draw[fill=color0,draw opacity=0] (axis cs:-1.36,0) rectangle (axis cs:-1.28,0.199287312468243);
\draw[fill=color0,draw opacity=0] (axis cs:-1.28,0) rectangle (axis cs:-1.2,0.208396475338961);
\draw[fill=color0,draw opacity=0] (axis cs:-1.2,0) rectangle (axis cs:-1.12,0.239094632780953);
\draw[fill=color0,draw opacity=0] (axis cs:-1.12,0) rectangle (axis cs:-1.04,0.34698389210294);
\draw[fill=color0,draw opacity=0] (axis cs:-1.04,0) rectangle (axis cs:-0.96,0.478997111810377);
\draw[fill=color0,draw opacity=0] (axis cs:-0.96,0) rectangle (axis cs:-0.88,0.440749770460238);
\draw[fill=color0,draw opacity=0] (axis cs:-0.88,0) rectangle (axis cs:-0.8,0.294334602116223);
\draw[fill=color0,draw opacity=0] (axis cs:-0.8,0) rectangle (axis cs:-0.72,0.251769461851829);
\draw[fill=color0,draw opacity=0] (axis cs:-0.72,0) rectangle (axis cs:-0.64,0.249485206942353);
\draw[fill=color0,draw opacity=0] (axis cs:-0.64,0) rectangle (axis cs:-0.56,0.244749556520266);
\draw[fill=color0,draw opacity=0] (axis cs:-0.56,0) rectangle (axis cs:-0.48,0.254499425036325);
\draw[fill=color0,draw opacity=0] (axis cs:-0.48,0) rectangle (axis cs:-0.4,0.263274306700778);
\draw[fill=color0,draw opacity=0] (axis cs:-0.4,0) rectangle (axis cs:-0.32,0.272494896640251);
\draw[fill=color0,draw opacity=0] (axis cs:-0.32,0) rectangle (axis cs:-0.24,0.279737656109323);
\draw[fill=color0,draw opacity=0] (axis cs:-0.24,0) rectangle (axis cs:-0.16,0.288484681006589);
\draw[fill=color0,draw opacity=0] (axis cs:-0.16,0) rectangle (axis cs:-0.0800000000000001,0.293387472031805);
\draw[fill=color0,draw opacity=0] (axis cs:-0.0800000000000001,0) rectangle (axis cs:0,0.297565987110116);
\draw[fill=color0,draw opacity=0] (axis cs:0,0) rectangle (axis cs:0.0800000000000001,0.299794528485216);
\draw[fill=color0,draw opacity=0] (axis cs:0.0800000000000001,0) rectangle (axis cs:0.16,0.296674570560077);
\draw[fill=color0,draw opacity=0] (axis cs:0.16,0) rectangle (axis cs:0.24,0.292440341947388);
\draw[fill=color0,draw opacity=0] (axis cs:0.24,0) rectangle (axis cs:0.32,0.288038972731567);
\draw[fill=color0,draw opacity=0] (axis cs:0.32,0) rectangle (axis cs:0.4,0.285281152779882);
\draw[fill=color0,draw opacity=0] (axis cs:0.4,0) rectangle (axis cs:0.48,0.281158351235949);
\draw[fill=color0,draw opacity=0] (axis cs:0.48,0) rectangle (axis cs:0.56,0.283108324939161);
\draw[fill=color0,draw opacity=0] (axis cs:0.56,0) rectangle (axis cs:0.64,0.300630231500881);
\draw[fill=color0,draw opacity=0] (axis cs:0.64,0) rectangle (axis cs:0.72,0.317845713623519);
\draw[fill=color0,draw opacity=0] (axis cs:0.72,0) rectangle (axis cs:0.8,0.366929337410078);
\draw[fill=color0,draw opacity=0] (axis cs:0.8,0) rectangle (axis cs:0.88,0.452366042377942);
\draw[fill=color0,draw opacity=0] (axis cs:0.88,0) rectangle (axis cs:0.96,0.560645296440573);
\draw[fill=color0,draw opacity=0] (axis cs:0.96,0) rectangle (axis cs:1.04,0.555658935113789);
\draw[fill=color0,draw opacity=0] (axis cs:1.04,0) rectangle (axis cs:1.12,0.369018594949234);
\draw[fill=color0,draw opacity=0] (axis cs:1.12,0) rectangle (axis cs:1.2,0.216447081056507);
\draw[fill=color0,draw opacity=0] (axis cs:1.2,0) rectangle (axis cs:1.28,0.167168459899626);
\draw[fill=color0,draw opacity=0] (axis cs:1.28,0) rectangle (axis cs:1.36,0.161457822625935);
\draw[fill=color0,draw opacity=0] (axis cs:1.36,0) rectangle (axis cs:1.44,0.17106840730605);
\draw[fill=color0,draw opacity=0] (axis cs:1.44,0) rectangle (axis cs:1.52,0.158867143277382);
\draw[fill=color0,draw opacity=0] (axis cs:1.52,0) rectangle (axis cs:1.6,0.0972201174887012);
\draw[fill=color0,draw opacity=0] (axis cs:1.6,0) rectangle (axis cs:1.68,0.054627120457119);
\draw[fill=color0,draw opacity=0] (axis cs:1.68,0) rectangle (axis cs:1.76,0.0412837289737121);
\draw[fill=color0,draw opacity=0] (axis cs:1.76,0) rectangle (axis cs:1.84,0.0360745135094178);
\draw[fill=color0,draw opacity=0] (axis cs:1.84,0) rectangle (axis cs:1.92,0.0310045818810672);
\draw[fill=color0,draw opacity=0] (axis cs:1.92,0) rectangle (axis cs:2,0.0257396528823954);
\draw[fill=color0,draw opacity=0] (axis cs:2,0) rectangle (axis cs:2.08,0.0224804111213129);
\draw[fill=color0,draw opacity=0] (axis cs:2.08,0) rectangle (axis cs:2.16,0.0186083204820781);
\draw[fill=color0,draw opacity=0] (axis cs:2.16,0) rectangle (axis cs:2.24,0.0162683520382239);
\draw[fill=color0,draw opacity=0] (axis cs:2.24,0) rectangle (axis cs:2.32,0.0144019486365784);
\draw[fill=color0,draw opacity=0] (axis cs:2.32,0) rectangle (axis cs:2.4,0.0124241181661779);
\draw[fill=color0,draw opacity=0] (axis cs:2.4,0) rectangle (axis cs:2.48,0.0107527121348535);
\draw[fill=color0,draw opacity=0] (axis cs:2.48,0) rectangle (axis cs:2.56,0.00933201700822777);
\draw[fill=color0,draw opacity=0] (axis cs:2.56,0) rectangle (axis cs:2.64,0.00807846248473448);
\draw[fill=color0,draw opacity=0] (axis cs:2.64,0) rectangle (axis cs:2.72,0.00749347037377104);
\draw[fill=color0,draw opacity=0] (axis cs:2.72,0) rectangle (axis cs:2.8,0.0054877831361817);
\draw[fill=color0,draw opacity=0] (axis cs:2.8,0) rectangle (axis cs:2.88,0.00571063727369162);
\draw[fill=color0,draw opacity=0] (axis cs:2.88,0) rectangle (axis cs:2.96,0.00490279102521817);
\draw[fill=color0,draw opacity=0] (axis cs:2.96,0) rectangle (axis cs:3.04,0.0042899421470659);
\draw[fill=color0,draw opacity=0] (axis cs:3.04,0) rectangle (axis cs:3.12,0.00314781469232757);
\draw[fill=color0,draw opacity=0] (axis cs:3.12,0) rectangle (axis cs:3.2,0.00300853085638388);
\draw[fill=color0,draw opacity=0] (axis cs:3.2,0) rectangle (axis cs:3.28,0.00261853611574152);
\draw[fill=color0,draw opacity=0] (axis cs:3.28,0) rectangle (axis cs:3.36,0.00231211167666539);
\draw[fill=color0,draw opacity=0] (axis cs:3.36,0) rectangle (axis cs:3.44,0.00211711430634421);
\draw[fill=color0,draw opacity=0] (axis cs:3.44,0) rectangle (axis cs:3.52,0.00211711430634421);
\draw[fill=color0,draw opacity=0] (axis cs:3.52,0) rectangle (axis cs:3.6,0.00164354926413564);
\draw[fill=color0,draw opacity=0] (axis cs:3.6,0) rectangle (axis cs:3.68,0.00169926279851313);
\draw[fill=color0,draw opacity=0] (axis cs:3.68,0) rectangle (axis cs:3.76,0.00136498159224824);
\draw[fill=color0,draw opacity=0] (axis cs:3.76,0) rectangle (axis cs:3.84,0.00122569775630454);
\draw[fill=color0,draw opacity=0] (axis cs:3.84,0) rectangle (axis cs:3.92,0.000835703015662188);
\draw[fill=color0,draw opacity=0] (axis cs:3.92,0) rectangle (axis cs:4,0.000807846248473448);
\addplot [very thick, color1, forget plot]
table [row sep=\\]{%
-4	0.000133830225764885 \\
-3.99199199199199	0.000138182046410498 \\
-3.98398398398398	0.000142666228042307 \\
-3.97597597597598	0.000147286481503254 \\
-3.96796796796797	0.000152046611333821 \\
-3.95995995995996	0.000156950517814782 \\
-3.95195195195195	0.00016200219904485 \\
-3.94394394394394	0.00016720575305353 \\
-3.93593593593594	0.000172565379949496 \\
-3.92792792792793	0.00017808538410479 \\
-3.91991991991992	0.000183770176375138 \\
-3.91191191191191	0.000189624276356684 \\
-3.9039039039039	0.000195652314679408 \\
-3.8958958958959	0.000201859035337517 \\
-3.88788788788789	0.000208249298057069 \\
-3.87987987987988	0.000214828080701088 \\
-3.87187187187187	0.000221600481712426 \\
-3.86386386386386	0.0002285717225946 \\
-3.85585585585586	0.000235747150430845 \\
-3.84784784784785	0.000243132240441603 \\
-3.83983983983984	0.000250732598580649 \\
-3.83183183183183	0.000258553964170059 \\
-3.82382382382382	0.000266602212574213 \\
-3.81581581581582	0.000274883357912987 \\
-3.80780780780781	0.000283403555814325 \\
-3.7997997997998	0.000292169106206322 \\
-3.79179179179179	0.000301186456148957 \\
-3.78378378378378	0.0003104622027056 \\
-3.77577577577578	0.000320003095854415 \\
-3.76776776776777	0.000329816041439723 \\
-3.75975975975976	0.000339908104163431 \\
-3.75175175175175	0.00035028651061658 \\
-3.74374374374374	0.000360958652351047 \\
-3.73573573573574	0.000371932088991451 \\
-3.72772772772773	0.000383214551387261 \\
-3.71971971971972	0.000394813944805093 \\
-3.71171171171171	0.000406738352161197 \\
-3.7037037037037	0.000418996037294058 \\
-3.6956956956957	0.00043159544827708 \\
-3.68768768768769	0.00044454522077123 \\
-3.67967967967968	0.000457854181417586 \\
-3.67167167167167	0.000471531351269624 \\
-3.66366366366366	0.000485585949265103 \\
-3.65565565565566	0.000500027395737409 \\
-3.64764764764765	0.000514865315966112 \\
-3.63963963963964	0.000530109543766577 \\
-3.63163163163163	0.000545770125118336 \\
-3.62362362362362	0.000561857321831991 \\
-3.61561561561562	0.00057838161525435 \\
-3.60760760760761	0.000595353710011453 \\
-3.5995995995996	0.000612784537789192 \\
-3.59159159159159	0.000630685261151105 \\
-3.58358358358358	0.000649067277392973 \\
-3.57557557557558	0.000667942222433796 \\
-3.56756756756757	0.000687321974742665 \\
-3.55955955955956	0.000707218659301081 \\
-3.55155155155155	0.000727644651600166 \\
-3.54354354354354	0.000748612581672245 \\
-3.53553553553554	0.000770135338156225 \\
-3.52752752752753	0.000792226072396121 \\
-3.51951951951952	0.00081489820257214 \\
-3.51151151151151	0.000838165417863603 \\
-3.5035035035035	0.000862041682643007 \\
-3.4954954954955	0.000886541240700502 \\
-3.48748748748749	0.00091167861949796 \\
-3.47947947947948	0.000937468634451866 \\
-3.47147147147147	0.000963926393244149 \\
-3.46346346346346	0.000991067300160059 \\
-3.45545545545546	0.0010189070604522 \\
-3.44744744744745	0.00104746168472971 \\
-3.43943943943944	0.0010767474933716 \\
-3.43143143143143	0.00110678112096323 \\
-3.42342342342342	0.00113757952075483 \\
-3.41541541541542	0.00116915996914087 \\
-3.40740740740741	0.00120154007015926 \\
-3.3993993993994	0.00123473776000907 \\
-3.39139139139139	0.00126877131158549 \\
-3.38338338338338	0.00130365933903088 \\
-3.37537537537538	0.00133942080230046 \\
-3.36736736736737	0.00137607501174126 \\
-3.35935935935936	0.001413641632683 \\
-3.35135135135135	0.00145214069003938 \\
-3.34334334334334	0.0014915925729182 \\
-3.33533533533534	0.00153201803923895 \\
-3.32732732732733	0.00157343822035606 \\
-3.31931931931932	0.00161587462568624 \\
-3.31131131131131	0.00165934914733831 \\
-3.3033033033033	0.00170388406474362 \\
-3.2952952952953	0.00174950204928538 \\
-3.28728728728729	0.00179622616892505 \\
-3.27927927927928	0.00184407989282385 \\
-3.27127127127127	0.00189308709595757 \\
-3.26326326326326	0.00194327206372255 \\
-3.25525525525526	0.00199465949653091 \\
-3.24724724724725	0.00204727451439301 \\
-3.23923923923924	0.00210114266148476 \\
-3.23123123123123	0.00215628991069791 \\
-3.22322322322322	0.00221274266817093 \\
-3.21521521521522	0.00227052777779815 \\
-3.20720720720721	0.00232967252571501 \\
-3.1991991991992	0.00239020464475689 \\
-3.19119119119119	0.00245215231888918 \\
-3.18318318318318	0.00251554418760605 \\
-3.17517517517518	0.00258040935029545 \\
-3.16716716716717	0.00264677737056774 \\
-3.15915915915916	0.00271467828054534 \\
-3.15115115115115	0.00278414258511061 \\
-3.14314314314314	0.00285520126610952 \\
-3.13513513513514	0.00292788578650791 \\
-3.12712712712713	0.00300222809449793 \\
-3.11911911911912	0.00307826062755151 \\
-3.11111111111111	0.00315601631641804 \\
-3.1031031031031	0.00323552858906328 \\
-3.0950950950951	0.00331683137454643 \\
-3.08708708708709	0.00339995910683238 \\
-3.07907907907908	0.00348494672853595 \\
-3.07107107107107	0.00357182969459489 \\
-3.06306306306306	0.00366064397586864 \\
-3.05505505505506	0.00375142606265932 \\
-3.04704704704705	0.00384421296815189 \\
-3.03903903903904	0.00393904223176991 \\
-3.03103103103103	0.00403595192244368 \\
-3.02302302302302	0.0041349806417872 \\
-3.01501501501502	0.00423616752718052 \\
-3.00700700700701	0.00433955225475399 \\
-2.998998998999	0.00444517504227068 \\
-2.99099099099099	0.00455307665190355 \\
-2.98298298298298	0.00466329839290368 \\
-2.97497497497497	0.00477588212415564 \\
-2.96696696696697	0.00489087025661667 \\
-2.95895895895896	0.00500830575563558 \\
-2.95095095095095	0.00512823214314763 \\
-2.94294294294294	0.00525069349974172 \\
-2.93493493493493	0.00537573446659573 \\
-2.92692692692693	0.00550340024727644 \\
-2.91891891891892	0.00563373660939975 \\
-2.91091091091091	0.0057667898861475 \\
-2.9029029029029	0.00590260697763674 \\
-2.89489489489489	0.00604123535213746 \\
-2.88688688688689	0.00618272304713484 \\
-2.87887887887888	0.00632711867023164 \\
-2.87087087087087	0.00647447139988703 \\
-2.86286286286286	0.00662483098598741 \\
-2.85485485485485	0.0067782477502453 \\
-2.84684684684685	0.0069347725864219 \\
-2.83883883883884	0.00709445696036946 \\
-2.83083083083083	0.00725735290988893 \\
-2.82282282282282	0.00742351304439893 \\
-2.81481481481481	0.00759299054441176 \\
-2.80680680680681	0.0077658391608121 \\
-2.7987987987988	0.00794211321393437 \\
-2.79079079079079	0.00812186759243432 \\
-2.78278278278278	0.00830515775195071 \\
-2.77477477477477	0.00849203971355293 \\
-2.76676676676677	0.00868257006197001 \\
-2.75875875875876	0.00887680594359712 \\
-2.75075075075075	0.00907480506427511 \\
-2.74274274274274	0.00927662568683898 \\
-2.73473473473473	0.00948232662843099 \\
-2.72672672672673	0.00969196725757422 \\
-2.71871871871872	0.00990560749100246 \\
-2.71071071071071	0.0101233077902421 \\
-2.7027027027027	0.0103451291579421 \\
-2.69469469469469	0.0105711331339476 \\
-2.68668668668669	0.0108013817911136 \\
-2.67867867867868	0.011035937730854 \\
-2.67067067067067	0.0112748640784219 \\
-2.66266266266266	0.0115182244779185 \\
-2.65465465465465	0.0117660830870247 \\
-2.64664664664665	0.0120185045714526 \\
-2.63863863863864	0.0122755540991133 \\
-2.63063063063063	0.0125372973339956 \\
-2.62262262262262	0.0128038004297541 \\
-2.61461461461461	0.0130751300230007 \\
-2.60660660660661	0.0133513532262973 \\
-2.5985985985986	0.0136325376208454 \\
-2.59059059059059	0.0139187512488692 \\
-2.58258258258258	0.014210062605689 \\
-2.57457457457457	0.0145065406314803 \\
-2.56656656656657	0.0148082547027168 \\
-2.55855855855856	0.0151152746232934 \\
-2.55055055055055	0.0154276706153247 \\
-2.54254254254254	0.0157455133096184 \\
-2.53453453453453	0.0160688737358178 \\
-2.52652652652653	0.016397823312213 \\
-2.51851851851852	0.0167324338352151 \\
-2.51051051051051	0.0170727774684938 \\
-2.5025025025025	0.0174189267317727 \\
-2.49449449449449	0.0177709544892815 \\
-2.48648648648649	0.0181289339378627 \\
-2.47847847847848	0.0184929385947284 \\
-2.47047047047047	0.0188630422848682 \\
-2.46246246246246	0.0192393191281025 \\
-2.45445445445445	0.0196218435257817 \\
-2.44644644644645	0.0200106901471285 \\
-2.43843843843844	0.020405933915221 \\
-2.43043043043043	0.0208076499926159 \\
-2.42242242242242	0.0212159137666093 \\
-2.41441441441441	0.0216308008341344 \\
-2.40640640640641	0.0220523869862944 \\
-2.3983983983984	0.0224807481925294 \\
-2.39039039039039	0.022915960584417 \\
-2.38238238238238	0.0233581004391047 \\
-2.37437437437437	0.023807244162374 \\
-2.36636636636637	0.0242634682713357 \\
-2.35835835835836	0.0247268493767558 \\
-2.35035035035035	0.0251974641650118 \\
-2.34234234234234	0.0256753893796796 \\
-2.33433433433433	0.0261607018027501 \\
-2.32632632632633	0.0266534782354775 \\
-2.31831831831832	0.0271537954788581 \\
-2.31031031031031	0.0276617303137407 \\
-2.3023023023023	0.0281773594805703 \\
-2.29429429429429	0.0287007596587648 \\
-2.28628628628629	0.0292320074457259 \\
-2.27827827827828	0.029771179335487 \\
-2.27027027027027	0.0303183516969979 \\
-2.26226226226226	0.0308736007520491 \\
-2.25425425425425	0.0314370025528369 \\
-2.24624624624625	0.0320086329591723 \\
-2.23823823823824	0.0325885676153351 \\
-2.23023023023023	0.0331768819265761 \\
-2.22222222222222	0.0337736510352706 \\
-2.21421421421421	0.0343789497967251 \\
-2.20620620620621	0.0349928527546413 \\
-2.1981981981982	0.0356154341162401 \\
-2.19019019019019	0.0362467677270497 \\
-2.18218218218218	0.0368869270453615 \\
-2.17417417417417	0.0375359851163567 \\
-2.16616616616617	0.03819401454591 \\
-2.15815815815816	0.0388610874740724 \\
-2.15015015015015	0.03953727554824 \\
-2.14214214214214	0.0402226498960119 \\
-2.13413413413413	0.0409172810977437 \\
-2.12612612612613	0.0416212391588003 \\
-2.11811811811812	0.0423345934815158 \\
-2.11011011011011	0.0430574128368641 \\
-2.1021021021021	0.043789765335848 \\
-2.09409409409409	0.0445317184006117 \\
-2.08608608608609	0.0452833387352837 \\
-2.07807807807808	0.0460446922965577 \\
-2.07007007007007	0.0468158442640166 \\
-2.06206206206206	0.0475968590102091 \\
-2.05405405405405	0.0483878000704834 \\
-2.04604604604605	0.0491887301125894 \\
-2.03803803803804	0.0499997109060533 \\
-2.03003003003003	0.0508208032913361 \\
-2.02202202202202	0.0516520671487823 \\
-2.01401401401401	0.0524935613673682 \\
-2.00600600600601	0.053345343813258 \\
-1.997997997998	0.0542074712981785 \\
-1.98998998998999	0.0550799995476186 \\
-1.98198198198198	0.0559629831688668 \\
-1.97397397397397	0.0568564756188931 \\
-1.96596596596597	0.0577605291720876 \\
-1.95795795795796	0.0586751948878651 \\
-1.94994994994995	0.0596005225781457 \\
-1.94194194194194	0.0605365607747232 \\
-1.93393393393393	0.0614833566965309 \\
-1.92592592592593	0.062440956216817 \\
-1.91791791791792	0.06340940383024 \\
-1.90990990990991	0.0643887426198963 \\
-1.9019019019019	0.0653790142242912 \\
-1.89389389389389	0.0663802588042656 \\
-1.88588588588589	0.0673925150098903 \\
-1.87787787787788	0.0684158199473405 \\
-1.86986986986987	0.0694502091457625 \\
-1.86186186186186	0.0704957165241465 \\
-1.85385385385385	0.0715523743582165 \\
-1.84584584584585	0.0726202132473521 \\
-1.83783783783784	0.0736992620815557 \\
-1.82982982982983	0.0747895480084762 \\
-1.82182182182182	0.0758910964005057 \\
-1.81381381381381	0.0770039308219614 \\
-1.80580580580581	0.0781280729963669 \\
-1.7977977977978	0.0792635427738473 \\
-1.78978978978979	0.0804103580986525 \\
-1.78178178178178	0.0815685349768227 \\
-1.77377377377377	0.0827380874440109 \\
-1.76576576576577	0.0839190275334777 \\
-1.75775775775776	0.0851113652442723 \\
-1.74974974974975	0.086315108509615 \\
-1.74174174174174	0.0875302631654975 \\
-1.73373373373373	0.0887568329195139 \\
-1.72572572572573	0.0899948193199405 \\
-1.71771771771772	0.0912442217250772 \\
-1.70970970970971	0.0925050372728685 \\
-1.7017017017017	0.0937772608508177 \\
-1.69369369369369	0.0950608850662112 \\
-1.68568568568569	0.0963559002166691 \\
-1.67767767767768	0.0976622942610365 \\
-1.66966966966967	0.0989800527906321 \\
-1.66166166166166	0.100309159000872 \\
-1.65365365365365	0.10164959366328 \\
-1.64564564564565	0.103001335097907 \\
-1.63763763763764	0.10436435914617 \\
-1.62962962962963	0.105738639144131 \\
-1.62162162162162	0.107124145896225 \\
-1.61361361361361	0.108520847649465 \\
-1.60560560560561	0.10992871006813 \\
-1.5975975975976	0.111347696208955 \\
-1.58958958958959	0.112777766496842 \\
-1.58158158158158	0.114218878701105 \\
-1.57357357357357	0.115670987912262 \\
-1.56556556556557	0.117134046519405 \\
-1.55755755755756	0.11860800418814 \\
-1.54954954954955	0.120092807839133 \\
-1.54154154154154	0.121588401627275 \\
-1.53353353353353	0.123094726921466 \\
-1.52552552552553	0.124611722285057 \\
-1.51751751751752	0.12613932345695 \\
-1.50950950950951	0.127677463333369 \\
-1.5015015015015	0.129226071950339 \\
-1.49349349349349	0.130785076466857 \\
-1.48548548548549	0.132354401148798 \\
-1.47747747747748	0.133933967353555 \\
-1.46946946946947	0.13552369351543 \\
-1.46146146146146	0.137123495131801 \\
-1.45345345345345	0.13873328475006 \\
-1.44544544544545	0.140352971955359 \\
-1.43743743743744	0.141982463359159 \\
-1.42942942942943	0.143621662588612 \\
-1.42142142142142	0.14527047027677 \\
-1.41341341341341	0.146928784053658 \\
-1.40540540540541	0.148596498538208 \\
-1.3973973973974	0.150273505331067 \\
-1.38938938938939	0.151959693008306 \\
-1.38138138138138	0.153654947116025 \\
-1.37337337337337	0.155359150165875 \\
-1.36536536536537	0.157072181631514 \\
-1.35735735735736	0.158793917945997 \\
-1.34934934934935	0.160524232500121 \\
-1.34134134134134	0.16226299564173 \\
-1.33333333333333	0.164010074675994 \\
-1.32532532532533	0.165765333866673 \\
-1.31731731731732	0.167528634438376 \\
-1.30930930930931	0.169299834579821 \\
-1.3013013013013	0.17107878944811 \\
-1.29329329329329	0.172865351174024 \\
-1.28528528528529	0.174659368868355 \\
-1.27727727727728	0.176460688629273 \\
-1.26926926926927	0.178269153550739 \\
-1.26126126126126	0.180084603731981 \\
-1.25325325325325	0.181906876288021 \\
-1.24524524524525	0.183735805361283 \\
-1.23723723723724	0.185571222134271 \\
-1.22922922922923	0.187412954843324 \\
-1.22122122122122	0.18926082879347 \\
-1.21321321321321	0.191114666374361 \\
-1.20520520520521	0.192974287077311 \\
-1.1971971971972	0.194839507513435 \\
-1.18918918918919	0.19671014143289 \\
-1.18118118118118	0.198585999745227 \\
-1.17317317317317	0.200466890540856 \\
-1.16516516516517	0.202352619113618 \\
-1.15715715715716	0.204242987984477 \\
-1.14914914914915	0.206137796926331 \\
-1.14114114114114	0.208036842989938 \\
-1.13313313313313	0.209939920530956 \\
-1.12512512512513	0.21184682123811 \\
-1.11711711711712	0.21375733416247 \\
-1.10910910910911	0.215671245747847 \\
-1.1011011011011	0.217588339862305 \\
-1.09309309309309	0.219508397830784 \\
-1.08508508508509	0.221431198468836 \\
-1.07707707707708	0.223356518117463 \\
-1.06906906906907	0.225284130679063 \\
-1.06106106106106	0.22721380765447 \\
-1.05305305305305	0.229145318181096 \\
-1.04504504504505	0.231078429072154 \\
-1.03703703703704	0.233012904856966 \\
-1.02902902902903	0.234948507822352 \\
-1.02102102102102	0.236884998055083 \\
-1.01301301301301	0.238822133485401 \\
-1.00500500500501	0.24075966993159 \\
-0.996996996996997	0.242697361145593 \\
-0.988988988988989	0.244634958859672 \\
-0.980980980980981	0.246572212834085 \\
-0.972972972972973	0.248508870905785 \\
-0.964964964964965	0.250444679038131 \\
-0.956956956956957	0.252379381371581 \\
-0.948948948948949	0.254312720275385 \\
-0.940940940940941	0.256244436400228 \\
-0.932932932932933	0.258174268731856 \\
-0.924924924924925	0.260101954645624 \\
-0.916916916916917	0.262027229961987 \\
-0.908908908908909	0.263949829002911 \\
-0.900900900900901	0.265869484649172 \\
-0.892892892892893	0.267785928398552 \\
-0.884884884884885	0.269698890424904 \\
-0.876876876876877	0.271608099638066 \\
-0.868868868868869	0.273513283744617 \\
-0.860860860860861	0.275414169309444 \\
-0.852852852852853	0.277310481818125 \\
-0.844844844844845	0.279201945740075 \\
-0.836836836836837	0.281088284592474 \\
-0.828828828828829	0.28296922100493 \\
-0.820820820820821	0.284844476784867 \\
-0.812812812812813	0.286713772983616 \\
-0.804804804804805	0.288576829963198 \\
-0.796796796796797	0.290433367463758 \\
-0.788788788788789	0.292283104671645 \\
-0.780780780780781	0.294125760288111 \\
-0.772772772772773	0.295961052598604 \\
-0.764764764764765	0.29778869954263 \\
-0.756756756756757	0.299608418784177 \\
-0.748748748748749	0.301419927782648 \\
-0.740740740740741	0.303222943864308 \\
-0.732732732732733	0.305017184294205 \\
-0.724724724724725	0.306802366348536 \\
-0.716716716716717	0.308578207387456 \\
-0.708708708708709	0.310344424928268 \\
-0.700700700700701	0.312100736719007 \\
-0.692692692692693	0.313846860812361 \\
-0.684684684684685	0.31558251563992 \\
-0.676676676676677	0.317307420086723 \\
-0.668668668668669	0.319021293566066 \\
-0.660660660660661	0.320723856094559 \\
-0.652652652652653	0.322414828367396 \\
-0.644644644644645	0.324093931833804 \\
-0.636636636636636	0.325760888772662 \\
-0.628628628628629	0.327415422368237 \\
-0.620620620620621	0.329057256786028 \\
-0.612612612612613	0.330686117248681 \\
-0.604604604604605	0.33230173011195 \\
-0.596596596596596	0.333903822940662 \\
-0.588588588588589	0.335492124584683 \\
-0.580580580580581	0.337066365254824 \\
-0.572572572572573	0.338626276598684 \\
-0.564564564564565	0.340171591776383 \\
-0.556556556556556	0.341702045536168 \\
-0.548548548548549	0.343217374289848 \\
-0.54054054054054	0.344717316188045 \\
-0.532532532532533	0.346201611195215 \\
-0.524524524524525	0.347670001164422 \\
-0.516516516516516	0.349122229911825 \\
-0.508508508508509	0.350558043290856 \\
-0.5005005005005	0.351977189266054 \\
-0.492492492492492	0.353379417986526 \\
-0.484484484484485	0.354764481859011 \\
-0.476476476476476	0.356132135620502 \\
-0.468468468468469	0.357482136410418 \\
-0.46046046046046	0.358814243842273 \\
-0.452452452452452	0.360128220074839 \\
-0.444444444444445	0.361423829882744 \\
-0.436436436436436	0.362700840726495 \\
-0.428428428428429	0.363959022821904 \\
-0.42042042042042	0.365198149208855 \\
-0.412412412412412	0.366417995819425 \\
-0.404404404404405	0.3676183415453 \\
-0.396396396396396	0.368798968304465 \\
-0.388388388388389	0.369959661107154 \\
-0.38038038038038	0.37110020812101 \\
-0.372372372372372	0.372220400735444 \\
-0.364364364364365	0.373320033625156 \\
-0.356356356356356	0.374398904812801 \\
-0.348348348348348	0.375456815730761 \\
-0.34034034034034	0.376493571282005 \\
-0.332332332332332	0.377508979900006 \\
-0.324324324324325	0.378502853607702 \\
-0.316316316316316	0.379475008075448 \\
-0.308308308308308	0.380425262677969 \\
-0.3003003003003	0.38135344055026 \\
-0.292292292292292	0.382259368642425 \\
-0.284284284284285	0.38314287777343 \\
-0.276276276276276	0.384003802683735 \\
-0.268268268268268	0.3848419820868 \\
-0.26026026026026	0.385657258719425 \\
-0.252252252252252	0.386449479390916 \\
-0.244244244244244	0.387218495031049 \\
-0.236236236236236	0.387964160736805 \\
-0.228228228228228	0.38868633581787 \\
-0.22022022022022	0.389384883840873 \\
-0.212212212212212	0.390059672672329 \\
-0.204204204204204	0.390710574520297 \\
-0.196196196196196	0.391337465974705 \\
-0.188188188188188	0.39194022804635 \\
-0.18018018018018	0.392518746204527 \\
-0.172172172172172	0.393072910413305 \\
-0.164164164164164	0.393602615166401 \\
-0.156156156156156	0.394107759520658 \\
-0.148148148148148	0.394588247128102 \\
-0.14014014014014	0.395043986266573 \\
-0.132132132132132	0.3954748898689 \\
-0.124124124124124	0.395880875550626 \\
-0.116116116116116	0.396261865636259 \\
-0.108108108108108	0.396617787184037 \\
-0.1001001001001	0.396948572009207 \\
-0.0920920920920922	0.397254156705792 \\
-0.084084084084084	0.397534482666846 \\
-0.0760760760760761	0.39778949610319 \\
-0.0680680680680683	0.39801914806061 \\
-0.06006006006006	0.398223394435522 \\
-0.0520520520520522	0.398402195989086 \\
-0.0440440440440439	0.398555518359772 \\
-0.0360360360360361	0.398683332074364 \\
-0.0280280280280278	0.398785612557403 \\
-0.02002002002002	0.398862340139061 \\
-0.0120120120120122	0.398913500061449 \\
-0.00400400400400391	0.398939082483343 \\
0.00400400400400436	0.398939082483343 \\
0.0120120120120122	0.398913500061449 \\
0.02002002002002	0.398862340139061 \\
0.0280280280280278	0.398785612557403 \\
0.0360360360360357	0.398683332074364 \\
0.0440440440440444	0.398555518359772 \\
0.0520520520520522	0.398402195989086 \\
0.06006006006006	0.398223394435522 \\
0.0680680680680679	0.39801914806061 \\
0.0760760760760757	0.39778949610319 \\
0.0840840840840844	0.397534482666846 \\
0.0920920920920922	0.397254156705792 \\
0.1001001001001	0.396948572009207 \\
0.108108108108108	0.396617787184037 \\
0.116116116116116	0.396261865636259 \\
0.124124124124124	0.395880875550626 \\
0.132132132132132	0.3954748898689 \\
0.14014014014014	0.395043986266573 \\
0.148148148148148	0.394588247128102 \\
0.156156156156156	0.394107759520658 \\
0.164164164164164	0.393602615166401 \\
0.172172172172172	0.393072910413305 \\
0.18018018018018	0.392518746204527 \\
0.188188188188188	0.39194022804635 \\
0.196196196196196	0.391337465974705 \\
0.204204204204204	0.390710574520297 \\
0.212212212212212	0.390059672672329 \\
0.22022022022022	0.389384883840873 \\
0.228228228228228	0.388686335817871 \\
0.236236236236236	0.387964160736805 \\
0.244244244244245	0.387218495031049 \\
0.252252252252252	0.386449479390916 \\
0.26026026026026	0.385657258719425 \\
0.268268268268268	0.3848419820868 \\
0.276276276276276	0.384003802683735 \\
0.284284284284285	0.38314287777343 \\
0.292292292292292	0.382259368642425 \\
0.3003003003003	0.38135344055026 \\
0.308308308308308	0.380425262677969 \\
0.316316316316316	0.379475008075448 \\
0.324324324324325	0.378502853607702 \\
0.332332332332332	0.377508979900006 \\
0.34034034034034	0.376493571282005 \\
0.348348348348348	0.375456815730762 \\
0.356356356356356	0.374398904812801 \\
0.364364364364365	0.373320033625156 \\
0.372372372372372	0.372220400735444 \\
0.38038038038038	0.37110020812101 \\
0.388388388388388	0.369959661107154 \\
0.396396396396397	0.368798968304465 \\
0.404404404404405	0.3676183415453 \\
0.412412412412412	0.366417995819425 \\
0.42042042042042	0.365198149208855 \\
0.428428428428428	0.363959022821904 \\
0.436436436436437	0.362700840726495 \\
0.444444444444445	0.361423829882744 \\
0.452452452452452	0.360128220074839 \\
0.46046046046046	0.358814243842273 \\
0.468468468468468	0.357482136410418 \\
0.476476476476477	0.356132135620502 \\
0.484484484484485	0.354764481859011 \\
0.492492492492492	0.353379417986526 \\
0.5005005005005	0.351977189266054 \\
0.508508508508508	0.350558043290856 \\
0.516516516516517	0.349122229911825 \\
0.524524524524525	0.347670001164422 \\
0.532532532532533	0.346201611195215 \\
0.54054054054054	0.344717316188045 \\
0.548548548548548	0.343217374289848 \\
0.556556556556557	0.341702045536168 \\
0.564564564564565	0.340171591776383 \\
0.572572572572573	0.338626276598684 \\
0.58058058058058	0.337066365254824 \\
0.588588588588588	0.335492124584683 \\
0.596596596596597	0.333903822940662 \\
0.604604604604605	0.33230173011195 \\
0.612612612612613	0.330686117248681 \\
0.62062062062062	0.329057256786028 \\
0.628628628628628	0.327415422368237 \\
0.636636636636637	0.325760888772662 \\
0.644644644644645	0.324093931833804 \\
0.652652652652653	0.322414828367396 \\
0.66066066066066	0.32072385609456 \\
0.668668668668668	0.319021293566066 \\
0.676676676676677	0.317307420086723 \\
0.684684684684685	0.31558251563992 \\
0.692692692692693	0.313846860812361 \\
0.7007007007007	0.312100736719007 \\
0.708708708708708	0.310344424928268 \\
0.716716716716717	0.308578207387456 \\
0.724724724724725	0.306802366348536 \\
0.732732732732733	0.305017184294205 \\
0.74074074074074	0.303222943864308 \\
0.748748748748748	0.301419927782648 \\
0.756756756756757	0.299608418784177 \\
0.764764764764765	0.29778869954263 \\
0.772772772772773	0.295961052598604 \\
0.780780780780781	0.294125760288111 \\
0.788788788788788	0.292283104671645 \\
0.796796796796797	0.290433367463758 \\
0.804804804804805	0.288576829963198 \\
0.812812812812813	0.286713772983616 \\
0.820820820820821	0.284844476784867 \\
0.828828828828828	0.282969221004931 \\
0.836836836836837	0.281088284592474 \\
0.844844844844845	0.279201945740075 \\
0.852852852852853	0.277310481818125 \\
0.860860860860861	0.275414169309444 \\
0.868868868868869	0.273513283744617 \\
0.876876876876877	0.271608099638066 \\
0.884884884884885	0.269698890424904 \\
0.892892892892893	0.267785928398552 \\
0.900900900900901	0.265869484649172 \\
0.908908908908909	0.263949829002911 \\
0.916916916916917	0.262027229961987 \\
0.924924924924925	0.260101954645624 \\
0.932932932932933	0.258174268731856 \\
0.940940940940941	0.256244436400228 \\
0.948948948948949	0.254312720275384 \\
0.956956956956957	0.252379381371581 \\
0.964964964964965	0.250444679038131 \\
0.972972972972973	0.248508870905785 \\
0.980980980980981	0.246572212834085 \\
0.988988988988989	0.244634958859672 \\
0.996996996996997	0.242697361145593 \\
1.00500500500501	0.24075966993159 \\
1.01301301301301	0.238822133485401 \\
1.02102102102102	0.236884998055083 \\
1.02902902902903	0.234948507822352 \\
1.03703703703704	0.233012904856966 \\
1.04504504504505	0.231078429072154 \\
1.05305305305305	0.229145318181096 \\
1.06106106106106	0.22721380765447 \\
1.06906906906907	0.225284130679062 \\
1.07707707707708	0.223356518117463 \\
1.08508508508509	0.221431198468836 \\
1.09309309309309	0.219508397830784 \\
1.1011011011011	0.217588339862305 \\
1.10910910910911	0.215671245747847 \\
1.11711711711712	0.21375733416247 \\
1.12512512512513	0.21184682123811 \\
1.13313313313313	0.209939920530956 \\
1.14114114114114	0.208036842989938 \\
1.14914914914915	0.206137796926331 \\
1.15715715715716	0.204242987984477 \\
1.16516516516517	0.202352619113618 \\
1.17317317317317	0.200466890540856 \\
1.18118118118118	0.198585999745228 \\
1.18918918918919	0.19671014143289 \\
1.1971971971972	0.194839507513435 \\
1.20520520520521	0.192974287077311 \\
1.21321321321321	0.191114666374361 \\
1.22122122122122	0.18926082879347 \\
1.22922922922923	0.187412954843324 \\
1.23723723723724	0.185571222134271 \\
1.24524524524525	0.183735805361283 \\
1.25325325325325	0.181906876288021 \\
1.26126126126126	0.180084603731981 \\
1.26926926926927	0.178269153550739 \\
1.27727727727728	0.176460688629273 \\
1.28528528528529	0.174659368868355 \\
1.29329329329329	0.172865351174024 \\
1.3013013013013	0.17107878944811 \\
1.30930930930931	0.169299834579821 \\
1.31731731731732	0.167528634438376 \\
1.32532532532533	0.165765333866673 \\
1.33333333333333	0.164010074675994 \\
1.34134134134134	0.16226299564173 \\
1.34934934934935	0.160524232500121 \\
1.35735735735736	0.158793917945997 \\
1.36536536536537	0.157072181631514 \\
1.37337337337337	0.155359150165875 \\
1.38138138138138	0.153654947116025 \\
1.38938938938939	0.151959693008306 \\
1.3973973973974	0.150273505331067 \\
1.40540540540541	0.148596498538208 \\
1.41341341341341	0.146928784053658 \\
1.42142142142142	0.14527047027677 \\
1.42942942942943	0.143621662588612 \\
1.43743743743744	0.141982463359159 \\
1.44544544544545	0.140352971955359 \\
1.45345345345345	0.13873328475006 \\
1.46146146146146	0.137123495131801 \\
1.46946946946947	0.13552369351543 \\
1.47747747747748	0.133933967353555 \\
1.48548548548549	0.132354401148798 \\
1.49349349349349	0.130785076466857 \\
1.5015015015015	0.129226071950339 \\
1.50950950950951	0.127677463333369 \\
1.51751751751752	0.12613932345695 \\
1.52552552552553	0.124611722285057 \\
1.53353353353353	0.123094726921466 \\
1.54154154154154	0.121588401627275 \\
1.54954954954955	0.120092807839133 \\
1.55755755755756	0.11860800418814 \\
1.56556556556557	0.117134046519405 \\
1.57357357357357	0.115670987912262 \\
1.58158158158158	0.114218878701105 \\
1.58958958958959	0.112777766496842 \\
1.5975975975976	0.111347696208955 \\
1.60560560560561	0.10992871006813 \\
1.61361361361361	0.108520847649465 \\
1.62162162162162	0.107124145896225 \\
1.62962962962963	0.105738639144131 \\
1.63763763763764	0.10436435914617 \\
1.64564564564565	0.103001335097907 \\
1.65365365365365	0.10164959366328 \\
1.66166166166166	0.100309159000872 \\
1.66966966966967	0.0989800527906321 \\
1.67767767767768	0.0976622942610365 \\
1.68568568568569	0.0963559002166692 \\
1.69369369369369	0.0950608850662112 \\
1.7017017017017	0.0937772608508176 \\
1.70970970970971	0.0925050372728685 \\
1.71771771771772	0.0912442217250772 \\
1.72572572572573	0.0899948193199405 \\
1.73373373373373	0.088756832919514 \\
1.74174174174174	0.0875302631654974 \\
1.74974974974975	0.086315108509615 \\
1.75775775775776	0.0851113652442723 \\
1.76576576576577	0.0839190275334778 \\
1.77377377377377	0.082738087444011 \\
1.78178178178178	0.0815685349768226 \\
1.78978978978979	0.0804103580986525 \\
1.7977977977978	0.0792635427738473 \\
1.80580580580581	0.0781280729963669 \\
1.81381381381381	0.0770039308219615 \\
1.82182182182182	0.0758910964005057 \\
1.82982982982983	0.0747895480084762 \\
1.83783783783784	0.0736992620815557 \\
1.84584584584585	0.0726202132473522 \\
1.85385385385385	0.0715523743582165 \\
1.86186186186186	0.0704957165241465 \\
1.86986986986987	0.0694502091457625 \\
1.87787787787788	0.0684158199473405 \\
1.88588588588589	0.0673925150098903 \\
1.89389389389389	0.0663802588042655 \\
1.9019019019019	0.0653790142242912 \\
1.90990990990991	0.0643887426198963 \\
1.91791791791792	0.06340940383024 \\
1.92592592592593	0.0624409562168171 \\
1.93393393393393	0.0614833566965309 \\
1.94194194194194	0.0605365607747232 \\
1.94994994994995	0.0596005225781457 \\
1.95795795795796	0.0586751948878651 \\
1.96596596596597	0.0577605291720876 \\
1.97397397397397	0.056856475618893 \\
1.98198198198198	0.0559629831688668 \\
1.98998998998999	0.0550799995476186 \\
1.997997997998	0.0542074712981785 \\
2.00600600600601	0.0533453438132581 \\
2.01401401401401	0.0524935613673681 \\
2.02202202202202	0.0516520671487823 \\
2.03003003003003	0.0508208032913361 \\
2.03803803803804	0.0499997109060533 \\
2.04604604604605	0.0491887301125894 \\
2.05405405405405	0.0483878000704834 \\
2.06206206206206	0.0475968590102091 \\
2.07007007007007	0.0468158442640166 \\
2.07807807807808	0.0460446922965577 \\
2.08608608608609	0.0452833387352837 \\
2.09409409409409	0.0445317184006117 \\
2.1021021021021	0.043789765335848 \\
2.11011011011011	0.0430574128368641 \\
2.11811811811812	0.0423345934815158 \\
2.12612612612613	0.0416212391588004 \\
2.13413413413413	0.0409172810977437 \\
2.14214214214214	0.0402226498960119 \\
2.15015015015015	0.03953727554824 \\
2.15815815815816	0.0388610874740724 \\
2.16616616616617	0.03819401454591 \\
2.17417417417417	0.0375359851163567 \\
2.18218218218218	0.0368869270453615 \\
2.19019019019019	0.0362467677270497 \\
2.1981981981982	0.0356154341162401 \\
2.20620620620621	0.0349928527546413 \\
2.21421421421421	0.0343789497967251 \\
2.22222222222222	0.0337736510352706 \\
2.23023023023023	0.0331768819265761 \\
2.23823823823824	0.0325885676153351 \\
2.24624624624625	0.0320086329591724 \\
2.25425425425425	0.0314370025528369 \\
2.26226226226226	0.0308736007520491 \\
2.27027027027027	0.0303183516969979 \\
2.27827827827828	0.029771179335487 \\
2.28628628628629	0.0292320074457259 \\
2.29429429429429	0.0287007596587648 \\
2.3023023023023	0.0281773594805703 \\
2.31031031031031	0.0276617303137407 \\
2.31831831831832	0.0271537954788581 \\
2.32632632632633	0.0266534782354776 \\
2.33433433433433	0.0261607018027501 \\
2.34234234234234	0.0256753893796796 \\
2.35035035035035	0.0251974641650118 \\
2.35835835835836	0.0247268493767558 \\
2.36636636636637	0.0242634682713357 \\
2.37437437437437	0.023807244162374 \\
2.38238238238238	0.0233581004391047 \\
2.39039039039039	0.022915960584417 \\
2.3983983983984	0.0224807481925294 \\
2.40640640640641	0.0220523869862944 \\
2.41441441441441	0.0216308008341344 \\
2.42242242242242	0.0212159137666093 \\
2.43043043043043	0.0208076499926159 \\
2.43843843843844	0.020405933915221 \\
2.44644644644645	0.0200106901471285 \\
2.45445445445445	0.0196218435257817 \\
2.46246246246246	0.0192393191281025 \\
2.47047047047047	0.0188630422848682 \\
2.47847847847848	0.0184929385947284 \\
2.48648648648649	0.0181289339378626 \\
2.49449449449449	0.0177709544892815 \\
2.5025025025025	0.0174189267317727 \\
2.51051051051051	0.0170727774684938 \\
2.51851851851852	0.0167324338352152 \\
2.52652652652653	0.016397823312213 \\
2.53453453453453	0.0160688737358178 \\
2.54254254254254	0.0157455133096184 \\
2.55055055055055	0.0154276706153247 \\
2.55855855855856	0.0151152746232934 \\
2.56656656656657	0.0148082547027168 \\
2.57457457457457	0.0145065406314803 \\
2.58258258258258	0.014210062605689 \\
2.59059059059059	0.0139187512488692 \\
2.5985985985986	0.0136325376208454 \\
2.60660660660661	0.0133513532262973 \\
2.61461461461461	0.0130751300230007 \\
2.62262262262262	0.0128038004297541 \\
2.63063063063063	0.0125372973339956 \\
2.63863863863864	0.0122755540991133 \\
2.64664664664665	0.0120185045714526 \\
2.65465465465465	0.0117660830870247 \\
2.66266266266266	0.0115182244779185 \\
2.67067067067067	0.0112748640784219 \\
2.67867867867868	0.011035937730854 \\
2.68668668668669	0.0108013817911136 \\
2.69469469469469	0.0105711331339476 \\
2.7027027027027	0.0103451291579421 \\
2.71071071071071	0.0101233077902421 \\
2.71871871871872	0.00990560749100247 \\
2.72672672672673	0.00969196725757422 \\
2.73473473473473	0.00948232662843099 \\
2.74274274274274	0.00927662568683898 \\
2.75075075075075	0.00907480506427511 \\
2.75875875875876	0.00887680594359712 \\
2.76676676676677	0.00868257006197001 \\
2.77477477477477	0.00849203971355293 \\
2.78278278278278	0.00830515775195071 \\
2.79079079079079	0.00812186759243432 \\
2.7987987987988	0.00794211321393438 \\
2.80680680680681	0.0077658391608121 \\
2.81481481481481	0.00759299054441176 \\
2.82282282282282	0.00742351304439893 \\
2.83083083083083	0.00725735290988893 \\
2.83883883883884	0.00709445696036947 \\
2.84684684684685	0.0069347725864219 \\
2.85485485485485	0.0067782477502453 \\
2.86286286286286	0.00662483098598741 \\
2.87087087087087	0.00647447139988703 \\
2.87887887887888	0.00632711867023165 \\
2.88688688688689	0.00618272304713484 \\
2.89489489489489	0.00604123535213746 \\
2.9029029029029	0.00590260697763674 \\
2.91091091091091	0.0057667898861475 \\
2.91891891891892	0.00563373660939975 \\
2.92692692692693	0.00550340024727644 \\
2.93493493493493	0.00537573446659573 \\
2.94294294294294	0.00525069349974172 \\
2.95095095095095	0.00512823214314764 \\
2.95895895895896	0.00500830575563558 \\
2.96696696696697	0.00489087025661667 \\
2.97497497497497	0.00477588212415564 \\
2.98298298298298	0.00466329839290368 \\
2.99099099099099	0.00455307665190356 \\
2.998998998999	0.00444517504227067 \\
3.00700700700701	0.00433955225475399 \\
3.01501501501502	0.00423616752718052 \\
3.02302302302302	0.0041349806417872 \\
3.03103103103103	0.00403595192244368 \\
3.03903903903904	0.00393904223176991 \\
3.04704704704705	0.00384421296815189 \\
3.05505505505506	0.00375142606265932 \\
3.06306306306306	0.00366064397586864 \\
3.07107107107107	0.00357182969459489 \\
3.07907907907908	0.00348494672853594 \\
3.08708708708709	0.00339995910683238 \\
3.0950950950951	0.00331683137454643 \\
3.1031031031031	0.00323552858906328 \\
3.11111111111111	0.00315601631641805 \\
3.11911911911912	0.00307826062755151 \\
3.12712712712713	0.00300222809449793 \\
3.13513513513514	0.00292788578650791 \\
3.14314314314314	0.00285520126610952 \\
3.15115115115115	0.00278414258511062 \\
3.15915915915916	0.00271467828054533 \\
3.16716716716717	0.00264677737056774 \\
3.17517517517518	0.00258040935029545 \\
3.18318318318318	0.00251554418760605 \\
3.19119119119119	0.00245215231888919 \\
3.1991991991992	0.00239020464475689 \\
3.20720720720721	0.00232967252571501 \\
3.21521521521522	0.00227052777779815 \\
3.22322322322322	0.00221274266817093 \\
3.23123123123123	0.00215628991069792 \\
3.23923923923924	0.00210114266148476 \\
3.24724724724725	0.00204727451439301 \\
3.25525525525526	0.00199465949653091 \\
3.26326326326326	0.00194327206372255 \\
3.27127127127127	0.00189308709595757 \\
3.27927927927928	0.00184407989282385 \\
3.28728728728729	0.00179622616892505 \\
3.2952952952953	0.00174950204928538 \\
3.3033033033033	0.00170388406474362 \\
3.31131131131131	0.00165934914733832 \\
3.31931931931932	0.00161587462568624 \\
3.32732732732733	0.00157343822035606 \\
3.33533533533534	0.00153201803923895 \\
3.34334334334334	0.0014915925729182 \\
3.35135135135135	0.00145214069003938 \\
3.35935935935936	0.001413641632683 \\
3.36736736736737	0.00137607501174126 \\
3.37537537537538	0.00133942080230046 \\
3.38338338338338	0.00130365933903089 \\
3.39139139139139	0.00126877131158549 \\
3.3993993993994	0.00123473776000907 \\
3.40740740740741	0.00120154007015926 \\
3.41541541541542	0.00116915996914087 \\
3.42342342342342	0.00113757952075483 \\
3.43143143143143	0.00110678112096324 \\
3.43943943943944	0.0010767474933716 \\
3.44744744744745	0.00104746168472971 \\
3.45545545545546	0.0010189070604522 \\
3.46346346346346	0.000991067300160061 \\
3.47147147147147	0.000963926393244148 \\
3.47947947947948	0.000937468634451866 \\
3.48748748748749	0.00091167861949796 \\
3.4954954954955	0.000886541240700502 \\
3.5035035035035	0.000862041682643008 \\
3.51151151151151	0.000838165417863601 \\
3.51951951951952	0.00081489820257214 \\
3.52752752752753	0.000792226072396121 \\
3.53553553553554	0.000770135338156225 \\
3.54354354354354	0.000748612581672247 \\
3.55155155155155	0.000727644651600165 \\
3.55955955955956	0.000707218659301081 \\
3.56756756756757	0.000687321974742665 \\
3.57557557557558	0.000667942222433796 \\
3.58358358358358	0.000649067277392974 \\
3.59159159159159	0.000630685261151104 \\
3.5995995995996	0.000612784537789192 \\
3.60760760760761	0.000595353710011453 \\
3.61561561561562	0.00057838161525435 \\
3.62362362362362	0.000561857321831992 \\
3.63163163163163	0.000545770125118335 \\
3.63963963963964	0.000530109543766577 \\
3.64764764764765	0.000514865315966112 \\
3.65565565565566	0.000500027395737409 \\
3.66366366366366	0.000485585949265104 \\
3.67167167167167	0.000471531351269623 \\
3.67967967967968	0.000457854181417586 \\
3.68768768768769	0.00044454522077123 \\
3.6956956956957	0.00043159544827708 \\
3.7037037037037	0.000418996037294059 \\
3.71171171171171	0.000406738352161196 \\
3.71971971971972	0.000394813944805093 \\
3.72772772772773	0.000383214551387261 \\
3.73573573573574	0.000371932088991452 \\
3.74374374374374	0.000360958652351047 \\
3.75175175175175	0.000350286510616579 \\
3.75975975975976	0.000339908104163431 \\
3.76776776776777	0.000329816041439723 \\
3.77577577577578	0.000320003095854415 \\
3.78378378378378	0.0003104622027056 \\
3.79179179179179	0.000301186456148956 \\
3.7997997997998	0.000292169106206322 \\
3.80780780780781	0.000283403555814325 \\
3.81581581581582	0.000274883357912987 \\
3.82382382382382	0.000266602212574214 \\
3.83183183183183	0.000258553964170059 \\
3.83983983983984	0.000250732598580649 \\
3.84784784784785	0.000243132240441603 \\
3.85585585585586	0.000235747150430845 \\
3.86386386386386	0.000228571722594601 \\
3.87187187187187	0.000221600481712426 \\
3.87987987987988	0.000214828080701088 \\
3.88788788788789	0.000208249298057069 \\
3.8958958958959	0.000201859035337517 \\
3.9039039039039	0.000195652314679408 \\
3.91191191191191	0.000189624276356684 \\
3.91991991991992	0.000183770176375138 \\
3.92792792792793	0.00017808538410479 \\
3.93593593593594	0.000172565379949497 \\
3.94394394394394	0.00016720575305353 \\
3.95195195195195	0.00016200219904485 \\
3.95995995995996	0.000156950517814782 \\
3.96796796796797	0.000152046611333821 \\
3.97597597597598	0.000147286481503255 \\
3.98398398398398	0.000142666228042307 \\
3.99199199199199	0.000138182046410498 \\
4	0.000133830225764885 \\
};
\end{axis}

\end{tikzpicture}

%% file: content/4-Conclusion.tex
\section{Conclusion}
In this paper, we introduced the \emph{Recurrent Kalman Network} that jointly learns high-dimensional representations of the system in a latent space with locally linear transition models and factorized covariances.
The update equations in the high-dimensional space are based on the update equations of the classical Kalman filter, however, due to the factorization assumptions, they simplify to scalar operations that can be performed much faster and with greater numerical stability.
Our model outperforms generic LSTMs and GRUs on various state estimation tasks while providing reasonable uncertainty estimates. Additionally, it outperformed several generative models on an image imputation task and we demonstrated its applicability to real world scenarios. Training is straight forward and can be done in an end-to-end fashion.

In future work, we want to exploit the principled notion of a variance provided by our approach in scenarios where such a notion is beneficial, e.g. reinforcement learning. 
Similar to \cite{fraccaro2017kvae} we could expand our approach to not just filter but smooth over trajectories in offline post-processing scenarios which could potentially increase the estimation performance significantly.
Additionally, different, more complex, factorization assumptions can be investigated in the future.

%% file: content/A-Preliminaries.tex
The Kalman filter \citep{kalman1960filter} works by iteratively applying two steps, predict and update. It assumes additive Gaussian noise with zero mean and  covariances $\cmat{\Sigma}^\textrm{trans}$ and $\cmat{\Sigma}^\textrm{obs}$ on both transitions and observations, which need to be given to the filter.
During the prediction step the transition model $\cmat{A}$ is used to infer the next prior state estimate $\left(\cvec{x}_{t+1}^-, \cmat{\Sigma}_{t+1}^- \right)$, i.e., a-priori to the observation, from the current posterior estimate $\left(\cvec{x}_{t}^+, \cmat{\Sigma}_{t}^+ \right)$,  by 
\begin{align*}
\cvec{x}^-_{t+1} &= \cmat{A} \cvec{x}^+_{t} \\ \text{and} \quad \cmat{\Sigma}^-_{t+1} &= \cmat{A} \cmat{\Sigma}_{t}^+ \cmat{A}^T + \cmat{\Sigma}^\mathrm{trans}.
\end{align*}
The prior estimate is then updated using the current observation $\cvec{w_t}$ and the observation model $\cmat{H}$ to obtain the posterior estimate $\left(\cvec{x}_t^+, \cmat{\Sigma}_t^+ \right)$, i.e.,
\begin{align*}
\cvec{x}_t^+ &= \cvec{x}_t^- + \cmat{Q}_t \left(\cvec{w}_t - \cmat{H} \cvec{x}_t^-  \right) \\ \text{and} \quad   \cmat{\Sigma}^+_t &= \left(\cmat{I} - \cmat{Q}_t \cmat{H} \right) \cmat{\Sigma}^-_t \\ \text{, with} \quad \cmat{Q}_t &= \cmat{\Sigma}^-_t \cmat{H}^T\left(\cmat{H} \cmat{\Sigma}^-_t \cmat{H}^T  + \cmat{\Sigma}^\mathrm{obs} \right)^{-1}, 
\end{align*}
where $\cmat{I}$ denotes the identity matrix.
The matrix $\cmat{Q}_t$ is referred to as the Kalman gain.
The whole update step can be interpreted as a weighted average between state and observation estimate, where the weighting, i.e., $\cmat{Q}_t$, depends on the uncertainty about those estimates. 

If currently no observation is present or future states should be predicted, the update step is omitted.

%% file: content/B-Appendix.tex
\renewcommand\thesubsection{\Alph{subsection}}
\newcommand{\cdiag}[1]{\boldsymbol{\hat{\mathrm{#1}}}}
\setcounter{subsection}{0}
\subsection{Simplified Kalman Filter Formulas}
As stated above the simple latent observation model  $\cmat{H} = \left[\begin{array}{cc} \cmat{I}_m & \cmat{0}_{m \times (n-m)} \end{array}\right]$, as well as the assumed factorization of the covariance matrices allow us to simplify the Kalman Filter equations. 
\subsubsection{Notation}
In the following derivations we neglect the time indices $t$ and $t+1$ for brevity.
For any matrix $\cmat{M}$, $\cdiag{M}$ denotes a diagonal matrix with the same diagonal as $\cmat{M}$, $\cvec{m}$ denotes a vector containing those diagonal elements and $M^{(ij)}$ denotes the entry at row $i$ and column $j$. Similarly, $v^{(i)}$ denotes the $i$-th entry of a vector $\cvec{v}$. The point wise product between two vectors of same length (Hadamat Product) will be denoted by $\odot$ and the point wise division by $\oslash$.
\subsubsection{Prediction Step}
\begin{align*}
&\text{Mean:} \quad &\cvec{z}^- &= \cmat{A} \cvec{z}^+\\
&\text{Covariance:} \quad & \cmat{\Sigma}^- &= \cmat{A} \cmat{\Sigma}^+ \cmat{A}^T + \cdiag{\Sigma}^\mathrm{trans}
\end{align*}
The computation of the mean can not be further simplified, however, depending on the state size and bandwidth, sparse matrix multiplications may be exploited. For the covariance, let $\cmat{T} = \cmat{A}\cmat{\Sigma}^+$. Then, 
\begin{align*}
\cmat{T} &= \cmat{A} \cmat{\Sigma}^+ = 
\left[\begin{array}{cc}
\cmat{B}_{11} & \cmat{B}_{12} \\
\cmat{B}_{21} & \cmat{B}_{22}
\end{array}\right] 
\left[\begin{array}{cc} \cdiag{\Sigma}^{\mathrm{u},+} & \cdiag{\Sigma}^{\mathrm{s},+} \\ 
\cdiag{\Sigma}^{\mathrm{s},+} & \cdiag{\Sigma}^{\mathrm{l},+} 
\end{array}\right] \\ &= 
\left[\begin{array}{cc}
\cmat{B}_{11}\cdiag{\Sigma}^{\mathrm{u},+} + \cmat{B}_{12}\cdiag{\Sigma}^{\mathrm{s},+} & \cmat{B}_{11}\cdiag{\Sigma}^{\mathrm{s},+} + \cmat{B}_{12} \cdiag{\Sigma}^{\mathrm{l},+} \\
\cmat{B}_{21}\cdiag{\Sigma}^{\mathrm{u},+} + \cmat{B}_{22}\cdiag{\Sigma}^{\mathrm{s},+} & \cmat{B}_{21}\cdiag{\Sigma}^{\mathrm{s},+} + \cmat{B}_{22} \cdiag{\Sigma}^{\mathrm{l},+}
\end{array}\right] 
\end{align*}
and
\begin{align*}
\cmat{\Sigma}^- &= \cmat{T} \cmat{A}^T + \cdiag{\Sigma}^\mathrm{trans} \\ =& 
\left[\begin{array}{cc}
\cmat{B}_{11}\cdiag{\Sigma}^{\mathrm{u},+} + \cmat{B}_{12}\cdiag{\Sigma}^{\mathrm{s},+} & \cmat{B}_{11}\cdiag{\Sigma}^{\mathrm{s},+} + \cmat{B}_{12} \cdiag{\Sigma}^{\mathrm{l},+} \\
\cmat{B}_{21}\cdiag{\Sigma}^{\mathrm{u},+} + \cmat{B}_{22}\cdiag{\Sigma}^{\mathrm{s},+} & \cmat{B}_{21}\cdiag{\Sigma}^{\mathrm{s},+} + \cmat{B}_{22} \cdiag{\Sigma}^{\mathrm{l},+}
\end{array}\right] \dots \\
\dots & 
\left[\begin{array}{cc}
\cmat{B}_{11}^T & \cmat{B}_{21}^T \\
\cmat{B}_{12}^T & \cmat{B}_{22}^T
\end{array}\right]  + \cdiag{\Sigma}^\mathrm{trans} =  
\left[\begin{array}{cc}
\cdiag{\Sigma}^{\mathrm{u},-} & \cdiag{\Sigma}^{\mathrm{s},-} \\ 
\cdiag{\Sigma}^{\mathrm{s},-} & \cdiag{\Sigma}^{\mathrm{l},-} 
\end{array}\right] ,
\end{align*}
with 
\begin{align*}
\cmat{\Sigma}^{\mathrm{u},-} 
=&\left(\cmat{B}_{11} \cdiag{\Sigma}^{\mathrm{u},+} + \cmat{B}_{12} \cdiag{\Sigma}^{\mathrm{s},+} \right) \cmat{B}_{11}^T \dots \\ \dots  &+ \left(\cmat{B}_{11}  \cdiag{\Sigma}^{\mathrm{s},+} + \cmat{B}_{12} \cdiag{\Sigma}^{\mathrm{l},+} \right) \cmat{B}_{12}^T   + \cdiag{\Sigma}^{\mathrm{u}, \mathrm{trans}}\\
=&\cmat{B}_{11} \cdiag{\Sigma}^{\mathrm{u},+} \cmat{B}_{11}^T + \cmat{B}_{12} \cdiag{\Sigma}^{\mathrm{s},+} \cmat{B}_{11}^T  \dots  \\ \dots &+ \cmat{B}_{11} \cdiag{\Sigma}^{\mathrm{s},+} \cmat{B}_{12}^T + \cmat{B}_{12} \cdiag{\Sigma}^{\mathrm{l},+} \cmat{B}_{12}^T + \cdiag{\Sigma}^{\mathrm{u}, \mathrm{trans}}\\
\cmat{\Sigma}^{\mathrm{l},-} =&\left(\cmat{B}_{21} \cdiag{\Sigma}^{\mathrm{u},+} + \cmat{B}_{22} \cdiag{\Sigma}^{\mathrm{s},+} \right) \cmat{B}_{21}^T  \dots \\ \dots  &+ \left(\cmat{B}_{21}\cdiag{\Sigma}^{\mathrm{s},+} + \cmat{B}_{22} \cdiag{\Sigma}^{\mathrm{l},+} \right) \cmat{B}_{22}^T + \cdiag{\Sigma}^{\mathrm{l}, \mathrm{trans}}\\
=&\cmat{B}_{21} \cdiag{\Sigma}^{\mathrm{u},+} \cmat{B}_{21}^T + \cmat{B}_{22} \cdiag{\Sigma}^{\mathrm{s},+} \cmat{B}_{21}^T \dots \\ \dots  &+ \cmat{B}_{21}\cdiag{\Sigma}^{\mathrm{s},+}  \cmat{B}_{22}^T + \cmat{B}_{22} \cdiag{\Sigma}^{\mathrm{l},+} \cmat{B}_{22}^T + \cdiag{\Sigma}^{\mathrm{l}, \mathrm{trans}}\\
\cmat{\Sigma}^{\mathrm{s},-} 
=&\left(\cmat{B}_{21} \cdiag{\Sigma}^{\mathrm{u},+} + \cmat{B}_{22} \cdiag{\Sigma}^{\mathrm{s},+} \right) \cmat{B}_{11}^T  \dots \\ \dots  &+ \left(\cmat{B}_{21}\cdiag{\Sigma}^{\mathrm{s},+} + \cmat{B}_{22} \cdiag{\Sigma}^{\mathrm{l},+} \right) \cmat{B}_{12}^T\\
=&\cmat{B}_{21} \cdiag{\Sigma}^{\mathrm{u},+} \cmat{B}_{11}^T + \cmat{B}_{22} \cdiag{\Sigma}^{\mathrm{s},+} \cmat{B}_{11}^T \dots \\ \dots  &+ \cmat{B}_{21}\cdiag{\Sigma}^{\mathrm{s},+}  \cmat{B}_{12}^T + \cmat{B}_{22} \cdiag{\Sigma}^{\mathrm{l},+} \cmat{B}_{12}^T 
\end{align*}
Since we are only interested in the diagonal parts of $\cmat{\Sigma}^{\mathrm{u},- }, \cmat{\Sigma}^{\mathrm{l},- }$ and $\cmat{\Sigma}^{\mathrm{s},- }$ i.e. $\cdiag{\Sigma}^{\mathrm{u},- }, \cdiag{\Sigma}^{\mathrm{l},- }$ and $\cdiag{\Sigma}^{\mathrm{s},- }$, we can further simplify these equations by realizing two properties of the terms above. First, for any matrix $\cmat{M}, \cmat{N}$ and a diagonal matrix $\cdiag{\Sigma}$ it holds that
$$\left(\cmat{M} \cdiag{\Sigma} \cmat{N}^T\right)^{(ii)} = \sum_{k=1}^n A^{(ik)} B^{(ik)} \sigma^{(k)} = \left(\cmat{N} \cdiag{\Sigma} \cmat{M}^T\right)_{ii}.$$ 
Hence, we can simplify the  equations for the upper and lower part to
\begin{align*}
\cmat{\Sigma}^{u,-} 
=&\cmat{B}_{11} \cdiag{\Sigma}^{\mathrm{u},+} \cmat{B}_{11}^T + 2 \cdot \cmat{B}_{12} \cdiag{\Sigma}^{\mathrm{s},+} \cmat{B}_{11}^T \dots \\ \dots  &+ \cmat{B}_{12} \cdiag{\Sigma}^{\mathrm{l},+} \cmat{B}_{12}^T + \cdiag{\Sigma}^{\mathrm{u}, \textrm{trans}},\\
\cmat{\Sigma}^{\mathrm{l},-} 
=&\cmat{B}_{21} \cdiag{\Sigma}^{\mathrm{u},+} \cmat{B}_{21}^T +  2 \cdot \cmat{B}_{22} \cdiag{\Sigma}^{\mathrm{s},+} \cmat{B}_{21}^T \dots \\ \dots  &+ \cmat{B}_{22} \cdiag{\Sigma}^{\mathrm{l},+} \cmat{B}_{22}^T + \cdiag{\Sigma}^{\mathrm{l}, \mathrm{trans}}.\\
\end{align*}
Second, since we are only interested in the diagonal of the result it is sufficient to compute only the diagonals of the individual parts of the sums which are almost all of the same structure i.e. $\cmat{S} = \cmat{M} \cdiag{\Sigma}^+ \cmat{N}^T $. Let $\cmat{T} = \cmat{M} \cdiag{\Sigma}^+$, then each element of $\cmat{T}$ can be computed as 
$$ T^{(ij)} = \sum_{k=1}^n M^{(ik)} \hat{\Sigma}^{(kj)} = M^{(ij)} \sigma^{(j)}.$$
Consequently, the elements of $\cmat{S}  = \cmat{T} \cmat{A}^T$ can be computed as
$$ S^{(ij)} = \sum_{k=1}^n T^{(ik)} A^{(kj)} = \sum_{k=1}^n M^{(ik)} \sigma_k N^{(jk)}. $$
Ultimately, we are not interested in $\cmat{S}$ but only in $\cdiag{S}$ $$\hat{S}^{(ii)} = \sum_{k=1}^n M^{(ik)} N^{(ik)} \sigma^(k).$$ Using this we obtain can obtain the entries of $\cvec{\sigma}^{u,-}, \cvec{\sigma}^{l,-}$ and $\cvec{\sigma}^{s,-}$ by
\begin{align}
\sigma^{u,-, (i)} &= \label{eq:var1_pred} \\  \sum_{k=1}^m & \left(B_{11}^{(ik)}\right)^2 \sigma^{u,+, (i)} + 2 \sum_{k=1}^m B_{11}^{(ik)} B_{12}^{(ik)} \sigma^{s,+, (i)} ... \nonumber \\ ... + \sum_{k=1}^m & \left(B_{12}^{(ik)}\right)^2 \sigma^{l,+, (i)} +  \sigma^{u, \textrm{trans}, (i)}  \nonumber \\
\sigma^{l,-, (i)} &= \label{eq:var2_pred} \\  \sum_{k=1}^m & \left(B_{21}^{(ik)}\right) \sigma^{u,+, (i)} + 2 \sum_{k=1}^m B_{22}^{(ik)} B_{21}^{(ik)} \sigma^{s,+, (i)}  ... \nonumber \\ ... + \sum_{k=1}^m & \left(B_{22}^{(ik)}\right)^2 \sigma^{l,+, (i)} +  \sigma^{l, \textrm{trans}, (i)} \nonumber
\\
\sigma^{s,-, (i)}  &= \label{eq:var3_pred} \\ \sum_{k=1}^m & B_{21}^{(ik)} B_{11}^{(ik)} \sigma^{u,+, (i)} +  \sum_{k=1}^m B_{22}^{(ik)} B_{11}^{(ik)} \sigma^{s,+, (i)} \text{...} \nonumber  \\  \text{...} +  \sum_{k=1}^m & B_{21}^{(ik)} B_{12}^{(ik)} \sigma^{s,+, (i)}  + \sum_{k=1}^m B_{22}^{(ik)} B_{12}^{(ik)} \sigma^{l,+, (i)}, \nonumber
\end{align}
which can be implemented efficiently using elementwise matrix multiplication and sum reduction.
\subsubsection{Update Step}
\begin{align*}
&\text{Kalman Gain} \quad &\cmat{Q} &= \cmat{\Sigma}^- \cmat{H}^T\left(\cmat{H} \cmat{\Sigma}^- \cmat{H}^T  + \cmat{\Sigma}^\mathrm{obs} \right)^{-1} \\
&\text{Mean} \quad &\cvec{z}^+ &= \cvec{z}^- + \cmat{Q} \left(\cvec{w} - \cmat{H} \cvec{z}^-  \right) \\
&\text{Covariance} \quad &\cmat{\Sigma}^+ &= \left(\cmat{I} - \cmat{Q} \cmat{H} \right) \cmat{\Sigma}^- 
\end{align*}
First, note that 
$$\cmat{\Sigma}^- \cmat{H}^T = \left[\begin{array}{c} \cdiag{\Sigma}^{\mathrm{u},-} \\ \cdiag{\Sigma}^{\mathrm{s},-} \end{array}\right] $$ and $$\cmat{H} \cmat{\Sigma}^- \cmat{H}^T  + \cdiag{\Sigma}^\mathrm{obs} = \cdiag{\Sigma}^{u,-} + \cdiag{\Sigma}^\mathrm{obs}
$$
and thus the computation of the Kalman Gain only involves diagonal matrices. Hence the Kalman Gain matrix also consists of two diagonal matrices, i.e., $\cmat{Q} =
\left[\begin{array}{c}\cdiag{Q}^\mathrm{u} \\
\cdiag{Q}^\mathrm{l}
\end{array}\right] $ whose diagonals can be computed by 

\begin{align}
\cvec{q}^\mathrm{u} &= \cvec{\sigma}^{\mathrm{u},-} \oslash \left( \cvec{\sigma}^{\mathrm{u},-} + \cvec{\sigma}^\mathrm{obs}  \right) \\ \text{and} \quad \cvec{q}^\mathrm{l} &= \cvec{\sigma}^{\mathrm{s},-} \oslash \left( \cmat{\sigma}^{\mathrm{u},-} + \cmat{\sigma}^\mathrm{obs} \right).\end{align}

Using this result, the mean update can be simplified to 
\begin{equation}\cvec{z}^+ = \cvec{z}^- +
\left[\begin{array}{cc}
\cmat{q}_\mathrm{u} \\
\cmat{q}_\mathrm{l}
\end{array}\right]\odot
\left[\begin{array}{cc}
\cvec{w} - \cvec{z}^{\mathrm{u},-} \\
\cvec{w} - \cvec{z}^{\mathrm{u},-} 
\end{array}\right].
\label{eq:gains_ap}
\end{equation}
For the covariance we get:
\begin{align*}
&\left[\begin{array}{cc}
\cdiag{\Sigma}^{\mathrm{u}, +} & \cdiag{\Sigma}^{\mathrm{s}, +} \\
\cdiag{\Sigma}^{\mathrm{s}, +} & \cdiag{\Sigma}^{\mathrm{l}, +}\\
\end{array}\right]  =  
\left(\cmat{I}_n - \cmat{Q}\cmat{H} \right) \cmat{\Sigma}^- = \\ &
\left[\begin{array}{cc}
\cmat{I}_m - \cdiag{Q}^\mathrm{u} & \cmat{0}_{m \times m} \\
-\cdiag{Q}^\mathrm{l} & \cmat{I}_m 
\end{array}\right]
\left[\begin{array}{cc}
\cdiag{\Sigma}^{\mathrm{u}, -} & \cdiag{\Sigma}^{\mathrm{s}, -} \\
\cdiag{\Sigma}^{\mathrm{s}, -} & \cdiag{\Sigma}^{\mathrm{l}, -}\\
\end{array}\right] \\  =& 
\left[\begin{array}{cc}
\left(\cmat{I}_m - \cdiag{Q}^\mathrm{u} \right) \cdiag{\Sigma}^{\mathrm{u}, -} & \left(\cmat{I}_m - \cdiag{Q}^\mathrm{u} \right) \cdiag{\Sigma}^{\mathrm{s},-} \\
-\cdiag{Q}^\mathrm{l}\cdiag{\Sigma}^{\mathrm{u}, -} + \cdiag{\Sigma}^{\mathrm{s}, -} & -\cdiag{Q}^\mathrm{l} \cdiag{\Sigma}^{\mathrm{s},-} + \cdiag{\Sigma}^{\mathrm{l},-}
\end{array}\right].
\end{align*}
Hence the diagonals of the individual parts can be computed as 
\begin{align}
\cvec{\sigma}^{\mathrm{u},+} &= \left( \cvec{1}_m - \cvec{q}^\mathrm{u} \right) \odot \cvec{\sigma}^{\mathrm{u},-} \label{eq:var1_ap} \\
\cvec{\sigma}^{\mathrm{s},+} &= \left( \cvec{1}_m - \cvec{q}^\mathrm{u} \right) \odot \cvec{\sigma}^{\mathrm{s},-} \label{eq:var2_ap}\\
\cvec{\sigma}^{\mathrm{l},+} &= \cvec{\sigma}^{\mathrm{l}, -} - \cvec{q}^\mathrm{l} \odot \cvec{\sigma}^{\mathrm{s},-} \label{eq:var3_ap}.
\end{align}
\subsection{Root Mean Square Error Results}
To evaluate the actual prediction performance of our approach we repeated some experiments using the RMSE as loss function. Other than that and removing the variance output of the decoder no changes were made to the model, hyperparameters and learning procedure. The results can be found in \autoref{tab:rmse_res}.
\begin{table}[t]
\caption{RMSE Results}
\label{tab:rmse_res}
\vskip 0.15in
\centering
\begin{tabular}{lc}
\toprule
Model & RMSE  \\
\midrule
\multicolumn{2}{c}{Pendulum} \\
\midrule
RKN ($ m = 15, b = 3, K = 15$) & $0.0779 \pm 0.0082$ \\
RKN ($ m = b = 15, K = 15$) & $0.0758 \pm 0.0094$ \\ 
LSTM ($ m = 50 $) & $0.0920 \pm 0.0774$ \\
LSTM ($ m = 6 $)  & $ 0.0959 \pm 0.0100$ \\
GRU ($ m = 50 $)  & $ 0.0821 \pm 0.0084$ \\
GRU ($ m = 8 $)   & $0.0916 \pm  0.0087$ \\
\midrule
\multicolumn{2}{c}{Multiple Pendulums} \\
\midrule
RKN ($m = 45, b = 3, k = 15$) & $0.0878 \pm 0.0036$ \\ 
LSTM ($m = 50$) & $0.098 \pm 0.0036$  \\ 
LSTM ($m = 12$) & $0.104 \pm 0.0043$ \\
GRU ($m = 50$)  & $0.112 \pm 0.0371$  \\
GRU ($m = 14$)  & $0.105 \pm 0.0055$ \\
\midrule
\multicolumn{2}{c}{Quad Link (without additional noise )} \\
\midrule
RKN ($m = 100, b = 25, k = 15$) & $0.103 \pm 0.00076$ \\
LSTM ($m = 100$) & $0.175 \pm 0.182$  \\
LSTM ($m = 25$) & $0.118 \pm 0.0049$  \\
GRU ($m = 100$) &  $0.278 \pm 0.105$ \\ 
GRU ($m = 25$) &  $0.121 \pm 0.0021$ \\
\midrule
\multicolumn{2}{c}{Quad Link (with additional noise )} \\
\midrule
RKN ($m = 100, b = 25, k = 15$) & $0.171 \pm 0.0039$ \\
LSTM ($m = 75$) & $0.175 \pm 0.0022$  \\
GRU ($m = 25$) &  $0.204 \pm 0.0023$ \\ 
\bottomrule
\end{tabular}
\end{table}

\subsection{Visualization of Imputation Results}
\begin{figure*}[t]
\subfigure[Pendulum]{\includegraphics[width=.45\textwidth]{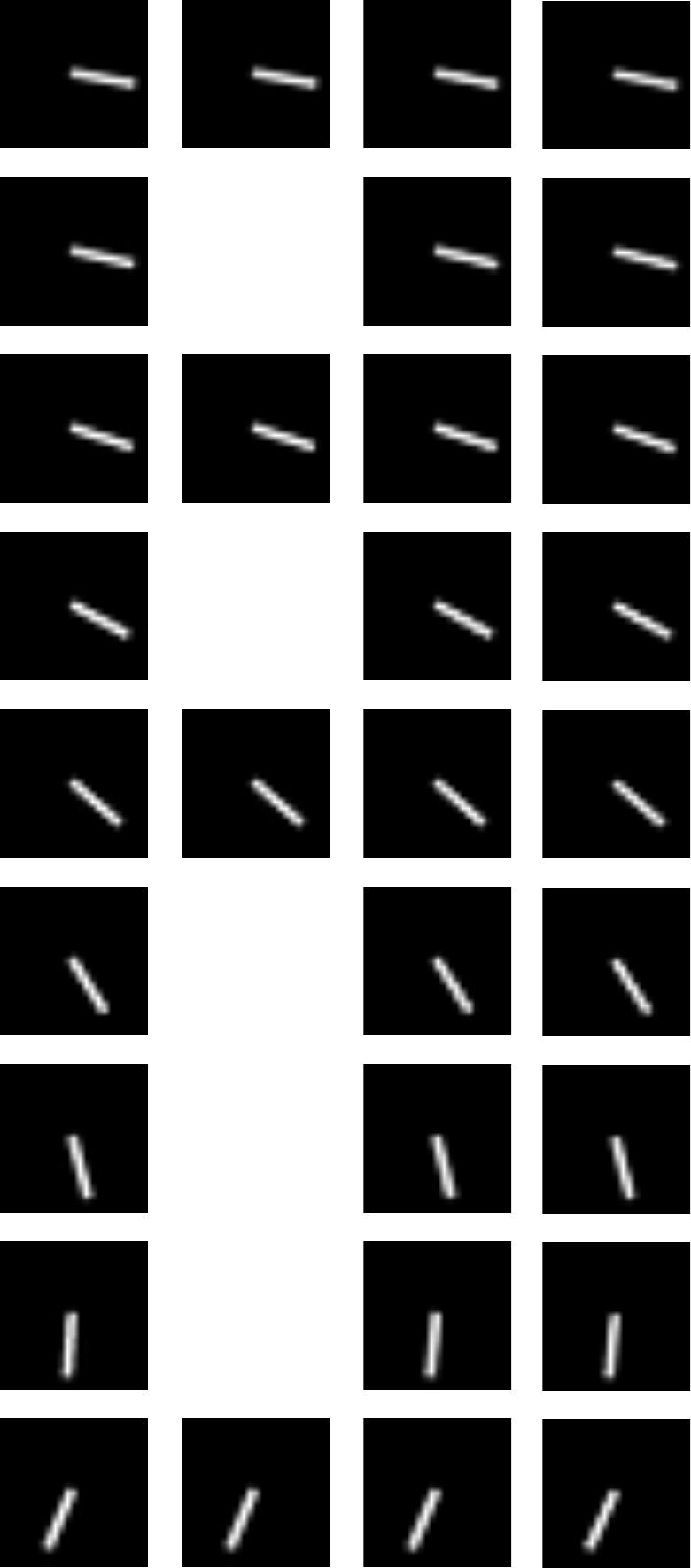}}%
\hspace{0.1\textwidth}%
\subfigure[Quad Link]{\includegraphics[width=.45\textwidth]{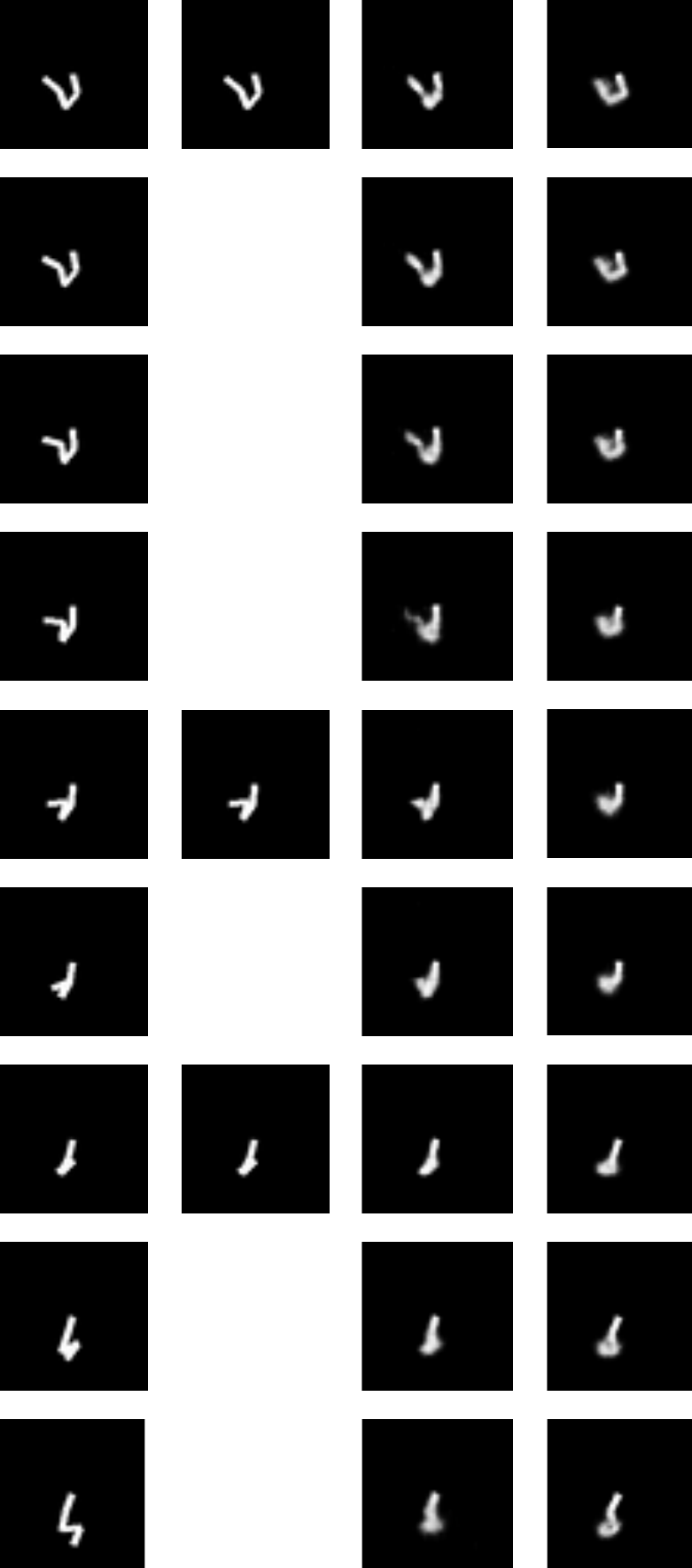}}
\caption{Each of (a) and (b) shows from left to right: true images, input to the models, imputation results for RKF, imputation results for KVAE\cite{fraccaro2017kvae}.}
\label{fig:imp_res}
\end{figure*}

Exemplary results of the data imputation experiments conducted for the Pendulum and Quad Link experiment can be found in \autoref{fig:imp_res}

\subsection{Network Architectures and Hyper Parameters}
\label{sec:apendixHP}

For all experiments Adam \cite{kingma2014adam} with default parameters ($\alpha=10^{-3}$, $\beta_1 = 0.9$, $\beta_2 = 0.999$ and $\varepsilon=10^{-8}$) was used as an optimizer. The gradients were computed using (truncated) Backpropagation Trough Time (BpTT) \cite{werbos1990bptt}. Further in all (transposed) convolutional layers layer normalization (LN) \cite{ba2016layer} was employed to normalize the filter responses. "Same" padding was used. The elu activation function \cite{clevert2015elu} plus a constant 1 is denoted by $(\textrm{elu} + 1)$ was used to ensure that the variance outputs are positive.

\subsubsection{Pendulum and Multiple Pendulum Experiments}
\begin{figure*}[b]
    \centering
    \includegraphics[width=\textwidth]{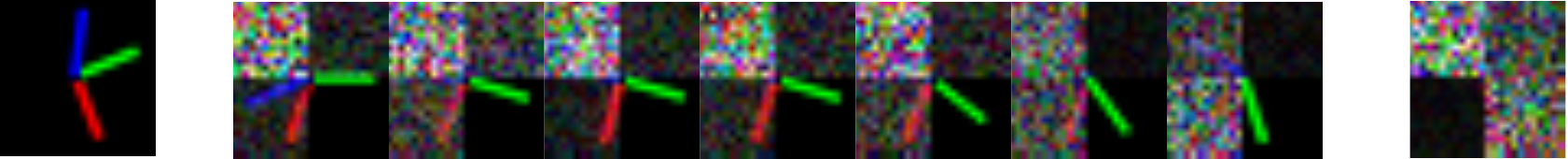}
    \caption{Example images for the multiple pendulum experiments. \textbf{Left}: Noise free image. \textbf{Middle}: sequence of images showing how the noise affects different pendulums differently. \textbf{Right}: Image without useful information.}
    \label{fig:mult_pend_exp}
\end{figure*}

\paragraph{Observations.} \emph{Pendulum}: Grayscale images of size $24 \times 24$ pixels.  \emph{Multiple Pendulum}: RGB images of size $24 \times 24$ pixels. See \autoref{fig:mult_pend_exp} for examples. \\
\paragraph{Dataset.}
$1000$ Train and $500$ Test sequences of length $150$. For the filtering experiments noise according to section \ref{sec:PendulumDataSet} was added , for imputation $50\%$ of the images were removed randomly. \\
\paragraph{Encoder.} 2 convolution + 1 fully connected + linear output \& $(\textrm{elu} + 1)$ output:
\begin{itemize}
	\item Convolution 1: 12, $5 \times 5$ filter, ReLU, $2 \times 2$ max pool with $2 \times 2$ stride   
	\item Convolution 2: 12, $3 \times 3$ filter with $2 \times 2$ stride, ReLU, $2 \times 2$ max pool with $2 \times 2$ stride 
	\item \emph{Pendulum}: Fully Connected 1: 30, ReLU
    \item \emph{Multiple Pendulum}: Fully Connected 1: 90, ReLU
	\end{itemize}
\textbf{Transition Model} \emph{Pendulum}: 15 dimensional latent observation, 30 dimensional latent
state. \emph{Multiple Pendulum}: 45 dimensional latent observation, 90 dimensonal latent state. \emph{Both:}  bandwidth: 3, number of basis: 15
\begin{itemize}
    \item $\alpha(\cvec{z}_t)$: No hidden layers - softmax output  
\end{itemize}

\paragraph{Decoder (for $\cvec{s}^+_t$).} 1 fully connected + linear output:
\begin{itemize}
	\item Fully Connected 1: 10, ReLU
\end{itemize} 
\paragraph{Decoder (for $\cvec{o}^+_t$).}: 1 fully connected + 2 transposed convolution + transposed convolution output:
\begin{itemize}
	\item Fully Connected 1: 144 ReLU
	\item Transposed Convolution 1: 16, $5 \times 5$ filter with $4 \times 4$ stride, ReLU
 	\item Transposed Convolution 2: 12, $3 \times 3$ filter with $2 \times 2$ stride, ReLU
	\item Transposed Convolution Out: \emph{Pendulum}: 1 \emph{Multiple Pendulum}: 3, $3 \times 3$ filter with $1 \times 1$ stride, Sigmoid
\end{itemize}
\paragraph{Decoder (for $\cvec{\sigma}_t^+$ or $\sigma_t^+$).} 1 fully connected + $(\textrm{elu} + 1)$:
\begin{itemize}
	\item Fully Connected 1: 10, ReLU
\end{itemize} 
\subsubsection{Quad Link}
\paragraph{Observations.} Grayscale images of size $48 x 48$ pixels. \\
\paragraph{Dataset.} $4000$ Train and $1000$ Test sequences of length 150. For the filtering with additional noise experiments noise according to section D was added, for imputation $50\%$ of the images were removed randomly. \\
\paragraph{Encoder.} 2 convolution + 1 fully connected + linear output \& $(\textrm{elu} + 1)$ output:
\begin{itemize}
	\item Convolution 1: 12, $5 \times 5$ filter with $2 \times 2$ stride, ReLU, $2 \times 2$ max pool with $2 \times 2$ stride   
	\item Convolution 2: 12, $3 \times 3$ filter with $2 \times 2$ stride, ReLU, $2 \times 2$ max pool with $2 \times 2$ stride 
	\item Fully Connected 1: 200 ReLU
\end{itemize}
\paragraph{Transition Model.} 100 dimensional latent observation, 200 dimensional latent
state, bandwidth: 3, number of basis: 15
\begin{itemize}
    \item $\alpha(\cvec{z}_t)$: No hidden layers - softmax output  
\end{itemize}
\paragraph{Decoder (for $\cvec{s}^+_t$).} 1 fully connected + linear output:
\begin{itemize}
	\item Fully Connected 1: 10, ReLU
\end{itemize} 
\paragraph{Decoder (for $\cvec{o}^+_t$).} 1 fully connected + 2 transposed convolution + transposed convolution output:
\begin{itemize}
	\item Fully Connected 1: 144 ReLU
	\item Transposed Convolution 1: 16, $5 \times 5$ filter with $4 \times 4$ stride, ReLU
 	\item Transposed Convolution 2: 12, $3 \times 3$ filter with $4 \times 4$ stride, ReLU
	\item Transposed Convolution Out: 1, $1 \times 1$ stride, Sigmoid
\end{itemize}
\paragraph{Decoder (for $\cvec{\sigma}_t^+$ or $\sigma_t^+$.} 1 fully connected + $(\textrm{elu} + 1)$:
\begin{itemize}
	\item Fully Connected 1: 10, ReLU
\end{itemize} 
\subsubsection{Kitti}
\paragraph{Observation and Data Set.} For this experiment, our encoder is based on the pose network proposed by \cite{zhou2017sfmlearner} which helps us to speed-up the training process. Specifically we extract features from the conv6 layer of the pose network by running the model on the KITTI odometry dataset. The training dataset for this experiment comprised of sequences 00, 01, 02, 08, 09. Sequences 03, 04, 05, 06, 07, 10 were used for testing. \\
\paragraph{Encoder.} Pose Network of \cite{zhou2017sfmlearner} up to layer conv6 + 1 Convolution
\begin{itemize}
    \item Convolution 1: 50, 1x1 filter,with 1x1 stride
\end{itemize}
\paragraph{Transition Model.}
50 dimensional latent observations, 100 dimensional latent state, bandwidth 1, number of basis 16
\begin{itemize}
    \item $\alpha(\cvec{z}_t)$: No hidden layers - softmax output  
\end{itemize}
\paragraph{Decoder (for $\cvec{s}^+_t$).} 2 fully connected + linear output:
\begin{itemize}
	\item Fully Connected 1: 50, ReLU
\end{itemize}
\subsubsection{Pneumatic Brook Robot Arm}
\paragraph{Observations and Data Set.}
$6$ sequences of $30,000$ samples each of input currents and observed joint positions, sampled at 100Hz. $5$ sequences were used for training, $1$ for testing.
\paragraph{Encoder.} 1 fully connected + linear output \& $(\textrm{elu} + 1)$ output.
\begin{itemize}
    \item Fully Connected, 30 ReLU 
\end{itemize}
\paragraph{Transition Model.} 30 dimensional latent observation, 60 dimensional latent state, bandwidth 3, number of basis 32

\paragraph{Decoder (for $\cvec{s}^+_t$.} 1 fully connected + linear output
\begin{itemize}
    \item 30 ReLU
\end{itemize}

\subsection{Observation Noise generation process}
\label{sec:PendulumDataSet}
Let $\mathcal{U}(x,y)$ denote the uniform distribution from $x$ to $y$.To generate the noise for the pendulum task for each sequence a sequence of factors $f_t$ of same length was generated. To correlate the factors they were sampled as $f_0 \sim \mathcal{U}(0, 1)$  and $f_{t+1} = \text{min}(\text{max}(0, f_t + r_t), 1)$ with $r_t \sim \mathcal{U}(-0.2, 0.2)$. Afterwards, for each sequence two thresholds $t_1 \sim \mathcal{U}(0.0, 0.25)$ and $t_2 \sim \mathcal{U}(0.75, 1)$ were sampled. All $f_t < t_1$ were set to $0$, all $f_t > t_2$ to 1 and the rest was linearly mapped to the interval $[0, 1]$. Finally, for each image $\cvec{i}_t$ an image consisting of pure uniformly distributed noise $\cvec{i}_t^{noise}$ was sampled and the observation computed as $\cvec{o}_t$ = $f_t \cdot \cvec{i}_t + (1 - f_t) \cdot \cvec{i}_t^{noise}$.